\documentclass[11pt]{article}
\usepackage[margin=1in]{geometry}
\usepackage{times}
\usepackage{setspace}
\usepackage[hyphens]{url}
\usepackage{graphicx}
\usepackage[round]{natbib}
\usepackage{caption}
\usepackage{algorithm}
\usepackage{algorithmic}
\usepackage{booktabs}
\usepackage{placeins}
\usepackage{subcaption}
\usepackage{amsmath,amssymb,amsfonts}
\usepackage{mathtools}
\usepackage{amsthm}
\usepackage{xcolor}
\newcommand{\rev}[1]{#1}   
\newcommand{\upd}[1]{#1}   
\newcommand{\Kd}[1]{\mathcal{K}^{\dagger}_{#1}}

\newcommand{\T}{\mathcal{T}}
\newcommand{\nrm}[1]{\lVert #1\rVert}
\newtheorem{theorem}{Theorem}

\newtheorem{definition}{Definition}
\newtheorem{proposition}{Proposition}
\newtheorem{remark}{Remark}
\newtheorem{corollary}{Corollary}
\newtheorem{lemma}{Lemma}

\theoremstyle{definition}

\theoremstyle{plain}
\newcommand{\coderepo}{\url{https://anonymous.4open.science/r/opq-RL-2027-0D23}}
\DeclareMathOperator{\Tr}{Tr}
\DeclareMathOperator{\ad}{ad}

\DeclareMathOperator{\LSE}{LSE}
\title{Operator-Valued Q-Learning: Compositional Bellman Recursions for Order-Sensitive Decision Making}

\author{Swagatam Das \\ Electronics and Communication Sciences Unit, \\ Indian Statistical Institute, Kolkata. \and Swagato Das \\ Indian Statistical Institute, Kolkata.}
\date{}

\usepackage[hidelinks]{hyperref}  
\begin{document}
\maketitle

\begin{abstract}
In quantum pulse sequences, spacecraft attitude control, and in-hand
reorientation, actions act on a hidden physical state as non-commuting
operators, so the outcome of a plan depends on the order of its actions.
We let the value function carry this structure: each $Q(s,a)$ is an
$N\times N$ operator, and the Bellman backup applies the action's
dual map to the continuation value,
$Q(s,a)=\Kd{a}\bigl(R(s,a)+\gamma\,\mathbb{E}\,V(s')\bigr)$. Unrolled along a
plan, the recursion builds the ordered product of the actions' operators,
and the value at any latent state is a linear trace readout.
\rev{This is the operator form of $\alpha$-vector evaluation:} it is exact for policies
that do not observe the latent, the backup contracts for unitary and
orthogonal dynamics, and an asynchronous operator temporal-difference (TD)
scheme learns it with almost-sure convergence.
\rev{Control is where the operator setting departs from classical RL and
POMDPs. Classical values live in a lattice, so a maximum over actions always
exists; self-adjoint operators form an anti-lattice (Kadison), and there is no
maximum.} An operator log-sum-exp nevertheless gives one operator
per state--action pair whose readout upper-bounds the value of every policy at
every latent state\rev{, and to which the sampled update provably converges for a
finite horizon}, reducing to soft $Q$-learning \rev{and to the QMDP bound} when
actions commute\rev{; we characterise what fails without a lattice and give a
sufficient condition for near-optimal greedy control}. We
also prove that scalar values on observations that ignore action order
must confuse swapped histories, that exact history tables
grow exponentially for free action groups but polynomially for nilpotent
ones, and a Hankel-rank lower bound on the operator dimension.
\rev{In control experiments with model-matched baselines and planners, the learned
operator policy matches 1000-simulation tree search on $SO(3)$ pointing for
horizons up to $20$ at about $10^4$ times lower decision cost, transfers
zero-shot to unseen initial states, and, when the initial state is hidden, acts
exactly on its belief mean, where plugging the mean into a DQN does not; its
certificate stays valid at every horizon. A planner with the same model
knowledge is more sample-efficient at short horizons, and a DQN given the
operators eventually matches or exceeds OpQ; keeping a few operators per state
closes the remaining gap to the optimum.}
\end{abstract}

\section{Introduction}\label{sec:intro}

Value functions in reinforcement learning (RL) are scalars
\citep{sutton2018rl,watkins1992q}, or, in distributional RL, probability
measures \citep{bellemare2017distributional}; either way, their values can
be combined in any order. Yet in quantum control, attitude control, and
robotic manipulation, actions compose as non-commuting operators, and the
order of a plan determines its outcome. The issue is particularly relevant
when each episode is expensive, as in quantum and robotic control.
\rev{Tracking the latent and fitting a generic function approximator to it
learns the order-dependence from data without guarantees; planning over
action histories is exact but its cost grows as $K^T$ per decision.}
We show that the value
function itself can carry this non-commutative structure, with exact
representation, convergence and certificate guarantees. Entropy-regularised
RL \citep{haarnoja2018soft,ziebart2010maximum} supplies the aggregation we
use for control.

\paragraph{A simple example of the problem.}
Consider a rectangular object, and rotate it $90^\circ$ about its
vertical axis, then $90^\circ$ about its horizontal axis. Now perform
the same two rotations in the opposite order: the object ends up
facing a different way. The same signature appears whenever a quantum
gate is applied before versus after another: the objects being
composed are matrices, and matrix multiplication does not commute. If
the agent's observation records \emph{which} actions were taken (a
step counter, action counts, a task phase) but not \emph{in which
order}, a scalar $Q$-function on that observation assigns the same
number to both histories. Section~\ref{sec:osMDP} makes this precise:
with three steps or more, two histories that differ only by a swap
receive values that no scalar function of an order-blind observation can
both match (Proposition~\ref{thm:representation_gap}).

\begin{figure}[t]
  \centering
  \includegraphics[width=0.97\linewidth]{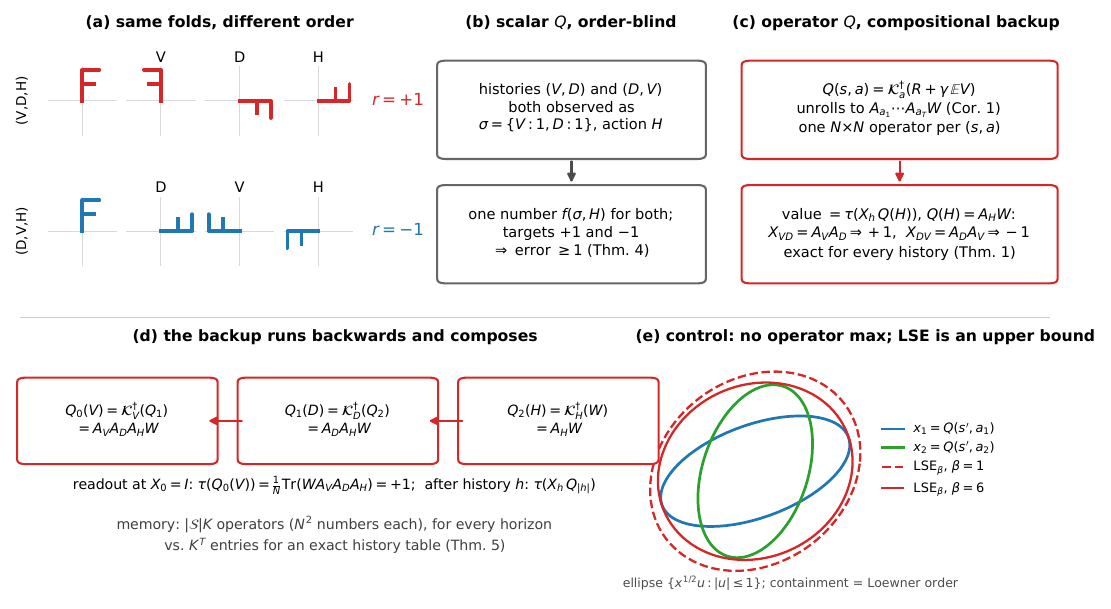}
  \caption{\textbf{The idea in one example} (Appendix~\ref{app:warmup}).
  (a)~$(V,D,H)$ and $(D,V,H)$ use the same folds but end mirrored, with
  rewards $\pm1$. (b)~After two steps both histories contain one $V$ and
  one $D$, so a scalar value on this order-blind observation errs by at
  least $1$ (Proposition~\ref{thm:representation_gap}). (c)~The compositional
  backup forms the ordered product, and one operator $Q(H)=A_HW$, read
  against the history operator $X_h$, returns both values exactly
  (Proposition~\ref{thm:exact_rep}). (d)~Unrolled, the recursion applies the dual
  maps backwards along the plan, so each stored operator already contains
  the rest of the plan (Cor.~\ref{cor:product}). (e)~For control there is
  no operator maximum (Kadison); the operator log-sum-exp is a Löwner upper
  bound of every $Q(s',a')$ (ellipse containment), which yields the
  upper-bound certificate of Theorem~\ref{thm:certificate}.}
  \label{fig:overview}
\end{figure}

In \textbf{quantum control}, an agent selects unitary pulses
$\{U_k=e^{-iH_k\Delta t}\}$; since $[H_j,H_k]\neq0$ for generic
Hamiltonians, pulse $j$ then $k$ reaches a physically different quantum
state than $k$ then $j$ \citep{niuclosedloop2022}. In \textbf{rigid-body
manipulation}, control inputs induce rotations composing the same way,
formalised by $SO(3)$ \citep{schuck2025so3} and inherited by in-hand
reorientation, spacecraft attitude control, and locomotion on $SE(3)$.
In each domain a plan $a_1,\dots,a_T$ has cumulative effect equal to an
\emph{ordered product} $X_T=A_{a_1}\cdots A_{a_T}$, with reward
$r=\tfrac1N\Tr(WX_T)$ for an observable $W$. Figure~\ref{fig:overview} shows the phenomenon and the operator
remedy, on a three-fold example worked out in Appendix~\ref{app:warmup}.

\paragraph{Observing the latent directly.}
If the latent $X_t$ is observed, the process on $(s_t,X_t)$ \emph{is}
Markov, and a scalar $Q(s,X,a)$ suffices in principle, but $X_t$ ranges over
a continuous non-abelian group, so a generic approximator must learn a
nonlinear-looking function on it.
\rev{If $X_t$ is not observed but the action maps and $X_0$ are known, it can
be tracked exactly from the agent's own actions, as a belief is in a POMDP;
exact tabular history summaries, by contrast, need $K^{T}$ entries for free
action groups (Proposition~\ref{thm:horizon}).}
Our representation is \emph{linear} in the latent, so one operator per
state--action pair is valid at every latent state, including \rev{unseen ones}
(Proposition~\ref{thm:exact_rep})\rev{, without retraining or
augmentation and with a decision cost that does not grow with the horizon}.

The central idea is to place order information in the \emph{codomain}
of the value function. If $Q(s,a)$ is an operator and the Bellman backup transports the
continuation value through the action's operator,
$Q(s,a)=\Kd{a}(R+\gamma\,\mathbb{E}V)$, then unrolling the recursion
produces exactly the ordered product $A_{a_1}\cdots A_{a_T}$, and the
scalar value at any latent state is a trace readout. For every policy
that does not look at the latent state this representation is
\emph{exact} (Proposition~\ref{thm:exact_rep}); the history enters only
through an $N^2$-dimensional \emph{history operator} $X_h$, which the
agent can maintain by one matrix product per step.

\paragraph{Operator aggregation for control.}
Control requires aggregating $\{Q(s',a')\}_{a'}$ into a continuation
value. For scalars this is a maximum; for self-adjoint operators the
Löwner order is only partial, and by Kadison's anti-lattice theorem two
incomparable self-adjoint operators have \emph{no} least upper bound
\citep{kadison1951order}.
\rev{However non-commuting a POMDP's transitions, its $\alpha$-vectors lie in a
lattice whose join gives the QMDP and fast-informed bounds
\citep{hauskrecht2000value}; the PSD cone is not a lattice
\citep[Thm.~3.21]{aliprantis2007cones} (Remark~\ref{rem:pomdp}).}
We use the operator log-sum-exp, which is a
Löwner upper bound of every argument, and prove that the resulting
soft operator recursion certifies an upper bound on the \rev{value of every
policy} at every latent state (Theorem~\ref{thm:certificate}); in the commuting
case it reduces exactly to soft $Q$-learning (Prop.~\ref{prop:reduction})
\rev{and to the QMDP bound}.
We also delimit what does \emph{not} survive: the operator log-sum-exp is
not monotone and not non-expansive outside the commuting sector
(Props.~\ref{prop:lse_lipschitz_new},~\ref{prop:monotone_new}).

\paragraph{Scope.} Three settings should be kept apart. The exact
representation (Proposition~\ref{thm:exact_rep}) concerns
\emph{latent-blind policy evaluation}: the policy sees only $s_t$, and its
value is linear in the latent. The certificate (Thm~\ref{thm:certificate})
bounds the value of every policy, latent-aware ones included. The control
experiments \rev{except E23} are \emph{latent-aware}: the agent observes $\rho_t$, or can
compute it from the action history because $\rho_0$ and the $\mathcal K_a$
are known, and uses it only to perform the greedy readout
$\arg\max_a\tau(\rho_tQ(s_t,a))$.
\rev{OpQ uses the action maps $\mathcal K_a$ but not the observable transition
kernel $P$, and its headline variant learns $R$ from scalar rewards; E19
compares it with learners and planners given exactly this knowledge.}
We do not claim to solve a POMDP in which
the latent must be inferred from rewards alone, nor the case of unknown
action maps, which we leave open (Sec.~\ref{sec:conclude}).
\rev{The motivating case of an unknown, unmeasurable initial state (e.g.\ a
single copy of an unknown quantum input) is covered by
Proposition~\ref{thm:exact_rep} and Theorem~\ref{thm:certificate}; E23 tests
it end to end with an initial state that is drawn from a prior with known mean
and never observed.}

\paragraph{\rev{Relation to prior work.}}
\rev{The closest connection is to $\alpha$-vector representations in POMDPs
\citep{smallwood1973optimal,kaelbling1998planning}. With the latent
vectorised, Proposition~\ref{thm:exact_rep} is the operator (Heisenberg
picture) form of blind-policy $\alpha$-vector evaluation
\citep{hauskrecht2000value}, which we state because everything else builds on
it. The new content concerns control without a lattice
(Remark~\ref{rem:pomdp}); keeping a small set of operators per state, as
point-based methods do \citep{pineau2003point}, closes the remaining gap (E20).}
Because the latent carries a representation of the action algebra, the
required operator dimension is related to the Hankel rank of the
ordered-product readout (Proposition~\ref{thm:upper}), and the nilpotency step
of the algebra controls the size of exact history statistics
(Proposition~\ref{thm:horizon}).

The ordered-product readout is a weighted
automaton, and our dimension results import Hankel-rank theory
\citep{carlyle1971realizations,fliess1974matrices}, as used for
observable operator models and predictive state representations
\citep{jaeger2000observable,littman2002predictive,balle2014spectral};
\rev{planning with such linear predictive representations is studied by
\citet{evendar2005planning}, and quantum analogues of POMDPs by
\citet{barry2014quantum}.}
Our contribution is to place these operators \emph{inside} a Bellman
recursion with an order-theoretic aggregation. Distributional RL
\citep{bellemare2017distributional} enriches the codomain with measures
that commute under convolution. Bayesian Bellman operators
\citep{fellows2021bayesian}, operator DQN \citep{liu2022operatordqn},
quantum RL \citep{dong2008qrl,skolik2022quantumagents}, geometric RL on
$SO(3)$ \citep{schuck2025so3}, free-probability embeddings
\citep{hayase2025free}, recurrent \citep{seyedsalehi2023rql} and equivariant
\citep{wang2021equivariant} $Q$-learning, and the analyses of
\citet{lee2024finaliteration,lee2023discretetime,meyn2024projectedbellman}
all retain scalar Bellman equations.
\rev{Model-based planners \citep{sutton1990dyna,kocsis2006bandit} with OpQ's
knowledge are compared in E19; gradient pulse optimisation (GRAPE;
\citealp{khaneja2005optimal}) plans continuous pulses with a known model, and
for a discrete calibrated gate set its counterpart is this search over gate
words.}
\citet{hoeftmann2026relative} use a non-scalar Bellman operator on value
differences; \citet{opguidedinvariance2026} prove value preservation when
Lie-group operators commute with the generator, parallel to
Proposition~\ref{prop:reduction}.

\paragraph{Contributions.}
(1) A formal problem class, operator-latent decision processes, that
separates latent-aware from latent-blind control and makes the
state-aliasing structure explicit (Sec.~\ref{sec:opq}).
(2) A \emph{compositional} operator Bellman recursion: exact representation
of latent-blind action values \rev{in operator $\alpha$-vector form}
(Proposition~\ref{thm:exact_rep}), contraction
and almost-sure convergence of an asynchronous operator TD scheme
\rev{for policy evaluation} under its sampling and step-size assumptions
(Thm~\ref{thm:td}), and \rev{one-step} policy
improvement (Prop.~\ref{prop:improve}).
(3) \rev{Control without a lattice:} an operator soft-Bellman recursion whose
\rev{finite-horizon iterates are} Löwner
upper bounds on every policy's value (Thm~\ref{thm:certificate})\rev{, valid
for any Löwner upper-bound aggregator (Corollary~\ref{cor:anybound}), with
almost-sure convergence of its sampled update for a finite horizon
(Thm~\ref{thm:sampled})}, exact
reduction to soft $Q$-learning under commutativity
(Prop.~\ref{prop:reduction}), a corrected Lipschitz analysis of the
operator log-sum-exp, including counterexamples to $\beta$-free
constants (Prop.~\ref{prop:lse_lipschitz_new}), \rev{and a sufficient
condition for near-optimal greedy control (Prop.~\ref{prop:nearopt}).}
(4) Representation results for order, \rev{stated as consequences of the
classical results they build on}: an aliasing lower bound
(Proposition~\ref{thm:representation_gap}), a free-versus-nilpotent dichotomy
for exact \emph{tabular} history representation (Proposition~\ref{thm:horizon}),
and a Hankel-rank lower bound on the operator dimension that is tight for
generic full-rank instances (Proposition~\ref{thm:upper}).
(5) Numerical checks of every statement; control experiments against
tabular, DQN and recurrent baselines, \rev{model-matched learners and tree
search up to horizon 20 (E19), certificate tightening (E20), and tests of
miscalibration, $\beta$ and a hidden latent (E21--E23);} and representation
experiments on four non-abelian domains (Sec.~\ref{sec:experiments}); code at
\coderepo.

\section{Preliminaries}\label{sec:prelim}

\paragraph{Classical RL.}
For an MDP $(\mathcal{S},\mathcal{A},P,r,\gamma)$ the optimal $Q^*$ satisfies
$Q^*(s,a)=\mathbb{E}[r+\gamma\max_{a'}Q^*(s',a')]$, and the Bellman operator
is a $\gamma$-contraction in $\|\cdot\|_\infty$
\citep{watkins1992q,sutton2018rl}.

\paragraph{Operators.}
We work in $\mathcal{M}_N=\mathbb{C}^{N\times N}$ with the normalised
trace $\tau=\tfrac1N\Tr$, a faithful tracial state, so
$(\mathcal{M}_N,\tau)$ is the canonical finite-dimensional
non-commutative probability space \citep{nica2006lectures}; the scalar
case is $N=1$. We write $\nrm{\cdot}$ for the operator norm,
$\nrm{x}_2^2=\tau(x^*x)$ for the normalised Hilbert--Schmidt norm,
$\nrm{x}_1=\tau(|x|)$ for the normalised trace norm, and $x\preceq y$
for the Löwner order on self-adjoint operators. For a linear map
$\mathcal{K}$ on $\mathcal{M}_N$ its \emph{dual} (Heisenberg picture)
$\mathcal{K}^\dagger$ is defined by
$\tau\bigl(Y\,\mathcal{K}(\rho)\bigr)=\tau\bigl(\mathcal{K}^\dagger(Y)\,\rho\bigr)$.
For self-adjoint $x_1,\dots,x_K$ the operator log-sum-exp is
\begin{equation}\label{eq:lse}
  \LSE_\beta(x_1,\dots,x_K)=\tfrac1\beta\log\Bigl(\textstyle\sum_i e^{\beta x_i}\Bigr),
\end{equation}
defined by functional calculus \citep{bhatia1997matrix}.

\begin{remark}[Group and algebra views]\label{rem:lie}
Writing $A_a=\exp(\Omega_a)$ with $\Omega_a$ in a Lie algebra
$\mathfrak g\subset\mathfrak{gl}(N)$, the latent evolves as
$X_t=X_{t-1}\exp(\Omega_{a_t})$. For $\mathfrak g=\mathfrak{so}(3)$
the $\Omega_a$ are real skew-symmetric; for quantum pulses
$\Omega_a=-iH_a\Delta t$. The algebra view is used in
Sec.~\ref{sec:osMDP} to state nilpotency.
\end{remark}

\section{Operator-Valued Q-Learning}\label{sec:opq}

\subsection{Operator-latent decision processes}

\begin{definition}[Operator-latent decision process]\label{def:oldp}
An OLDP is a tuple $(\mathcal S,\mathcal A,P,\{\mathcal K_a\},R,\gamma)$:
a finite \emph{observable} state space $\mathcal S$ and action set
$\mathcal A$ ($|\mathcal A|=K$) with transition kernel $P(s'\mid s,a)$
that does not depend on the latent; a \emph{latent} $\rho_t\in\mathcal
M_N$ updated by linear maps, $\rho_{t+1}=\mathcal K_{a_t}(\rho_t)$; and
rewards linear in the latent, $r_t=\tau\bigl(R(s_t,a_t)\,\rho_{t+1}\bigr)$,
with $\nrm{R(s,a)}\le R_{\max}$. Rewards are real: in CP, $R$ and
$\rho$ are self-adjoint; in MP we use real matrices, or replace
$\tau(\cdot)$ by $\operatorname{Re}\tau(\cdot)$, which is still
linear over $\mathbb R$. We write
$\kappa:=\max_a\nrm{\Kd a}_{\mathcal M_N\to\mathcal M_N}$ for the
operator-norm bound of the dual maps.
\end{definition}

Two cases cover all our examples. In the \emph{multiplicative picture}
(MP), $\mathcal K_a(X)=XA_a$ with $A_a\in GL(N)$, so
$X_T=X_0A_{a_1}\cdots A_{a_T}$ and $\Kd a(Y)=A_aY$; $\kappa=\max_a\nrm{A_a}$,
which equals $1$ for orthogonal or unitary $A_a$. The terminal reward
$\tfrac1N\Tr(WX_T)$ of Sec.~\ref{sec:intro} is the case $R=W$ at the final
step and $R=0$ before (time is recorded in $s$). In the
\emph{conjugation picture} (CP), $\mathcal K_a(\rho)=U_a\rho U_a^*$ with
$U_a$ unitary and $\rho$ a density operator (quantum control; or the
dyad $\rho=nn^\top$ of a marked body axis in rigid-body control), so
$\Kd a(Y)=U_a^*YU_a$, and $\kappa=1$. In CP the dual maps are positive
and unital, and $R$ is self-adjoint.

A \emph{latent-aware} policy chooses actions from $(s_t,\rho_t)$; the
process $(s_t,\rho_t)$ is an MDP for such policies. A \emph{latent-blind}
policy $\pi(a\mid s)$ sees only $s_t$. When $s_t$ does not record the
order of past actions (e.g.\ $s_t$ is the step counter or the action
counts), latent-blind control is a POMDP with deterministic,
action-driven latent dynamics. If $\rho_0$ and the $\mathcal K_a$ are
known, a latent-blind agent can nevertheless reconstruct
$\rho_t=\mathcal K_{a_t}\circ\cdots\circ\mathcal K_{a_1}(\rho_0)$ from its
own actions; in MP with $X_0=I$ this is the \emph{history operator}
$X_h=A_{a_1}\cdots A_{a_t}$, an $N^2$-dimensional sufficient statistic of
the history.

\subsection{The compositional Bellman recursion}

For a latent-blind policy $\pi$ define, for $Q:\mathcal S\times\mathcal A\to\mathcal M_N$,
\begin{equation}\label{eq:bellman}
  (\T^\pi Q)(s,a)\;=\;\Kd a\Bigl(R(s,a)+\gamma\sum_{s'}P(s'\mid s,a)\sum_{a'}\pi(a'\mid s')\,Q(s',a')\Bigr).
\end{equation}
The continuation value is transported \emph{through the action's dual
map} before it is combined with the current action; this is where
composition happens. Unrolling~\eqref{eq:bellman} along a deterministic
plan in MP gives $Q=A_{a_1}A_{a_2}\cdots A_{a_T}W$, the ordered product.
(Setting $\Kd a=\mathrm{id}$ removes the composition and gives a
permutation-symmetric recursion whose target, in a single-state
terminal-reward problem, does not depend on the current action at all;
such a recursion cannot represent order, and~\eqref{eq:bellman} is the
minimal change that can.) For a finite horizon $H$ we write
$Q^\pi_t=\T^\pi Q^\pi_{t+1}$, $Q^\pi_H=0$.

\begin{proposition}[Operator $\alpha$-vector evaluation of latent-blind policies]\label{thm:exact_rep}
Let $\pi$ be latent-blind, and let
$q^\pi(s,\rho,a)=\mathbb E^\pi\bigl[\sum_{t\ge0}\gamma^tr_t\,\big|\,s_0=s,\rho_0=\rho,a_0=a\bigr]$
be its action value in the full process. For a finite horizon, the
backward recursion $Q^\pi_t=\T^\pi Q^\pi_{t+1}$ satisfies
$q^\pi_t(s,\rho,a)=\tau\bigl(\rho\,Q^\pi_t(s,a)\bigr)$ for all
$s,a$ and all $\rho\in\mathcal M_N$. For an infinite horizon with
$\gamma\kappa<1$, $\T^\pi$ has a unique fixed point $Q^\pi$ and
$q^\pi(s,\rho,a)=\tau(\rho\,Q^\pi(s,a))$. (For $\rho$ outside the
physical set, $q^\pi$ is defined by the same expectation and is linear in
$\rho$.)
\end{proposition}

\begin{corollary}[Ordered products are Bellman fixed points]\label{cor:product}
In MP with a single observable state per step, terminal reward $W$ and
a deterministic plan $(a_1,\dots,a_T)$, the recursion (steps $t=0,\dots,T-1$, action $a_{t+1}$ at step $t$) gives
$Q_t(a_{t+1})=A_{a_{t+1}}\cdots A_{a_T}W$, so the return from $X_0=I$ is
$\tau(Q_0(a_1))=\tfrac1N\Tr(WA_{a_1}\cdots A_{a_T})$. After a history
$h=(a_1,\dots,a_t)$, the value of the remaining plan is
$\tau(X_h\,Q_t(a_{t+1}))$.
\end{corollary}

The product readout used in Sec.~\ref{sec:osMDP} and in experiments
E8, E13--E16 is therefore not a separate model: it is the fixed point
of~\eqref{eq:bellman} along each plan.

\begin{theorem}[Contraction and convergence of operator TD]\label{thm:td}
(i) $\T^\pi$ is a $\gamma\kappa$-contraction in
$\nrm{Q}_{\infty,\mathrm{op}}:=\max_{s,a}\nrm{Q(s,a)}$; if every $\Kd a$
is also non-expansive in $\nrm\cdot_2$ (unitary conjugation, orthogonal
or unitary multiplication), it is a $\gamma$-contraction in
$\max_{s,a}\nrm{Q(s,a)}_2$.
(ii) Let $(s_k,a_k)$ be drawn i.i.d.\ from a distribution $\mu$ with full
support, $s'_k\sim P(\cdot\mid s_k,a_k)$, $a'_k\sim\pi(\cdot\mid s'_k)$,
and update only the visited pair,
\begin{equation}\label{eq:update}
  Q_{k+1}(s_k,a_k)=Q_k(s_k,a_k)+\alpha_k\Bigl(\Kd{a_k}\bigl(R(s_k,a_k)+\gamma\,Q_k(s'_k,a'_k)\bigr)-Q_k(s_k,a_k)\Bigr),
\end{equation}
with $\sum_k\alpha_k=\infty$, $\sum_k\alpha_k^2<\infty$. If
$\gamma\kappa<1$ then $Q_k\to Q^\pi$ almost surely, without projection.
\end{theorem}

Part (ii) is a statement about the specific asynchronous scheme
above (i.i.d.\ state--action sampling with full support, on-policy next
actions, square-summable steps); it is not a convergence result for
arbitrary behaviour policies.

The update samples the observable transition and the next action only;
it uses the dual maps $\Kd a$ and the reward observable $R$, which are
known in the control problems that motivate us (gate unitaries and a
target observable; rotation generators and a target attitude). When
the $A_a$ are unknown they can be fitted from returns (E8,
Sec.~\ref{sec:osMDP}); combining such fitting with TD, with guarantees,
is left open.

\subsection{Control}

\begin{proposition}[Latent-aware policy improvement]\label{prop:improve}
Let $\pi$ be latent-blind with $\gamma\kappa<1$ and let
$\pi^+(s,\rho)\in\arg\max_a\tau(\rho\,Q^\pi(s,a))$. Then
$V^{\pi^+}(s,\rho)\ge V^\pi(s,\rho)$ for all $(s,\rho)$.
\end{proposition}

Proposition~\ref{prop:improve} is the practical controller when the
latent is observed or reconstructed: evaluate a latent-blind base policy
in operator form and act greedily on the scalar readout. To obtain a
single operator per $(s,a)$ that speaks for \emph{all} policies, we
need an aggregation over actions in the operator codomain. In CP
($Q$ self-adjoint, $\Kd a$ positive and unital) define
\begin{equation}\label{eq:soft}
  (\T_\beta Q)(s,a)=\Kd a\Bigl(R(s,a)+\gamma\sum_{s'}P(s'\mid s,a)\,\LSE_\beta\bigl(\{Q(s',a')\}_{a'}\bigr)\Bigr).
\end{equation}

\begin{theorem}[Löwner certificate]\label{thm:certificate}
In CP, let $V^\beta_H=0$, $V^\beta_t(s)=\LSE_\beta(\{Q^\beta_t(s,a')\}_{a'})$ and
$Q^\beta_t(s,a)=\Kd a\bigl(R(s,a)+\gamma\sum_{s'}P(s'\mid s,a)V^\beta_{t+1}(s')\bigr)$
(the finite-horizon form of~\eqref{eq:soft}). For every
policy $\pi$---latent-aware and history-dependent included, with
$q^\pi_t$ conditioned on any history that reaches $(s,\rho)$ at step
$t$---every $a$, and every $\rho\succeq0$,
\[
  q^\pi_t(s,\rho,a)\;\le\;\tau\bigl(\rho\,Q^\beta_t(s,a)\bigr).
\]
In particular $q^*_t(s,\rho,a)\le\tau(\rho\,Q^\beta_t(s,a))$, and
$Q^\pi_t(s,a)\preceq Q^\beta_t(s,a)$ for every latent-blind $\pi$.
For an infinite horizon, no contraction is needed: for every policy and
every $n\ge1$,
$q^\pi(s,\rho,a)\le\tau\bigl(\rho\,Q^{\beta,(n)}_0(s,a)\bigr)+\gamma^nR_{\max}\tau(\rho)/(1-\gamma)$,
where $Q^{\beta,(n)}$ is the $n$-step recursion. If, in addition, $\T_\beta$
contracts on the ball $\{\nrm{Q}_{\infty,\mathrm{op}}\le M_\beta\}$,
$M_\beta=(R_{\max}+\gamma\beta^{-1}\log K)/(1-\gamma)$---with modulus
$\gamma$ on the commuting sector, or $\gamma e^{2\beta M_\beta}$ in general
(Prop.~\ref{prop:lse_lipschitz_new})---then its fixed point is itself a
certificate.
\end{theorem}

The certificate needs only positivity of $\mathcal K_a$ and the Löwner
bound $\LSE_\beta(x)\succeq x_i$; it does \emph{not} need monotonicity
of $\LSE_\beta$, which fails in general (Prop.~\ref{prop:monotone_new}).
Its slack has two sources: softness ($\beta^{-1}\log K$ per step) and
non-commutativity---by Kadison's theorem no single operator can be the
least upper bound of incomparable $Q(s',a')$. E17 measures both.

\begin{proposition}[Exact commutative reduction]\label{prop:reduction}
In CP, suppose all $U_a$ and $R(s,a)$ lie in a common maximal abelian
subalgebra $\mathcal D$ (simultaneously diagonal in a basis $\{e_j\}$).
Then (i) $\T_\beta$ preserves $\mathcal D$ and acts on each coordinate
$j$ as the classical soft Bellman operator
$q_j(s,a)=r_j(s,a)+\gamma\sum_{s'}P(s'\mid s,a)\,\mathrm{lse}_\beta(q_j(s',\cdot))$;
(ii) for a pure eigenstate $\rho=Ne_je_j^*$ the certificate of
Theorem~\ref{thm:certificate} is the soft-optimal value of coordinate
$j$, exceeding the optimal value by at most
$\beta^{-1}\log K\sum_{k=1}^{H-t-1}\gamma^{k}$ and converging to it as
$\beta\to\infty$; (iii) for $N=1$, $\T_\beta$ is soft $Q$-learning's
Bellman operator.
\end{proposition}

In the commuting sector only the diagonal of the latent affects values,
and it never changes ($U_a^*YU_a=Y$ for $Y\in\mathcal D$). Order is then
irrelevant and nothing is lost relative to classical RL; the operator
structure is paid for only when a non-commuting action appears.

\begin{remark}[Why control differs from POMDPs]\label{rem:pomdp}
\rev{Order-sensitivity alone is generic: any POMDP with non-commuting
transition matrices has it. What is specific is the order structure of the
value space, not whether the physical transition operators commute. POMDP
$\alpha$-vectors live in a lattice, so the max-backup and QMDP exist whatever
the transition matrices; Proposition~\ref{prop:reduction} recovers this only
for the commuting sector of our formulation and is not a reduction of general
POMDPs. Self-adjoint operators do not form a lattice: $x_1=\mathrm{diag}(1,0)$
and $x_2=\mathrm{diag}(0,1)$ have the upper bounds $I$ and
$Y=\left(\begin{smallmatrix}2&1.2\\1.2&2\end{smallmatrix}\right)$; $I$ is
minimal (an upper bound $Z\preceq I$ must equal $I$) yet $I\not\preceq Y$, so
no least upper bound, and hence no max-backup over these two action values,
exists. Among diagonal operators $I$ is the join, the coordinatewise maximum,
which is why Proposition~\ref{prop:reduction} recovers QMDP. Monotonicity and
non-expansiveness fail with the lattice (Props.~\ref{prop:lse_lipschitz_new},
\ref{prop:monotone_new}). Unitary and orthogonal dynamics also never forget
order (E10: a constant per-depth commutator term), whereas scrambling
stochastic maps do.}
\end{remark}

\begin{corollary}[Any Löwner upper bound certifies]\label{cor:anybound}
\rev{Let $G$ map $K$-tuples of self-adjoint operators to self-adjoint operators
with $G(x_1,\dots,x_K)\succeq x_i$ for every $i$. Replacing $\LSE_\beta$ by $G$
in~\eqref{eq:soft}, the conclusions of Theorem~\ref{thm:certificate} for finite
horizons hold unchanged.}
\end{corollary}

\rev{The trace-minimal choice $\arg\min\{\tau(Y):Y\succeq x_i\ \forall i\}$ (an SDP)
is a minimal upper bound; by Kadison's theorem none is least.}

\begin{proposition}[When the greedy readout is near-optimal]\label{prop:nearopt}
\rev{In CP with a finite horizon, let $\hat Q_t(s,a)$ satisfy
$\tau(\rho\hat Q_t(s,a))\ge q^*_t(s,\rho,a)$ for all $(s,\rho,a)$, as $Q^\beta_t$
does, let $\delta_t(s,\rho,a)=\tau(\rho\hat Q_t(s,a))-q^*_t(s,\rho,a)$ be its
slack, and let $\hat\pi$ act greedily on $\tau(\rho_t\hat Q_t(s_t,a))$. With
$\varepsilon_t=\mathbb E^{\hat\pi}\bigl[\max_a\delta_t(s_t,\rho_t,a)-\min_a\delta_t(s_t,\rho_t,a)\bigr]$,
\[
  V^*(s_0,\rho_0)-V^{\hat\pi}(s_0,\rho_0)\;\le\;\sum_{t=0}^{H-1}\gamma^t\varepsilon_t.
\]
In the commuting sector at pure eigenstates,
$\varepsilon_t\le\beta^{-1}\log K\sum_{k=1}^{H-t-1}\gamma^k$ by
Prop.~\ref{prop:reduction}(ii), so the greedy policy becomes optimal as
$\beta\to\infty$.}
\end{proposition}

\rev{What matters is how the slack varies across actions, not its size: on
SO(3) pointing the certificate is $1.3\times V^*$, its spread $2\%$ of $V^*$,
and greedy control within $0.02$ of optimal (E20). The condition is sufficient,
not necessary, and needs the model.}

\begin{theorem}[Sampled OpQ-soft converges for a finite horizon]\label{thm:sampled}
\rev{In CP with horizon $H$ and $Q_H=0$, let each transition $(t,s,a,s')$,
$s'\sim P(\cdot\mid s,a)$, update one entry:
$Q_t(s,a)\leftarrow Q_t(s,a)+\alpha\bigl[\Kd a\bigl(\hat R(s,a)+\gamma\LSE_\beta(\{Q_{t+1}(s',a')\}_{a'})\bigr)-Q_t(s,a)\bigr]$.
If every reachable $(t,s,a)$ is updated infinitely often, each entry's step
sizes satisfy $\sum_n\alpha_n=\infty$ and $\sum_n\alpha_n^2<\infty$, and
$\hat R\to R$ almost surely, then $Q_t(s,a)\to Q^\beta_t(s,a)$ almost surely at
every reachable $(t,s,a)$.}
\end{theorem}

\rev{The proof (Appendix~\ref{app:proofs}) is backward induction on $t$: the
targets of step $t$ depend only on step $t{+}1$, which has already converged,
so it needs only continuity of $\LSE_\beta$ and
$\nrm{\Kd aX}\le\nrm X$, not the monotonicity or non-expansiveness that fail
(Props.~\ref{prop:lse_lipschitz_new}, \ref{prop:monotone_new}). The limit is
$Q^\beta$, not $Q^*$; greedy quality is governed by
Proposition~\ref{prop:nearopt}. E24 (App.~\ref{app:e24}) checks the theorem and
the step-size floor of E18--E23.}

\rev{\textbf{What is and is not proved.} (1) \emph{Evaluation}: operator TD
converges for a fixed policy (Thm~\ref{thm:td}). (2) \emph{Certificate}: the
exact finite-horizon recursion upper-bounds every policy's value
(Thm~\ref{thm:certificate}). (3) \emph{Control}: the sampled update converges
to that certificate for a finite horizon (Thm~\ref{thm:sampled}); for an
infinite horizon the exact fixed point certifies when $\T_\beta$ contracts, and
we prove no sampled convergence. (4) \emph{Greedy performance} is governed by
Proposition~\ref{prop:nearopt}, a sufficient condition. Appendix~\ref{app:scope}
tabulates these statements.}

\section{Properties of the Operator Log-Sum-Exp}\label{sec:theory}

Throughout, $x_i,y_i$ are self-adjoint with $\nrm{x_i},\nrm{y_i}\le M$.

\begin{proposition}[Well-posedness and Löwner properties]\label{prop:wellposed_new}
$\LSE_\beta$ is well defined and self-adjoint, and
(i) $\LSE_\beta(x)\succeq x_i$ for every $i$;
(ii) $\LSE_\beta(x_1+cI,\dots,x_K+cI)=\LSE_\beta(x)+cI$;
(iii) $\nrm{\LSE_\beta(x)}\le\max_i\nrm{x_i}+\beta^{-1}\log K$;
(iv) $U^*\LSE_\beta(x)U=\LSE_\beta(U^*x_1U,\dots,U^*x_KU)$ for unitary $U$;
(v) if all $x_i$ commute, $\LSE_\beta$ acts as the scalar log-sum-exp on the joint spectrum.
By Kadison's theorem \citep{kadison1951order} a pair $x_1,x_2$ has a
least upper bound in the Löwner order only if $x_1\preceq x_2$ or
$x_2\preceq x_1$; property (i) is therefore the natural replacement for
``$\max$'' in the non-commutative case.
\end{proposition}

\begin{proposition}[Lipschitz behaviour]\label{prop:lse_lipschitz_new}
Let $h_i=x_i-y_i$ and $\Delta=\LSE_\beta(x)-\LSE_\beta(y)$.
(a) \emph{Commuting sector} (all $x_i,y_j$ in one eigenbasis):
$\nrm{\Delta}\le\max_i\nrm{h_i}$ and
$\nrm{\Delta}_2\le\bigl(\sum_i\nrm{h_i}_2^2\bigr)^{1/2}$, for every
$\beta$; the first constant is attained (uniform shifts) and the second
is approached as $\beta\to\infty$.
(b) \emph{General}: for $p\in\{2,\infty\}$,
$\nrm{\Delta}_p\le e^{2\beta M}\,\tfrac1K\sum_i\nrm{h_i}_p$.
(c) \emph{No $\beta$-free constant for max-aggregated Hilbert--Schmidt
perturbations}: for $K=N$ and $x_i=e_ie_i^*$, perturbing each $x_i$ in
its own coordinate gives
$\nrm{\Delta}_2/\max_i\nrm{h_i}_2\to p_\beta\sqrt N$ with
$p_\beta=e^\beta/(e^\beta+N-1)$; any bound of the form
$\nrm{\Delta}_2\le C\max_i\nrm{h_i}_2$ therefore needs $C\ge p_\beta\sqrt N$.
\end{proposition}

\emph{Numerical observation} (E7, $80$-digit arithmetic). Outside the
commuting sector $\LSE_\beta$ is not non-expansive in operator norm:
local ratios $\nrm{\Delta}/\max_i\nrm{h_i}$ reach $0.97$, $1.17$ and
$1.48$ at $\beta=1,5,10$ \rev{(an observation, not a theorem)}.

Part (a) gives $\T_\beta$ modulus $\gamma$ on the commuting sector; (c) rules out a
$\beta$-linear constant under max-aggregation. Closing the gap between (b) and
the observed growth is open; no result depends on it, since
Theorem~\ref{thm:certificate} is proved by finite-horizon induction.

\begin{proposition}[Monotonicity]\label{prop:monotone_new}
On the commuting sector, $x_i\preceq y_i$ for all $i$ implies
$\LSE_\beta(x)\preceq\LSE_\beta(y)$, and hence $\T_\beta$ is monotone.
In general it is not: the matrix exponential is not operator monotone,
and E12 exhibits violations at every perturbation scale down to $10^{-4}$.
$\T^\pi$ is always monotone in CP, since $\Kd a$ is positive.
\end{proposition}

\rev{Actions are compared through the scalar readout $\tau(\rho Q(s,a))$, so the failure of
monotonicity affects the backup, not action selection; E22 measures how often
greedy actions switch late in training.}

\section{What Scalar Values Lose, and What Operators Retain}\label{sec:osMDP}

We now work in MP with $X_0=I$, terminal reward
$r(a_1,\dots,a_T)=\tfrac1N\Tr(WX_T)$, and a history $h=(a_1,\dots,a_{t})$
with history operator $X_h=A_{a_1}\cdots A_{a_t}$. An observation map
$\sigma$ on histories is \emph{order-blind} if $\sigma(h)=\sigma(h')$
whenever $h'$ is a permutation of $h$ (the step counter and the vector
of action counts are examples). Consider the last-step value
$q_T(h,a)=\tfrac1N\Tr(WX_hA_a)$ for $|h|=T-1$, which does not depend on
any continuation policy.

\begin{proposition}[Order-aliasing lower bound]\label{thm:representation_gap}
Let $T\ge3$ and let $\sigma$ be order-blind. For every scalar function
$f$ of $(\sigma(h),a)$,
\[
  \sup_{|h|=T-1,\,a}\bigl|f(\sigma(h),a)-q_T(h,a)\bigr|
  \;\ge\;\tfrac1{2N}\max_{a,b,c}\bigl|\Tr\bigl(W X_{g}[A_a,A_b]A_{c}\bigr)\bigr|,
\]
where $g$ is any fixed prefix of length $T-3$. If $[A_a,A_b]\ne0$ for
some $a,b$, the right-hand side is positive for Lebesgue-almost every
$W$. In contrast, $q_T(h,a)=\tau\bigl(X_h\,Q_T(a)\bigr)$ exactly, with
$Q_T(a)=\Kd a(W)=A_aW$ (Corollary~\ref{cor:product}).
\end{proposition}

The condition $T\ge3$ is necessary: for $T=2$ the histories $(a)$ and
$(b)$ are different multisets, an order-blind map may separate them, and
no aliasing occurs. The theorem is a statement about state aliasing, not
about the Markov property of the underlying process: with the latent
observed, the process is Markov and the bound does not apply.

The next result concerns \emph{exact tables} indexed by a history summary, not compact
approximators, which need no $K^T$ table (E13--E16, E18).

\begin{proposition}[Cost of exact tabular history representation: free versus nilpotent]\label{thm:horizon}
(a) If the $A_a$ generate a free group of rank $K$---for $SO(3)$ and
$SE(3)$ this holds for Lebesgue-almost every $K$-tuple, $K\ge2$
\citep{epstein1971almost}---then for almost every $W$ any exact scalar
table indexed by $(\sigma(h),a)$, for any map $\sigma$ of histories, must
separate all $K^{T-1}$ histories of length $T-1$, i.e.\ has at least
$K^T$ entries; in particular no order-blind $\sigma$ is exact.
(b) If $A_a=e^{\Omega_a}$ and the Lie algebra generated by the
$\Omega_a$ is nilpotent of step $2$, then $X_h$ depends on $h$ only
through the counts $n_a(h)$ and the signed pair counts
$L_{ab}(h)=\#\{i<j:h_i=a,h_j=b\}-\#\{i<j:h_i=b,h_j=a\}$, $a<b$; an exact
table therefore needs at most $K(T+1)^K(2T^2+1)^{K(K-1)/2}$ entries,
polynomial in $T$.
\end{proposition}

For step $m>2$ the analogous statistics are the iterated counts of
degree $<m$ (Chen's expansion, \citealp{chen1957integration}); we state
and prove only $m=2$, which covers the Heisenberg algebra
$\mathfrak n_3$ and bosonic displacements (E10, E15). For compact groups
the scalar deficit itself is bounded by $\nrm{W}$ uniformly in $T$; the
distinction between the two regimes is the \emph{size of an exact
history representation}, not a growing error.

\begin{proposition}[Operator sufficiency and minimal dimension]\label{thm:upper}
Let $f(w)=\tfrac1N\Tr(WA_w)$ on action words $w$, with Hankel matrix
$\mathbf H_f(u,v)=f(uv)$. (i) The $K$ operators $A_a$ and the readout
$W$ ($(K+1)N^2$ parameters) represent $f$ exactly for every horizon; by
Corollary~\ref{cor:product} this is the fixed point of~\eqref{eq:bellman}
along each plan. (ii) Every $d$-dimensional operator model
$g(w)=\tfrac1d\Tr(W_dQ_w)$, $Q_a\in\mathcal M_d$, has
$\operatorname{rank}\mathbf H_g\le d^2$; since
$\operatorname{rank}\mathbf H_f\le N^2$, exact representation requires
$d\ge(\operatorname{rank}\mathbf H_f)^{1/2}$, and $d=N$ suffices; that is,
the minimal exact dimension satisfies
$(\operatorname{rank}\mathbf H_f)^{1/2}\le d_{\min}\le N$, with
$d_{\min}=N$ whenever $\operatorname{rank}\mathbf H_f=N^2$.
(iii) For any finite prefix set $\mathcal U$ and suffix set $\mathcal V$,
every $d$-dimensional model has
$\max_{u,v}|g(uv)-f(uv)|\ge\bigl(\sum_{j>d^2}\sigma_j(\mathbf H_f[\mathcal U,\mathcal V])^2\bigr)^{1/2}/\sqrt{|\mathcal U||\mathcal V|}$.
\end{proposition}

Part (ii) is the Hankel-rank theory of weighted automata
\citep{carlyle1971realizations,fliess1974matrices}: a $d\times d$ operator model
is a weighted automaton of dimension at most $d^2$. The bound is tight when
$\operatorname{rank}\mathbf H_f=N^2$, as for generic orthogonal generators (E17),
so exact recovery occurs at $d=N$ (E8, E14, E16); Proposition~\ref{prop:structure}
refines it for unitary actions. An exact tabular history needs $K^T$ entries
(Prop.~\ref{thm:horizon}a); recurrent models need no table, and E13/E14 compare
them at matched parameter counts.

\paragraph{Extensions and practical cost.} Appendix~\ref{app:extensions}
proves three extensions (checked in E17b): greedy improvement on an
$\varepsilon$-accurate $\hat Q$ loses at most $2\varepsilon\nrm\rho_1/(1-\gamma\kappa)$
(Prop.~\ref{prop:robust}), so the TD guarantee carries over to one step of
policy improvement (OpQ-PI); Proposition~\ref{thm:exact_rep} and
Theorem~\ref{thm:certificate} hold for any positive maps, e.g.\ decohering
channels, and randomly perturbed actions (Prop.~\ref{prop:noisy}); and
Prop.~\ref{prop:structure} refines the operator dimension for structured action
algebras. One TD update costs $O(N^3)$, a soft update $O(KN^3)$, and memory is
$|\mathcal S|KN^2$ numbers, independent of the horizon (precision requirements:
Appendix~\ref{app:expdetails}). \rev{For $n$ qubits $N=2^n$, so both grow exponentially in $n$;
our experiments use $N\le4$.}

\section{Experiments}\label{sec:experiments}

We report three kinds of evidence. \textbf{(C) Control} (E18\rev{--E24}):
agents learn from sampled episodes, and their policies are evaluated exactly
against the optimal latent-aware value. \textbf{(A) Finite-instance checks of the
theory} (E17, E7, E9, E12; e.g.\ $0/30$ certificate violations). \textbf{(B) Representation} (E8, E10,
E13--E16): the ordered-product readout, the fixed point of~\eqref{eq:bellman} along
each plan (Corollary~\ref{cor:product}), is fitted to exact sequence returns
\rev{by supervised regression} (test MSE $\le3\times10^{-30}$ vs.\ $0.003$--$0.09$ for recurrent
models). (A) and (B) are in Appendix~\ref{app:prior}. Every number is produced by the released code
(\coderepo{}; App.~\ref{app:expdetails}).
\rev{Experiment IDs are fixed labels shared with the code and results, so they are
not consecutive (Table~\ref{tab:expsummary}).}

\subsection{E18: End-to-end control}\label{sec:e18}
\textbf{Setup.} We use three OLDPs in the conjugation picture, all with
observable state $s_t=(t,m_t)$:
\textbf{C1}, a random OLDP ($N{=}3$, $K{=}4$, two stochastic modes,
$H{=}6$, noisy rewards);
\textbf{C2}, $SO(3)$ pointing: six $\pm0.6$\,rad body rotations of a
marked axis $\rho=3nn^\top$, reward $(t_m\cdot n)^2$ for a mode-dependent
target, $H{=}8$;
\textbf{C3}, two-qubit state preparation with
$\{H_1,H_2,T_1,T_2,\mathrm{CNOT}\}$ and a terminal fidelity reward, $H{=}8$.
The initial latent is either fixed or drawn at random in every episode and
observed; a \emph{transfer} test trains with the fixed $\rho_0$ and evaluates
on $8$ unseen ones.
\textbf{Agents.}
\emph{OpQ-soft} learns one $N\times N$ operator per $(s,a)$ with the
sample form of~\eqref{eq:soft} ($\beta{=}5$), estimates $R(m,a)$ from scalar
rewards by least squares, and acts greedily on $\tau(\rho_tQ(s_t,a))$.
\emph{OpQ-PI} evaluates the uniform policy by operator
TD~\eqref{eq:update} and applies Prop.~\ref{prop:improve}.
Scalar baselines: $Q$-learning on $(t,m)$ (order-blind), $Q$-learning
on the full history (the $K^T$ table), a DQN with the latent $\rho_t$ as
input, and a GRU-based DRQN on the history.
The OpQ agents use the known dual maps $\Kd a$; DQN is given $\rho_t$ itself\rev{,
and E19 removes this difference}.
Scores are normalised so that $1$ is the exact dynamic-programming optimum
and $0$ is the uniform policy.

\begin{table}[t]
\centering\scriptsize\setlength{\tabcolsep}{3.2pt}
\begin{tabular}{lccc|ccc|ccc}
\toprule
& \multicolumn{3}{c|}{C1 random OLDP} & \multicolumn{3}{c|}{C2 $SO(3)$ pointing} & \multicolumn{3}{c}{C3 two-qubit prep.}\\
Agent & fixed & random & transfer & fixed & random & transfer & fixed & random & transfer\\
\midrule
OpQ-soft (learned $R$) & 0.84 & 0.95 & 0.95 & 0.98 & 0.99 & 0.99 & 0.83 & 0.68 & 0.68 \\
OpQ-soft (known $R$) & 0.84 & 0.95 & -- & 0.98 & 0.99 & -- & 0.79 & 0.68 & -- \\
OpQ-PI & 0.86 & 0.97 & 0.97 & 0.91 & 0.98 & 0.98 & 0.28 & 0.70 & 0.70 \\
Scalar $Q$ on $(t,m)$ & 0.87 & 0.51 & -- & 0.69 & 0.06 & -- & 0.19 & 0.10 & -- \\
Scalar $Q$, history table & 0.94 & 0.50 & -- & 0.75 & 0.05 & -- & 0.28 & -0.04 & -- \\
DQN on $\rho_t$ & 1.00 & 0.99 & 0.87 & 0.99 & 0.99 & 0.96 & 1.00 & 0.87 & 0.49 \\
DRQN (GRU) on history & 1.00 & 0.95 & 0.53 & 0.97 & 0.94 & 0.21 & 0.96 & -0.01 & 0.11 \\
\bottomrule
\end{tabular}
\caption{E18: normalised return of the greedy policy after $3000$ episodes
(mean over $5$ seeds; standard deviations, scores after $100$ episodes and
learning curves are in Appendix~\ref{app:e18full}). \emph{Transfer}:
trained with a fixed $\rho_0$, evaluated on $8$ unseen initial latents.}
\label{tab:e18}
\end{table}

\textbf{Results} (Table~\ref{tab:e18}).
\emph{(i) Order-blind scalar agents fail once the latent
varies} (random $\rho_0$: $0.51$, $0.06$, $0.10$ for $Q$ on $(t,m)$), as
Proposition~\ref{thm:representation_gap} predicts. \rev{The history table fails
there too: its key omits the random $\rho_0$.} OpQ reaches $0.95$--$0.99$ on
C1 and C2 while storing $432$ and $864$ numbers, compared
with $K^H=4096$ and $1.7\times10^6$ histories.
\emph{(ii) \rev{Against baselines that do not know $\mathcal K_a$,} OpQ learns fastest.} After $100$ episodes
with a random $\rho_0$, OpQ-soft scores $0.95/0.95$ on C1/C2, against
$0.89/0.73$ for DQN and $0.44/0.28$ for DRQN. \rev{E19 shows how much of this is
knowledge of $\mathcal K_a$.}
\emph{(iii) OpQ transfers zero-shot}, since Proposition~\ref{thm:exact_rep} makes the
readout valid at every $\rho$: trained on one latent, it keeps
$0.95$--$0.99$ on C1/C2 and $0.68$ on C3 at unseen latents, against
$0.87/0.96/0.49$ for DQN\rev{; DRQN's transfer column is uninformative
(constant $\rho_0$ in training)}.
\emph{(iv) With enough episodes, a DQN that observes $\rho_t$ matches or
exceeds OpQ} (C1 fixed: $1.00$ vs.\ $0.84$; C3: $1.00/0.87$ vs.\
$0.83/0.68$). These are the ceilings of the exact-model readout ($0.86$, $0.79$,
$0.68$; Table~\ref{tab:e18ref}; \rev{C3 fixed with learned $R$, $0.83\pm0.08$,
is within one s.d.}), the limit of Theorem~\ref{thm:sampled}, so the gap is
compression into one operator per $(s,a)$, not optimisation error; on C2 OpQ
is within $0.02$ of the optimum.

\rev{\subsection{E19: Controlling for knowledge of the operators}\label{sec:e19}}
\begin{table}[t]
\centering\scriptsize\setlength{\tabcolsep}{3.2pt}
\begin{tabular}{llcccccc}
\toprule
& & \multicolumn{2}{c}{C1 random OLDP} & \multicolumn{2}{c}{C2 $SO(3)$ pointing} & \multicolumn{2}{c}{C3 two-qubit}\\
Agent & $\mathcal K_a$ & 100 & 3000 & 100 & 3000 & 100 & 3000\\
\midrule
DQN on $\rho_t$ (nonlinear) & unknown & 0.89 & \textbf{0.99} & 0.73 & \textbf{0.99} & 0.04 & 0.87 \\
DQN $+\mathcal K_a$ data (nonlinear) & known & 0.93 & \textbf{0.99} & 0.82 & \textbf{0.99} & 0.22 & 0.86 \\
Linear-Q (linear) & unknown & 0.77 & 0.96 & 0.57 & 0.95 & 0.25 & 0.54 \\
OpQ-soft (linear) & known & \textbf{0.95} & 0.95 & \textbf{0.95} & \textbf{0.99} & \textbf{0.58} & 0.68 \\
\midrule
\upd{Linear-Q $+\mathcal K_a$ data (linear)} & \upd{known} & \upd{0.96} & \upd{0.97} & \upd{0.91} & \upd{0.97} & \upd{0.41} & \upd{0.69} \\
DQN $+\mathcal K_a$ data, 4 updates/step & known & 0.96 & 1.00 & 0.90 & 1.00 & 0.38 & 0.90 \\
Learned model $+$ exhaustive search & known & 1.00 & 1.00 & 1.00 & 1.00 & 1.00 & 1.00 \\
\bottomrule
\end{tabular}
\caption{E19, $2\times2$ comparison with random initial latents: normalised return after $100$ and $3000$ episodes (mean over $5$ seeds; standard deviations in Table~\ref{tab:e19-2x2-random}). Bold marks the best of the first four rows. \upd{Linear-Q $+\mathcal K_a$ data is a control.}}
\label{tab:e19}
\end{table}

\rev{E19 crosses the form of the value function (nonlinear or linear in $\rho$)
with knowledge of $\mathcal K_a$: DQN; DQN $+\mathcal K_a$ data, with half of
each minibatch generated by $\mathcal K_a$ from fresh pure latents and OpQ's
$\hat R$; Linear-Q, one linear function of $\rho$ per $(t,m,a)$ with OpQ's soft
backup but no operators; \upd{Linear-Q $+\mathcal K_a$ data, the linear class on $\mathcal K_a$-generated transitions;} and OpQ-soft. Planners know $\mathcal K_a$, learn
$\hat R,\hat P$, and run exhaustive search or UCT-100/1000
\citep{kocsis2006bandit} (Appendix~\ref{app:e19}).}

\rev{\textbf{Results} (Table~\ref{tab:e19}). (i) Linear structure without the
operators (Linear-Q) does not reproduce OpQ.
\upd{With $\mathcal K_a$-generated transitions the linear class reaches $0.96$ and $0.91$ at $100$
episodes on C1 and C2 (OpQ-soft $0.95$, $0.95$), so most of OpQ's early advantage there is
knowledge of $\mathcal K_a$; on C3 a gap remains ($0.41$ vs.\ $0.58$).} (ii) Much of the early advantage
in E18 is knowledge of $\mathcal K_a$: $\mathcal K_a$-generated transitions
raise DQN at 100 episodes, but OpQ-soft stays ahead on C2 and C3. The
DQN$+\mathcal K_a$ arms also receive synthetic latents (and four times the
gradient steps); all arms use the same environment episodes. (iii) At short
horizons a planner with the same knowledge is more sample-efficient than every
learner ($0.99/0.99/0.83$ after $25$ episodes). (iv) Held-out: OpQ-soft
$0.95/0.99/0.68$, DQN $0.86/0.96/0.46$, DQN $+\mathcal K_a$ data
$0.99/0.98/0.84$. (v) On C2 for $H=8$--$20$, OpQ-soft ($0.98$--$0.99$) and
UCT-1000 ($0.96$--$0.98$) are at parity, at $0.035$ vs.\ $250$--$417$\,ms per
decision; on C1 OpQ ($0.81$--$0.85$) stays below UCT-1000 ($0.94$--$0.98$). The
certificate stays valid at every horizon; the advantage concerns decision cost,
not training (Appendix~\ref{app:e19}).}

\rev{\subsection{E20: Tightening the single-operator certificate}\label{sec:e20}}
\rev{E20 uses the exact model and asks what closes the gap of E18(iv)
(Appendix~\ref{app:e20}). (a) The trace-minimal SDP upper bound of
Corollary~\ref{cor:anybound} tightens the certificate on C1 ($1.88\to1.67$
times $V^*$) but does not improve greedy control
(Proposition~\ref{prop:nearopt}). (b) $M$ operators per $(t,m)$ with point-based
backups \citep{pineau2003point} close the gap: C1 $0.86\to1.00$ at $M{=}32$, C3
$0.79\to1.00$ at $M{=}8$ ($0.95$ held-out at $M{=}64$); on C2 one LSE operator
($0.99$) beats sets up to $M{=}64$. This family is point-based value iteration
in operator form; the certificate adds an upper bracket for choosing $M$
(App.~\ref{app:scope}). (c) \emph{Miscalibrated operators} (E21,
Appendix~\ref{app:e21}). With $\tilde U_a=U_ae^{\varepsilon A_a}$ in the agent's
model, OpQ on SO(3) is as robust as planning ($\ge0.95$ up to $\varepsilon=0.1$),
but on C1 and C3 its near-tied greedy decisions at the training latent are less
robust (C3: $0.79\to0.49$ at $\varepsilon=0.03$, planning $1.00$).
(d) \emph{Hidden latent} (E23, App.~\ref{app:e23}). With $\rho_0$ drawn from a
prior and never observed, agents track the belief mean $\bar\rho_t$; OpQ trained
with the latent is unchanged on $\bar\rho_t$, since the plug-in is exact for
linear values, while a DQN drops to $0.85/0.45$ on C1/C3. With the reward
observable known OpQ learns fastest on C2/C3; a belief-mean DQN ends higher on
C3 ($1.00$ vs.\ $0.79$).}

\section{Conclusion}\label{sec:conclude}

We introduced operator-valued $Q$-learning for decision processes whose
actions compose non-commutatively: exact latent-blind evaluation, convergent
operator TD for policy evaluation, and \rev{a Löwner certificate that replaces
maximum-based bounds when values form no lattice, to which the sampled control
update converges for a finite horizon. The learned policy evaluates unseen
latents without retraining, matches 1000-simulation tree search on SO(3)
pointing up to $H{=}20$ at about $10^4$ times lower decision cost, and a few
operators per state close its gap. With a hidden initial state it acts exactly
on the belief mean. E19--E23 delimit the claims: much of the early advantage
over model-free baselines is knowledge of the operators; short-horizon planning
needs fewer episodes; and near-tied greedy decisions make the single operator
sensitive to miscalibrated gates. Learning unknown or uncertain operators
\citep{balle2014spectral} and scaling to large $N$ are open.}

\newpage
\clearpage

\subsubsection*{Reproducibility statement}
Assumptions and proofs are in Appendices~\ref{app:proofs}, \ref{app:lipschitz}, \ref{app:horizon} and~\ref{app:extensions}; the full experimental protocols, hyperparameters and seeds are in Appendix~\ref{app:expdetails}. The anonymized repository at \coderepo{} contains the code for every experiment, the raw results, and the scripts that produce every table and figure; the E7--E18 suite runs on one CPU core in about three hours\rev{; E19--E24 (Appendices~\ref{app:e19}, \ref{app:e20}, \ref{app:e21}, \ref{app:e22}, \ref{app:e23}, \ref{app:e24}) take about \upd{seven} CPU-hours for E19 and 100 minutes on two cores for E20--E24, all on CPU}.

\subsubsection*{Ethics statement}
This is a methods paper on reinforcement learning with operator-valued value functions, and we see no direct ethical concerns. Applications to quantum or robotic control should follow the safety practice of those domains.

\subsubsection*{AI use statement}
We used an AI assistant for language editing, restructuring the manuscript, literature search, and checking theorem statements, assumptions, and experimental design. All outcomes from the AI assistant were independently reviewed and verified by the authors. The authors take full
responsibility for the content, methodology, code, and reported results of the paper.

\bibliography{refs}
\bibliographystyle{plainnat}

\newpage
\appendix

\begin{center}
{\LARGE\bfseries Appendix}
\end{center}
\vspace{0.5em}

\noindent This appendix contains the full proofs of all theoretical
results, further theoretical extensions, and the complete experimental
results and protocols.
\begin{itemize}
\item Appendix~\ref{app:warmup} works through a fully explicit
low-dimensional example (paper folding).
\item Appendices~\ref{app:proofs}, \ref{app:lipschitz}
and~\ref{app:horizon} prove the results of Sections~\ref{sec:opq},
\ref{sec:theory} and~\ref{sec:osMDP}, respectively.
\item Appendix~\ref{app:extensions} states and proves the extensions:
robust improvement, noisy and random latent dynamics, and a
structure-refined operator dimension.
\item Appendix~\ref{app:expsummary} summarises all experiments.
\item \rev{Appendices~\ref{app:e19} and~\ref{app:e20} give the full results of the
model-matched comparison (E19) and of the tightening study (E20).}
\item \rev{Appendix~\ref{app:scope} tabulates what is proved and under which conditions, and
discusses when one operator suffices, how to choose $M$, and its cost;
Appendix~\ref{app:e21} tests miscalibrated operators (E21), Appendix~\ref{app:e22} the
sensitivity of the sampled update to $\beta$ (E22), Appendix~\ref{app:e23} control with a
hidden initial latent (E23), and Appendix~\ref{app:e24} the convergence of the sampled
update (E24).}
\item Appendix~\ref{app:prior} gives the full results.
\item Appendix~\ref{app:expdetails} gives the experimental protocols,
hyperparameters and computational requirements.
\end{itemize}

\section{A Worked Example: Folding Paper}\label{app:warmup}
\begin{figure}[h]
  \centering
  \includegraphics[width=\linewidth]{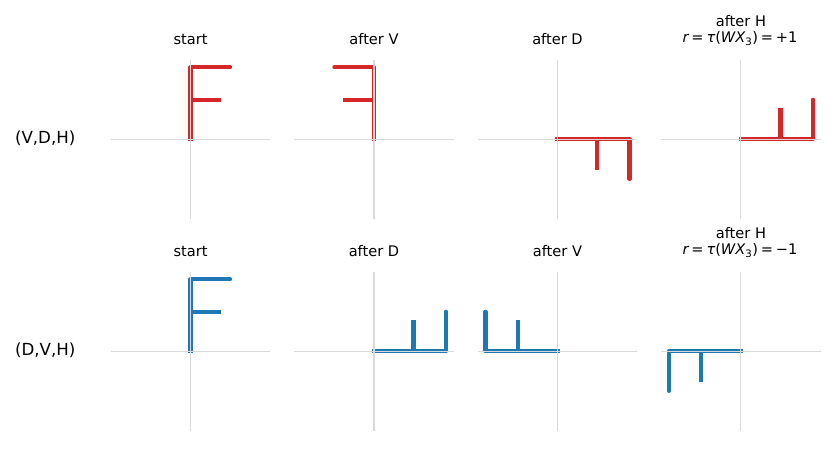}
  \caption{The plans $(V,D,H)$ and $(D,V,H)$ applied to a marked flap (drawn as the
  letter F). After two folds both histories contain one $V$ and one $D$, and the third
  action is the same, yet the readouts are $+1$ and $-1$. An order-blind scalar $Q$ must
  err by at least $1$ on one of them; the operator recursion gives both values exactly.}
  \label{fig:origami}
\end{figure}

\paragraph{A warm-up: folding paper.}
Track one marked flap of a square sheet, and idealise a fold along a crease
through the centre with unit normal $n$ as the reflection it applies to that
flap (layer and collision constraints ignored, as in rigid origami):
\begin{equation*}
  A_n \;=\; I-2nn^{\top} \;=\; e^{\,i\pi nn^{\top}}\in\mathbb{C}^{2\times2}.
\end{equation*}
Take the vertical, horizontal and diagonal creases,
\begin{equation*}
  A_V=\begin{pmatrix}-1&0\\0&1\end{pmatrix},\quad
  A_H=\begin{pmatrix}1&0\\0&-1\end{pmatrix},\quad
  A_D=\begin{pmatrix}0&1\\1&0\end{pmatrix}.
\end{equation*}
A plan $a_1,\dots,a_T$ moves the flap by the ordered product
$X_T=A_{a_1}\cdots A_{a_T}$ (acting on row vectors, $x^{\top}\mapsto x^{\top}X_T$).
This is the multiplicative picture of Definition~\ref{def:oldp} with $X_0=I$.

\smallskip\noindent\textit{Where order enters.}
$A_V$ and $A_H$ are simultaneously diagonal, $[A_V,A_H]=0$: folding in half
twice gives $-I$ in either order. The diagonal crease breaks this:
\begin{equation*}
  A_VA_D=\begin{pmatrix}0&-1\\1&0\end{pmatrix}=R_{90^\circ},\qquad
  A_DA_V=\begin{pmatrix}0&1\\-1&0\end{pmatrix}=R_{90^\circ}^{\top},\qquad
  [A_V,A_D]=\begin{pmatrix}0&-2\\2&0\end{pmatrix}\neq0.
\end{equation*}
Two reflections compose to a rotation by twice the angle between the creases,
with sign fixed by the order.

\smallskip\noindent\textit{Why a scalar value on an order-blind observation misses it.}
With two folds, a scalar $Q$ on the observation ``one fold made'' is \emph{not}
aliased: after $V$ the next action is $D$ and after $D$ it is $V$, so the two
plans query different entries $Q(1,D)$ and $Q(1,V)$. Aliasing requires a third
step (Proposition~\ref{thm:representation_gap}). Consider the plans $(V,D,H)$ and
$(D,V,H)$ with $W=(A_VA_DA_H)^{\top}$ and $N=2$, so that
$r=\tfrac1N\Tr(WX_3)$ is the cosine of the orientation error. Then
$r(V,D,H)=\tau(W\,R_{90^\circ}A_H)=1$ and
$r(D,V,H)=\tau\bigl(A_HR_{90^\circ}^{\top}R_{90^\circ}^{\top}A_H\bigr)=\tau(-I)=-1$.
Before the third fold both histories contain one $V$ and one $D$, so
any observation that records which folds were made (fold count, fold
counts per crease) is identical, the current action $H$ is identical,
and every scalar $Q$ on that observation errs by at least
$\tfrac12|1-(-1)|=1$ on one of them. The unweighted trace is blind as well,
$\Tr(A_VA_DA_H)=\Tr(A_DA_VA_H)=0$; order becomes visible only through the readout.

\smallskip\noindent\textit{What the operator recursion does.}
The last-step operator is $Q_2(H)=\Kd H(W)=A_HW$, and the recursion
\eqref{eq:bellman} along the plan gives $Q_0(a_1)=A_{a_1}A_{a_2}A_{a_3}W$
(Corollary~\ref{cor:product}). After either history $h$ the value of
folding $H$ is $\tau(X_h\,A_HW)$, with the history operator $X_h\in\{R_{90^\circ},R_{90^\circ}^\top\}$:
exactly $+1$ and $-1$. The model stores $K=3$ operators and a readout,
$(K+1)N^2=16$ numbers, for every horizon. In the commuting sector $\{V,H\}$
the operators share an eigenbasis and order is irrelevant.

\section{Proofs for Section~\ref{sec:opq}}\label{app:proofs}

Throughout, $\nrm{\cdot}_1$, $\nrm\cdot_2$, $\nrm\cdot$ are the normalised
trace, Hilbert--Schmidt and operator norms, and
$|\tau(Y\rho)|\le\nrm{Y}\,\nrm{\rho}_1$ (Hölder). Since the trace pairing
makes $(\mathcal M_N,\nrm\cdot_1)$ and $(\mathcal M_N,\nrm\cdot)$ dual,
$\nrm{\mathcal K_a}_{1\to1}=\nrm{\Kd a}_{\infty\to\infty}\le\kappa$.

\subsection{Proof of Proposition~\ref{thm:exact_rep} and Corollary~\ref{cor:product}}
\begin{proof}
\emph{Finite horizon.} Let $q^\pi_t$ be the expected discounted reward
collected from step $t$ to $H-1$. We show
$q^\pi_t(s,\rho,a)=\tau(\rho\,Q^\pi_t(s,a))$ by backward induction; the
case $t=H$ is $0=0$. Since $P$ does not depend on the latent and
$\pi(\cdot\mid s')$ does not depend on it either, the Markov property of
$(s_t,\rho_t)$ gives
\[
q^\pi_t(s,\rho,a)=\tau\bigl(R(s,a)\mathcal K_a(\rho)\bigr)+\gamma\sum_{s'}P(s'\mid s,a)\sum_{a'}\pi(a'\mid s')\,q^\pi_{t+1}\bigl(s',\mathcal K_a(\rho),a'\bigr).
\]
By the induction hypothesis the second term is
$\gamma\,\tau\bigl(\mathcal K_a(\rho)\,Y\bigr)$ with
$Y=\sum_{s'}P(s'\mid s,a)\sum_{a'}\pi(a'\mid s')Q^\pi_{t+1}(s',a')$. Using
the defining identity of the dual map and its linearity,
$q^\pi_t(s,\rho,a)=\tau\bigl(\rho\,\Kd a(R(s,a)+\gamma Y)\bigr)=\tau(\rho\,Q^\pi_t(s,a))$.

\emph{Infinite horizon.} By Theorem~\ref{thm:td}(i), $\T^\pi$ is a
$\gamma\kappa$-contraction, so $Q_n:=(\T^\pi)^n0\to Q^\pi$, the unique
fixed point. $Q_n$ is the time-$0$ operator of the $n$-step problem, so
by the finite-horizon case the $n$-step truncated return equals
$\tau(\rho Q_n(s,a))$. The tail is controlled by
$|r_t|\le R_{\max}\nrm{\rho_{t+1}}_1\le R_{\max}\kappa^{t+1}\nrm{\rho}_1$,
which is summable against $\gamma^t$ when $\gamma\kappa<1$. Letting
$n\to\infty$ gives $q^\pi(s,\rho,a)=\tau(\rho Q^\pi(s,a))$.

\emph{Corollary.} With $s_t=t$, a deterministic plan, $R=0$ before the
last step and $R=W$ at it, the recursion reads $Q_{T-1}(a_T)=\Kd{a_T}(W)=A_{a_T}W$ and
$Q_t(a_{t+1})=\Kd{a_{t+1}}(Q_{t+1}(a_{t+2}))=A_{a_{t+1}}Q_{t+1}(a_{t+2})$,
whence $Q_t(a_{t+1})=A_{a_{t+1}}\cdots A_{a_T}W$ and, from $X_0=I$,
$\tau(Q_0(a_1))=\tfrac1N\Tr(A_{a_1}\cdots A_{a_T}W)=\tfrac1N\Tr(WA_{a_1}\cdots A_{a_T})$
by cyclicity. After $h=(a_1,\dots,a_t)$ the latent is $X_h$, and the
finite-horizon case gives the value $\tau(X_hQ_t(a_{t+1}))=\tfrac1N\Tr(WA_{a_1}\cdots A_{a_T})$.
\end{proof}

\subsection{Proof of Theorem~\ref{thm:td}}
\begin{proof}
(i) Rewards cancel and
$\nrm{(\T^\pi Q_1-\T^\pi Q_2)(s,a)}=\gamma\nrm{\Kd a(\sum_{s'}P\sum_{a'}\pi\,(Q_1-Q_2)(s',a'))}\le\gamma\kappa\nrm{Q_1-Q_2}_{\infty,\mathrm{op}}$,
since $\sum_{s'}P\sum_{a'}\pi$ is a convex combination. The same
computation in $\nrm\cdot_2$ applies when $\nrm{\Kd a(Y)}_2\le\nrm Y_2$,
which holds for $Y\mapsto U^*YU$ and $Y\mapsto AY$ with $A$ unitary or
orthogonal. A consequence is the stability bound
$\nrm{Q-Q^\pi}_{\infty,\mathrm{op}}\le(1-\gamma\kappa)^{-1}\nrm{\T^\pi Q-Q}_{\infty,\mathrm{op}}$.

(ii) Identify $Q$ with a vector in $\mathbb C^{|\mathcal S||\mathcal A|N^2}$
and write $e_{(s,a)}$ for the block embedding. The update is
$Q_{k+1}=Q_k+\alpha_k\bigl(F(Q_k)+M_{k+1}\bigr)$ with
$F(Q)=D_\mu(\T^\pi Q-Q)$, where $D_\mu$ multiplies block $(s,a)$ by
$\mu(s,a)>0$, and
$M_{k+1}=e_{(s_k,a_k)}\bigl[\Kd{a_k}(R+\gamma Q_k(s'_k,a'_k))-Q_k(s_k,a_k)\bigr]-F(Q_k)$.
Given the past, $(s_k,a_k,s'_k,a'_k)$ is drawn afresh, so $M_{k+1}$ is a
martingale difference with
$\mathbb E[\nrm{M_{k+1}}^2\mid\mathcal F_k]\le C(1+\nrm{Q_k}^2)$. $F$ is
affine, $F(Q)=GQ+g$ with $G=D_\mu(\gamma\mathcal P^\pi-I)$ and
$(\mathcal P^\pi Q)(s,a)=\Kd a(\sum_{s'}P\sum_{a'}\pi\,Q(s',a'))$. For
$0<h\le1/\max\mu$ and every block,
$\nrm{((I+hG)Q)(s,a)}\le(1-h\mu(s,a))\nrm{Q(s,a)}+h\mu(s,a)\gamma\kappa\nrm{Q}_{\infty,\mathrm{op}}\le\bigl(1-h\mu_{\min}(1-\gamma\kappa)\bigr)\nrm Q_{\infty,\mathrm{op}}$,
so the logarithmic norm of $G$ in $\nrm\cdot_{\infty,\mathrm{op}}$ is at
most $-\mu_{\min}(1-\gamma\kappa)<0$. Every eigenvalue of $G$ therefore
has negative real part, and both $\dot Q=F(Q)$ and the scaled limit
$\dot Q=GQ$ have globally asymptotically stable equilibria ($Q^\pi$ and
$0$ respectively; $D_\mu$ is invertible, so $F(Q)=0$ iff $\T^\pi Q=Q$).
The conditions of the ODE method of \citet{borkar2000ode} hold, giving
$\sup_k\nrm{Q_k}<\infty$ and $Q_k\to Q^\pi$ almost surely.
\end{proof}

\subsection{Proof of Proposition~\ref{prop:improve}}
\begin{proof}
A latent-blind policy is a latent-aware policy that ignores $\rho$, and
by Proposition~\ref{thm:exact_rep} its action-value function in the MDP on
$(s,\rho)$ is $\tau(\rho Q^\pi(s,a))$. Hence
$V^\pi(s,\rho)=\sum_a\pi(a\mid s)\tau(\rho Q^\pi(s,a))\le\tau(\rho Q^\pi(s,\pi^+(s,\rho)))$.
Unrolling as in the classical policy-improvement theorem
\citep[\S4.2]{sutton2018rl}, $V^\pi(s_0,\rho_0)\le\mathbb
E^{\pi^+}[\sum_{t<n}\gamma^tr_t]+\gamma^n\mathbb E^{\pi^+}[V^\pi(s_n,\rho_n)]$.
Since $|V^\pi(s,\rho)|\le\nrm{Q^\pi}_{\infty,\mathrm{op}}\nrm\rho_1$ and
$\nrm{\rho_n}_1\le\kappa^n\nrm{\rho_0}_1$, the last term vanishes when
$\gamma\kappa<1$, and the first converges to $V^{\pi^+}(s_0,\rho_0)$.
\end{proof}

\subsection{Proof of Theorem~\ref{thm:certificate}}
\begin{proof}
In CP each $\mathcal K_a$ and $\Kd a$ is positive, so $Q^\beta_t$ is
self-adjoint for all $t$ and $\mathcal K_a(\rho)\succeq0$ whenever
$\rho\succeq0$. For an arbitrary (possibly history-dependent,
latent-aware) policy, let $q^\pi_t$ denote its action value from step
$t$; the continuation of $\pi$ after one more step is again a policy.
We prove by backward induction that
$q^\pi_t(s,\rho,a)\le\tau(\rho Q^\beta_t(s,a))$ for all policies, $s$,
$a$ and $\rho\succeq0$. At $t=H-1$ both sides equal
$\tau(R(s,a)\mathcal K_a(\rho))=\tau(\rho\,\Kd a(R(s,a)))$. For $t<H-1$,
writing $\rho'=\mathcal K_a(\rho)\succeq0$ and $\bar\pi(\cdot)$ for the
(history-dependent) distribution of the next action,
\begin{align*}
q^\pi_t(s,\rho,a)
&=\tau(R(s,a)\rho')+\gamma\sum_{s'}P(s'\mid s,a)\sum_{a'}\bar\pi(a')\,q^{\pi'}_{t+1}(s',\rho',a')\\
&\le\tau(R(s,a)\rho')+\gamma\sum_{s'}P(s'\mid s,a)\sum_{a'}\bar\pi(a')\,\tau\bigl(\rho'\,Q^\beta_{t+1}(s',a')\bigr)\\
&\le\tau(R(s,a)\rho')+\gamma\sum_{s'}P(s'\mid s,a)\,\tau\bigl(\rho'\,V^\beta_{t+1}(s')\bigr)
=\tau(\rho\,Q^\beta_t(s,a)).
\end{align*}
The first inequality is the induction hypothesis at $\rho'\succeq0$; the
second uses $Q^\beta_{t+1}(s',a')\preceq V^\beta_{t+1}(s')$
(Prop.~\ref{prop:wellposed_new}(i)) and that $Y\mapsto\tau(\rho'Y)$ is
monotone for $\rho'\succeq0$; the last equality is the dual identity.
Taking the supremum over policies gives the statement for $q^*_t$. For
latent-blind $\pi$, Proposition~\ref{thm:exact_rep} gives
$\tau(\rho(Q^\beta_t-Q^\pi_t))\ge0$ for every $\rho\succeq0$, i.e.\
$Q^\pi_t\preceq Q^\beta_t$.

\emph{Infinite horizon.} Unital positive maps have operator norm $1$
(Russo--Dye; \citealp[Ch.~2]{bhatia2007positive}), so by
Prop.~\ref{prop:wellposed_new}(iii)
$\nrm{\T_\beta Q}_{\infty,\mathrm{op}}\le R_{\max}+\gamma(\nrm Q_{\infty,\mathrm{op}}+\beta^{-1}\log K)$
and the ball of radius $M_\beta$ is invariant. On it $\LSE_\beta$ is
Lipschitz with constant $1$ on the commuting sector (which $\T_\beta$
preserves when all $U_a,R(s,a)$ lie in one maximal abelian subalgebra)
and $e^{2\beta M_\beta}$ in general (Prop.~\ref{prop:lse_lipschitz_new}(a,b)),
giving the stated moduli.

\emph{Truncation bound (no contraction needed).} Fix $n$. The
finite-horizon argument above, applied with horizon $n$, bounds the
$n$-step truncated return of every policy by
$\tau(\rho Q^{\beta,(n)}_0(s,a))$. In CP, $\rho_t\succeq0$ and
$\tau(\rho_t)=\tau(\rho)$, so $|r_t|\le R_{\max}\tau(\rho)$ and the
remaining rewards contribute at most
$\sum_{t\ge n}\gamma^tR_{\max}\tau(\rho)=\gamma^nR_{\max}\tau(\rho)/(1-\gamma)$.
This proves the truncation bound. When $\T_\beta$ contracts,
$Q^{\beta,(n)}_0\to Q^\beta$ as $n\to\infty$ and the tail vanishes, so the
fixed point is a certificate.
\end{proof}

\subsection{Proof of Proposition~\ref{prop:reduction}}
\begin{proof}
(i) For $Y\in\mathcal D$ and $U_a\in\mathcal D$ unitary,
$U_a^*YU_a=U_a^*U_aY=Y$, so $\Kd a$ is the identity on $\mathcal D$;
functional calculus acts entrywise on diagonal operators, so
$\LSE_\beta$ maps $\mathcal D^K$ to $\mathcal D$ coordinatewise, and
coordinate $j$ of $\T_\beta$ is the classical soft Bellman operator with
reward $r_j(s,a)=R(s,a)_{jj}$.
(ii) For $\rho=Ne_je_j^*$, $\mathcal K_a(\rho)=U_a\rho U_a^*=\rho$, so the
latent is constant and every policy faces the classical MDP with reward
$r_j$; hence $q^*_t(s,\rho,a)$ is its optimal value $q^*_{j,t}(s,a)$, while
$\tau(\rho Q^\beta_t(s,a))=(Q^\beta_t(s,a))_{jj}$ is its soft-optimal
value. With $\delta_t=\max_{s,a}(q^\beta_{j,t}-q^*_{j,t})\ge0$ we have
$\delta_{H-1}=0$ and, since $\mathrm{lse}_\beta(u)\le\max u+\beta^{-1}\log K$,
$\delta_t\le\gamma(\beta^{-1}\log K+\delta_{t+1})$, which gives the bound.
(iii) $N=1$ gives $\mathcal M_1=\mathbb C$, $\Kd a=\mathrm{id}$ and
$\LSE_\beta=\mathrm{lse}_\beta$.
\end{proof}

\subsection{\rev{Proof of Corollary~\ref{cor:anybound}}}
\begin{proof}
\rev{The proof of Theorem~\ref{thm:certificate} uses two properties of the
aggregator: that its output is self-adjoint, so $Q^\beta_t$ stays self-adjoint
under the positive maps $\Kd a$, and that it dominates each argument,
$V^\beta_{t+1}(s')\succeq Q^\beta_{t+1}(s',a')$, in the second inequality of the
induction. Both hold for $G$ by assumption, and no other property of $\LSE_\beta$
(monotonicity, smoothness, Lipschitz continuity) is used. The finite-horizon
statements, including $Q^\pi_t\preceq Q^G_t$ for latent-blind $\pi$, follow
verbatim. The trace-minimal $G$ exists because the feasible set
$\{Y:Y\succeq x_i\ \forall i\}$ is closed, convex and contains
$(\max_i\nrm{x_i})I$, and $\tau$ is bounded below on it by $\max_i\tau(x_i)$. It
is minimal in the Löwner order: if $Y'\preceq Y$ is also feasible then
$\tau(Y-Y')\ge0$ with equality only if $Y'=Y$, since $\tau$ is faithful.}
\end{proof}

\subsection{\rev{Proof of Proposition~\ref{prop:nearopt}}}
\begin{proof}
\rev{For latent-aware control the process $x_t=(s_t,\rho_t)$ is a finite-horizon
MDP, and for any deterministic policy $\pi$ the performance-difference identity,
obtained by telescoping
$V^*(x_t)=q^*(x_t,\pi(x_t))+[V^*(x_t)-q^*(x_t,\pi(x_t))]$ along trajectories of
$\pi$, gives
\[
  V^*(x_0)-V^\pi(x_0)=\sum_{t=0}^{H-1}\gamma^t\,\mathbb E^\pi\bigl[V^*(x_t)-q^*_t(x_t,\pi(x_t))\bigr].
\]
Fix $x=(s,\rho)$ at step $t$, let $a^*$ be optimal there and $\hat a=\hat\pi(x)$
the greedy readout. Since $\tau(\rho\hat Q_t(s,\hat a))\ge\tau(\rho\hat Q_t(s,a^*))$,
\[
  q^*_t(x,\hat a)=\tau(\rho\hat Q_t(s,\hat a))-\delta_t(x,\hat a)\ge\tau(\rho\hat Q_t(s,a^*))-\delta_t(x,\hat a)=q^*_t(x,a^*)+\delta_t(x,a^*)-\delta_t(x,\hat a),
\]
so $V^*_t(x)-q^*_t(x,\hat a)\le\max_a\delta_t(x,a)-\min_a\delta_t(x,a)$. Taking
expectations under $\hat\pi$ and summing gives the bound. In the commuting sector
at a pure eigenstate $\rho=Ne_je_j^*$, the latent is constant and
Prop.~\ref{prop:reduction}(ii) gives
$0\le\delta_t\le\beta^{-1}\log K\sum_{k=1}^{H-t-1}\gamma^k$ for every action,
which bounds the spread.}
\end{proof}

\subsection{\rev{Proof of Theorem~\ref{thm:sampled}}}
\rev{We use the Hilbert--Schmidt inner product $\langle X,Y\rangle=\Re\Tr(X^*Y)$ on
self-adjoint operators, with norm $\nrm\cdot_F$; all norms on this
finite-dimensional space are equivalent. Two facts are used: in CP, $\Kd a$ is
positive and unital, so $\nrm{\Kd aX}\le\nrm X$ in operator norm (Russo--Dye;
\citealp[Ch.~2]{bhatia2007positive}); and $\LSE_\beta$ is continuous, indeed
locally Lipschitz (Prop.~\ref{prop:lse_lipschitz_new}).}

\begin{lemma}[Stochastic approximation with vanishing bias]\label{lem:sa}
\rev{Let $(\mathcal F_n)$ be a filtration and
$x_{n+1}=x_n+\alpha_n(c-x_n+w_n+b_n)$, where $c$ is fixed,
$\alpha_n\in[0,1]$ are deterministic with $\sum_n\alpha_n=\infty$ and
$\sum_n\alpha_n^2<\infty$, $w_n$ is $\mathcal F_{n+1}$-measurable with
$\mathbb E[w_n\mid\mathcal F_n]=0$ and $\nrm{w_n}_F\le C$, and the $b_n$ are
random with $b_n\to0$ almost surely (no measurability is needed for them). Then
$x_n\to c$ almost surely.}
\end{lemma}
\begin{proof}
\rev{Write $x_n-c=u_n+v_n$ with $u_0=0$, $v_0=x_0-c$,
$u_{n+1}=(1-\alpha_n)u_n+\alpha_nw_n$ and $v_{n+1}=(1-\alpha_n)v_n+\alpha_nb_n$.
\emph{Noise.} Since $u_n$ depends only on $w_0,\dots,w_{n-1}$, it is
$\mathcal F_n$-measurable; as $\mathbb E[w_n\mid\mathcal F_n]=0$, the cross term
vanishes and
$\mathbb E[\nrm{u_{n+1}}_F^2\mid\mathcal F_n]=(1-\alpha_n)^2\nrm{u_n}_F^2+\alpha_n^2\mathbb E[\nrm{w_n}_F^2\mid\mathcal F_n]\le\nrm{u_n}_F^2-\alpha_n\nrm{u_n}_F^2+\alpha_n^2C^2$.
By the Robbins--Siegmund theorem \citep{robbins1971convergence}, $\nrm{u_n}_F^2$
converges almost surely and $\sum_n\alpha_n\nrm{u_n}_F^2<\infty$; as
$\sum_n\alpha_n=\infty$, the limit is $0$. \emph{Bias.} Fix a sample path and
$\varepsilon>0$, and take $n_0$ with $\nrm{b_n}_F\le\varepsilon$ for $n\ge n_0$.
Then $\nrm{v_{n+1}}_F-\varepsilon\le(1-\alpha_n)(\nrm{v_n}_F-\varepsilon)$, so
$\nrm{v_n}_F-\varepsilon\le\prod_{k=n_0}^{n-1}(1-\alpha_k)(\nrm{v_{n_0}}_F-\varepsilon)^+\to0$,
because $\sum_k\alpha_k=\infty$. Hence $\limsup_n\nrm{v_n}_F\le\varepsilon$ for
every $\varepsilon>0$.}
\end{proof}

\begin{proof}[\rev{Proof of Theorem~\ref{thm:sampled}}]
\rev{Call $(t,s)$ reachable if it has positive probability under some policy; the
reachable set is closed under transitions, so the targets of reachable entries
involve only reachable entries. Write
$V_{t+1}(s')=\LSE_\beta(\{Q_{t+1}(s',a')\}_{a'})$ for the current table. We show by
backward induction on $t=H,H-1,\dots,0$ that $Q_t(s,a)\to Q^\beta_t(s,a)$ almost
surely for every reachable $(t,s,a)$, and hence, by continuity of $\LSE_\beta$,
that $V_t(s)\to V^\beta_t(s)$ for every reachable $(t,s)$. For $t=H$ both sides
are $0$.}

\rev{Assume the claim at $t+1$, and fix a reachable $(t,s,a)$. The entry changes
only when it is updated; let $n=1,2,\dots$ index its updates, $k_n$ be the time of
the $n$-th one, $s'_n$ the sampled next state, and $\mathcal F_n$ the history
before $s'_n$ is drawn. The update is $x_{n+1}=x_n+\alpha_n(Y_n-x_n)$ with
$Y_n=\Kd a\bigl(\hat R_{k_n}(s,a)+\gamma V^{(k_n)}_{t+1}(s'_n)\bigr)$. Decompose
$Y_n=c+w_n+b_n$:
\begin{align*}
c&=\Kd a\Bigl(R(s,a)+\gamma\textstyle\sum_{s'}P(s'\mid s,a)V^\beta_{t+1}(s')\Bigr)=Q^\beta_t(s,a),\\
w_n&=\gamma\Kd a\Bigl(V^\beta_{t+1}(s'_n)-\textstyle\sum_{s'}P(s'\mid s,a)V^\beta_{t+1}(s')\Bigr),\\
b_n&=\Kd a\bigl(\hat R_{k_n}(s,a)-R(s,a)\bigr)+\gamma\Kd a\bigl(V^{(k_n)}_{t+1}(s'_n)-V^\beta_{t+1}(s'_n)\bigr).
\end{align*}
Since $s'_n\sim P(\cdot\mid s,a)$ given $\mathcal F_n$, $w_n$ is a martingale
difference, and it is bounded by $2\gamma\max_{s'}\nrm{V^\beta_{t+1}(s')}$ because
$V^\beta_{t+1}$ is a fixed, finite family. For $b_n$: $k_n\to\infty$ because the
entry is updated infinitely often; the first term tends to $0$ because
$\hat R\to R$ and $\nrm{\Kd aX}\le\nrm X$; and the second tends to $0$ because
$s'_n$ ranges over finitely many reachable states, at each of which
$V^{(k)}_{t+1}\to V^\beta_{t+1}$ by the induction hypothesis. Lemma~\ref{lem:sa}
gives $Q_t(s,a)\to Q^\beta_t(s,a)$ almost surely.}

\rev{\emph{What the theorem does not use, and what it does not give.} The induction
never compares two tables through $\T_\beta$, so neither monotonicity nor
non-expansiveness of $\LSE_\beta$ enters; it needs only that the targets of step
$t$ stop moving, which acyclicity in $t$ ensures. For an infinite horizon the
targets depend on the same table, and the standard argument
\citep{tsitsiklis1994asynchronous} needs $\T_\beta$ to contract, which
Theorem~\ref{thm:certificate} gives only under $\gamma e^{2\beta M_\beta}<1$. Two
consequences follow for the learned operators. First, they are asymptotically
certificates: for every policy $\pi$ and density $\rho$,
$q^\pi_t(s,\rho,a)\le\tau(\rho Q_t(s,a))+\tau(\rho)\nrm{Q_t(s,a)-Q^\beta_t(s,a)}$, and
the last term tends to $0$. Second, wherever the greedy action of $Q^\beta_t$ at
$(s,\rho)$ is unique, the learned greedy action equals it after finitely many
updates. The exploration condition holds for $\varepsilon$-greedy behaviour with
$\varepsilon$ bounded below, and the step sizes $n^{-p}$ with $p\in(1/2,1]$ are
Robbins--Monro; the condition on $\hat R$ holds when $R$ is known and, for the
least-squares estimate, under zero-mean reward noise and persistent excitation
(the observed latents keep spanning the self-adjoint operators). With a step-size
floor, as in E18--E23, the iterates do not converge but fluctuate around
$Q^\beta$; E24 measures by how much.}
\end{proof}

\section{Proofs for Section~\ref{sec:theory}}\label{app:lipschitz}

\subsection{Proof of Proposition~\ref{prop:wellposed_new}}
\begin{proof}
Each $e^{\beta x_i}\succeq e^{-\beta M}I$, so
$S=\sum_ie^{\beta x_i}\succeq Ke^{-\beta M}I\succ0$ and $\log S$ is
defined and self-adjoint. (i) $S\succeq e^{\beta x_i}$ and $\log$ is
operator monotone \citep[Ch.~V]{bhatia1997matrix}, so
$\log S\succeq\beta x_i$. (ii) $e^{\beta(x_i+cI)}=e^{\beta c}e^{\beta x_i}$.
(iii) With $m=\max_i\nrm{x_i}$, $S\preceq Ke^{\beta m}I$ and operator
monotonicity of $\log$ give $\LSE_\beta(x)\preceq(m+\beta^{-1}\log K)I$,
while (i) gives $\LSE_\beta(x)\succeq x_1\succeq-mI$. (iv) $U^*f(A)U=f(U^*AU)$
for any functional calculus. (v) In a joint eigenbasis all operations
are entrywise.
\end{proof}

\begin{lemma}[Scalar case]\label{lem:exact_lip}
For real $u,v\in\mathbb R^K$ and every $\beta>0$,
$|\mathrm{lse}_\beta(u)-\mathrm{lse}_\beta(v)|\le\max_i|u_i-v_i|$,
with equality for uniform shifts.
\end{lemma}
\begin{proof}
With $D=\max_i|u_i-v_i|$, monotonicity and the shift identity give
$\mathrm{lse}_\beta(v)-D\le\mathrm{lse}_\beta(u)\le\mathrm{lse}_\beta(v)+D$.
\end{proof}

\subsection{Proof of Proposition~\ref{prop:lse_lipschitz_new}}
\begin{proof}
(a) In the joint eigenbasis, coordinate $j$ of $\Delta$ is
$\Delta_j=\mathrm{lse}_\beta(x^{(j)})-\mathrm{lse}_\beta(y^{(j)})$, and by
Lemma~\ref{lem:exact_lip}, $|\Delta_j|\le\max_i|h_i^{(j)}|$. Taking
$\max_j$ gives the operator-norm bound; averaging squares gives
$\nrm\Delta_2^2=\tfrac1N\sum_j\Delta_j^2\le\tfrac1N\sum_j\sum_i|h_i^{(j)}|^2=\sum_i\nrm{h_i}_2^2$.
A uniform shift $h_i=cI$ attains the first bound. For the second, take
$K=2$, $x_1=\Delta_0 I$, $x_2=0$ and perturb only $x_1$: the ratio is the
softmax weight $(1+e^{-\beta\Delta_0})^{-1}\to1$.

(b) Let $X=\sum_ie^{\beta x_i}$, $Y=\sum_ie^{\beta y_i}$, both
$\succeq m:=Ke^{-\beta M}$. From $\log x=\int_0^\infty[(1+s)^{-1}-(x+s)^{-1}]ds$,
\[
\log X-\log Y=\int_0^\infty(Y+s)^{-1}(X-Y)(X+s)^{-1}\,ds,
\]
so for $p\in\{2,\infty\}$,
$\nrm{\log X-\log Y}_p\le\nrm{X-Y}_p\int_0^\infty(m+s)^{-2}ds=\nrm{X-Y}_p/m$.
(The one-line formula $\int_0^1((1-t)Y+tX)^{-1}(X-Y)\,dt$ is valid only
when $X$ and $Y$ commute.) By Duhamel's formula
$e^{\beta x}-e^{\beta y}=\beta\int_0^1e^{u\beta x}(x-y)e^{(1-u)\beta y}du$,
so $\nrm{e^{\beta x_i}-e^{\beta y_i}}_p\le\beta e^{\beta M}\nrm{h_i}_p$.
Combining and dividing by $\beta$,
$\nrm\Delta_p\le\tfrac{e^{\beta M}}{K}\,e^{\beta M}\sum_i\nrm{h_i}_p$.

(c) In the common basis, coordinate $j$ of the $K=N$ family has values
$1$ for action $j$ and $0$ otherwise. Perturbing $x_i$ by $\epsilon\,e_ie_i^*$
changes coordinate $j$ only through action $j$, with softmax weight
$p_\beta=e^\beta/(e^\beta+N-1)$, so $\Delta=\epsilon p_\beta I+O(\epsilon^2)$
and $\nrm\Delta_2=\epsilon p_\beta+O(\epsilon^2)$, whereas
$\nrm{h_i}_2=\epsilon/\sqrt N$ for each $i$ and
$(\sum_i\nrm{h_i}_2^2)^{1/2}=\epsilon$. Hence the max-aggregated ratio
tends to $p_\beta\sqrt N$ and the $\ell_2$-aggregated ratio to $p_\beta\le1$.
At $N=25$, $\beta=5$: $p_\beta\sqrt N=4.304$.

\end{proof}

\subsection{Proof of Proposition~\ref{prop:monotone_new}}
\begin{proof}
On the commuting sector $\LSE_\beta$ is the scalar log-sum-exp in each
eigen-coordinate, which is non-decreasing in each argument; $\Kd a$ is
positive, so $\T_\beta$ inherits monotonicity. In general the argument
fails because $x\mapsto e^{\beta x}$ is not operator monotone
\citep[Ch.~V]{bhatia1997matrix}; E12 gives explicit violations
(Sec.~\ref{sec:experiments}). $\T^\pi$ is affine with positive
linear part in CP, hence monotone.
\end{proof}

\section{Proofs for Section~\ref{sec:osMDP}}\label{app:horizon}

\subsection{Proof of Proposition~\ref{thm:representation_gap}}
\begin{proof}
Fix a prefix $g$ of length $T-3$ and actions $a,b,c$. The histories
$h=(g,a,b)$ and $h'=(g,b,a)$ are permutations of each other, so
$\sigma(h)=\sigma(h')$ and $f(\sigma(h),c)=f(\sigma(h'),c)$. Therefore
$\max\{|f-q_T(h,c)|,|f-q_T(h',c)|\}\ge\tfrac12|q_T(h,c)-q_T(h',c)|
=\tfrac1{2N}|\Tr(WX_g(A_aA_b-A_bA_a)A_c)|$. If $[A_a,A_b]\ne0$ then
$X_g[A_a,A_b]A_c\ne0$ (the outer factors are invertible), so
$W\mapsto\Tr(WX_g[A_a,A_b]A_c)$ is a non-zero linear functional whose
zero set is a hyperplane, of Lebesgue measure zero. The exact operator
form is Corollary~\ref{cor:product}.
\end{proof}

\subsection{Proof of Proposition~\ref{thm:horizon}}
\begin{proof}
(a) Distinct positive words are distinct reduced words, so in a free
group of rank $K$ the $K^{T-1}$ histories of length $T-1$ give pairwise
distinct $X_h$. For each of the finitely many pairs $h\ne h'$ and each
$a$, $\Tr(W(X_h-X_{h'})A_a)=0$ defines a hyperplane in $W$; outside their
union all values $q_T(h,a)$ with distinct $h$ differ for every $a$. An
exact table indexed by $(\sigma(h),a)$ must then have $\sigma(h)\ne\sigma(h')$
for all $h\ne h'$, i.e.\ at least $K^{T-1}$ observation values and $K^T$
entries. For $SO(3)$ and $SE(3)$ (connected, non-solvable), almost every
$K$-tuple generates a free group of rank $K$ \citep{epstein1971almost}.

(b) Step-$2$ nilpotency means every bracket is central. For central
$[X,Y]$, $e^Xe^Y=e^{X+Y+\frac12[X,Y]}$, and by induction
$e^{\Omega_{h_1}}\cdots e^{\Omega_{h_t}}=\exp\bigl(\sum_i\Omega_{h_i}+\tfrac12\sum_{i<j}[\Omega_{h_i},\Omega_{h_j}]\bigr)$
(the new bracket at each step is central, so it commutes past the
accumulated exponent). The first sum equals $\sum_an_a(h)\Omega_a$ and
the second $\sum_{a<b}L_{ab}(h)[\Omega_a,\Omega_b]$. Hence $X_h$, and
$q_T(h,\cdot)$, are functions of $(n(h),L(h))$, which takes at most
$(T+1)^K(2T^2+1)^{K(K-1)/2}$ values.

\emph{Bounded deficit for compact groups.} For orthogonal $X$,
$\nrm X_1=1$ and $|q_T|\le\nrm W$, so the constant $f=0$ already has
error at most $\nrm W$ for every $T$.
\end{proof}

\subsection{Proof of Proposition~\ref{thm:upper}}
\begin{proof}
(i) is Corollary~\ref{cor:product}. (ii) For any $d$-dimensional model,
$g(uv)=\tfrac1d\Tr(W_dQ_uQ_v)=\tfrac1d\sum_{i,j}(W_dQ_u)_{ji}(Q_v)_{ij}$
is an inner product between a vector in $\mathbb C^{d^2}$ depending only
on $u$ and one depending only on $v$, so every Hankel block has rank at
most $d^2$; with $d=N$, $W_d=W$, $Q_a=A_a$ this also shows
$\operatorname{rank}\mathbf H_f\le N^2$. Exactness forces
$\mathbf H_g=\mathbf H_f$. (iii) On the block
$B=\mathbf H_f[\mathcal U,\mathcal V]$ the model's block $G$ has rank at
most $d^2$, so $\nrm{B-G}_F\ge(\sum_{j>d^2}\sigma_j(B)^2)^{1/2}$ by the
Eckart--Young theorem, and the largest entry of $B-G$ is at least
$\nrm{B-G}_F/\sqrt{|\mathcal U||\mathcal V|}$.
\end{proof}

\subsection{Additional Implementation Notes}
\rev{\paragraph{Positivity holds without clipping.} If $x_1,\dots,x_K\succeq0$, then
$e^{\beta x_i}\succeq I$, so $\sum_ie^{\beta x_i}\succeq KI$, and since $\log$ is operator
monotone, $\LSE_\beta(x)\succeq\beta^{-1}\log K\,I\succeq0$. With $R\succeq0$ and positive
$\mathcal K_a^\dagger$, the soft recursion therefore preserves positive semidefiniteness
exactly. Numerically we evaluate
$\LSE_\beta(x)=cI+\beta^{-1}\log\sum_ie^{\beta(x_i-cI)}$ with $c=\max_i\lambda_{\max}(x_i)$,
which avoids overflow.}
\paragraph{Spectral clipping.} For the soft recursion with function
approximation one may clip eigenvalues to $[-M_\beta,M_\beta]$; by
Theorem~\ref{thm:certificate} the exact recursion never leaves this ball,
so clipping is a numerical safeguard only.
\paragraph{Numerical evaluation of $\LSE_\beta$.} $\sum_ie^{\beta x_i}$
has condition number up to $e^{2\beta M}$; for $\beta M\gtrsim10$ we
evaluate $\LSE_\beta$ in extended precision (E7, E17).

\section{Extensions}\label{app:extensions}

This appendix extends the results of Sections~\ref{sec:opq}
and~\ref{sec:osMDP} in three directions: approximate operator values,
noisy or random latent dynamics, and reducible action representations.
Experiment E17b (\texttt{experiments/e17b\_extension\_checks.py} in
\coderepo) checks each statement on finite instances.

\begin{proposition}[Robust latent-aware improvement]\label{prop:robust}
Let $\pi$ be latent-blind with $\gamma\kappa<1$, and let $\hat Q$ satisfy
$\max_{s,a}\nrm{\hat Q(s,a)-Q^\pi(s,a)}\le\varepsilon$. Let
$\hat\pi(s,\rho)\in\arg\max_a\tau(\rho\hat Q(s,a))$. Then for all
$(s,\rho)$,
\[
V^{\hat\pi}(s,\rho)\;\ge\;V^\pi(s,\rho)-\frac{2\varepsilon\,\nrm{\rho}_1}{1-\gamma\kappa}.
\]
\end{proposition}

\begin{proof}
By Hölder, $|\tau(\rho(\hat Q-Q^\pi)(s,a))|\le\varepsilon\nrm\rho_1$. Hence
$q^\pi(s,\rho,\hat\pi(s,\rho))\ge\tau(\rho\hat Q(s,\hat\pi))-\varepsilon\nrm\rho_1
\ge\tau(\rho\hat Q(s,a))-\varepsilon\nrm\rho_1\ge q^\pi(s,\rho,a)-2\varepsilon\nrm\rho_1$
for every $a$. Averaging over $a\sim\pi(\cdot\mid s)$ gives
$q^\pi(s,\rho,\hat\pi(s,\rho))\ge V^\pi(s,\rho)-2\varepsilon\nrm\rho_1$. Since
$q^\pi(s,\rho,a)=\mathbb E[r_0+\gamma V^\pi(s_1,\rho_1)]$, unrolling $n$
steps under $\hat\pi$ gives
$V^\pi(s_0,\rho_0)\le\mathbb E^{\hat\pi}\bigl[\sum_{t<n}\gamma^tr_t\bigr]
+2\varepsilon\sum_{t<n}\gamma^t\,\mathbb E\nrm{\rho_t}_1+\gamma^n\mathbb E\,V^\pi(s_n,\rho_n)$.
With $\nrm{\rho_t}_1\le\kappa^t\nrm{\rho_0}_1$ and
$|V^\pi(s,\rho)|\le\nrm{Q^\pi}_{\infty,\mathrm{op}}\nrm\rho_1$, letting
$n\to\infty$ proves the claim.
\end{proof}

Combined with Theorem~\ref{thm:td}, this turns an operator TD estimate
with error $\varepsilon$ into a latent-aware controller whose loss relative
to $\pi^+$ of Proposition~\ref{prop:improve} is $O(\varepsilon)$.

\begin{proposition}[Noisy and random latent dynamics]\label{prop:noisy}
Suppose that at each step, given the action $a$, the latent is updated by
a linear map $\mathcal K_a^\omega$, where $\omega\sim\nu_a$ is drawn
independently of the past and of $s'$, and
$\sup_\omega\nrm{(\mathcal K_a^\omega)^\dagger}\le\kappa$. Let
$\bar{\mathcal K}_a=\mathbb E_\omega\mathcal K_a^\omega$.
(i) Proposition~\ref{thm:exact_rep} and Corollary~\ref{cor:product} hold with
$\Kd a$ replaced by $\bar{\mathcal K}_a^\dagger$.
(ii) If every $\mathcal K_a^\omega$ is positive---for example, a
completely positive trace-preserving quantum channel with decoherence or
amplitude damping---then Theorem~\ref{thm:certificate} holds with
$\bar{\mathcal K}_a^\dagger$, for every policy, including policies that
observe the realised latent. In the infinite-horizon truncation bound,
$\tau(\rho)$ is replaced by $\sup_t\mathbb E\,\tau(\rho_t)$, which equals
$\tau(\rho)$ for trace-preserving maps.
\end{proposition}

\begin{proof}
(i) In the backward induction of Proposition~\ref{thm:exact_rep}, the one-step
identity becomes
$q^\pi_t(s,\rho,a)=\mathbb E_\omega\bigl[\tau\bigl(\mathcal K_a^\omega(\rho)\,(R(s,a)+\gamma Y)\bigr)\bigr]$
with $Y$ as before, using the independence of $\omega$ and $s'$. By
linearity this is
$\tau(\bar{\mathcal K}_a(\rho)(R+\gamma Y))=\tau(\rho\,\bar{\mathcal K}_a^\dagger(R+\gamma Y))$.
The infinite-horizon argument is unchanged, since
$\nrm{\bar{\mathcal K}_a^\dagger}\le\kappa$.
(ii) Apply the induction of Theorem~\ref{thm:certificate} conditionally on
$\omega$, with $\rho'=\mathcal K_a^\omega(\rho)\succeq0$. The two
inequalities hold for each $\omega$, and taking $\mathbb E_\omega$ of the
final expression gives $\tau(\rho\,\bar{\mathcal K}_a^\dagger(R+\gamma\sum_{s'}PV^\beta_{t+1}))$.
Positive maps preserve self-adjointness, so $Q^\beta_t$ remains
self-adjoint.
\end{proof}

\begin{proposition}[Structure-refined operator dimension]\label{prop:structure}
Let the $A_a$ be unitary and let $\mathcal A$ be the unital algebra they
generate. $\mathcal A$ is a finite-dimensional C$^*$-algebra, so
$\mathcal A\cong\bigoplus_{k=1}^r\mathcal M_{n_k}$, acting on
$\mathbb C^N$ with multiplicities $m_k$ ($\sum_km_kn_k=N$). For Lebesgue-almost
every $W$, $\operatorname{rank}\mathbf H_f=\dim\mathcal A=\sum_kn_k^2$, and
the minimal exact complex operator dimension satisfies
\[
\Bigl(\sum_kn_k^2\Bigr)^{1/2}\;\le\;d_{\min}\;\le\;\sum_kn_k .
\]
In particular $d_{\min}=N$ when the action representation is irreducible
($\mathcal A=\mathcal M_N$, the generic case), and $d_{\min}<N$ whenever an
irreducible component occurs with multiplicity $m_k>1$.
\end{proposition}

\begin{proof}
The span of all words $A_w$, including the empty word, is closed under
multiplication and contains $I$, so it equals $\mathcal A$. By
Cayley--Hamilton each $A_a^{-1}=A_a^*$ is a polynomial in $A_a$, so
$\mathcal A$ is $*$-closed, hence a C$^*$-algebra, and the structure
theorem applies. Since words span $\mathcal A$ on both sides,
$\operatorname{rank}\mathbf H_f$ equals the rank of the bilinear form
$B_W(X,Y)=\Tr(WXY)$ on $\mathcal A\times\mathcal A$. At $W=I$ this form is
non-degenerate: if $\Tr(XY)=0$ for all $Y\in\mathcal A$, take $Y=X^*$ to
get $X=0$. Its Gram determinant in a basis of $\mathcal A$ is a polynomial
in $W$ that does not vanish at $I$, so it is non-zero for almost every $W$,
and then $\operatorname{rank}\mathbf H_f=\dim\mathcal A$. The lower bound
follows from Proposition~\ref{thm:upper}(ii). For the upper bound, choose a
unitary $V$ with $V^*A_wV=\bigoplus_k(I_{m_k}\otimes B^{(k)}_w)$. Then
$\Tr(WA_w)=\sum_k\Tr(\widetilde W_kB^{(k)}_w)$, where $\widetilde W_k$ is
the partial trace over the multiplicity factor of the $k$-th diagonal
block of $V^*WV$. The model $Q_a=\bigoplus_kB^{(k)}_a$ with readout
proportional to $\bigoplus_k\widetilde W_k$ has dimension $\sum_kn_k$ and
reproduces $f$ exactly. Real orthogonal actions are covered by
complexification.
\end{proof}

E17b checks these statements on finite instances.
\begin{itemize}
\item Proposition~\ref{prop:noisy}, with random choices between two
amplitude-damping channels: the exact representation holds to
$3.9\times10^{-16}$, and the certificate has $0/20$ violations at
$\beta\in\{2,10\}$.
\item Proposition~\ref{prop:robust}, with $\varepsilon=0.05$: the bound
holds with margin $1.12$.
\item Proposition~\ref{prop:structure}, for a reducible $N{=}4$
representation with blocks $n=(2,1)$ and multiplicities $m=(1,2)$: the
Hankel rank is $5=\dim\mathcal A$, and an explicit $d{=}3$ model reproduces
all $243$ words of length $5$ to $1.8\times10^{-15}$, whereas a generic
$N{=}3$ representation has Hankel rank $9$.
\end{itemize}

\section{Summary of All Experiments}\label{app:expsummary}

Table~\ref{tab:map} maps each theoretical claim to the experiment that tests
it; Table~\ref{tab:expsummary} summarises the experiments.

\begin{table}[h]
\caption{Theory--experiment map.}
\label{tab:map}
\centering\scriptsize\setlength{\tabcolsep}{2.5pt}
\begin{tabular}{p{2.6cm}p{1.3cm}p{5.5cm}}
\toprule
Claim & Exp & Prediction / outcome \\
\midrule
P\ref{thm:exact_rep} exact representation & E17, E18 & readout $=$ exact expected return ($\le7\times10^{-16}$); zero-shot transfer across $\rho_0$ \\
T\ref{thm:td} operator TD & E17, E18 & $\max\nrm{Q_k-Q^\pi}\to0$; OpQ-PI \\
P\ref{prop:improve} improvement & E18 & OpQ-PI improves on the uniform policy \\
T\ref{thm:certificate} certificate & E17, E18 & $0$ violations; $Q^\beta-Q^\pi\succeq0$; slack consistent with OpQ-soft's ceiling \\
P\ref{prop:reduction} reduction & E9, E17 & exact collapse to soft $Q$-learning; slack $\to0$ at pure eigenstates \\
P\ref{prop:lse_lipschitz_new} Lipschitz & E7 & commuting ratio $\le1$; $4.30$ at $K{=}N{=}25$; non-commuting ratio $>1$, growing in $\beta$ \\
P\ref{prop:monotone_new} monotonicity & E12 & commuting only \\
P\ref{thm:representation_gap} aliasing & E17, E18 & positive swap gap; order-blind scalar agents fail when the latent varies \\
P\ref{thm:horizon} free vs.\ nilpotent & E10 & depth profile truncates for $\mathfrak n_3$, not for $\mathfrak{so}(3)$ \\
P\ref{thm:upper} dimension & E8, E14--E17 & exact at $d{=}N$ (real form: $2N_F$); Hankel rank $N^2$ \\
P\ref{thm:upper} scope & E13--E16 & sequence models vs.\ operator model \\
\rev{C\ref{cor:anybound} any upper bound} & \rev{E20} & \rev{SDP aggregator certifies; tighter on C1, not better control} \\
\rev{P\ref{prop:nearopt} slack spread} & \rev{E20} & \rev{bound holds on every instance; conservative by $5$--$21\times$} \\
\rev{knowledge of $K_a$} & \rev{E19} & \rev{$2\times2$ study, planners, horizons up to 20} \\
\rev{$M$-operator lower bound} & \rev{E21} & \rev{$L_M\le V^\ast$ in every case; reaches $V^\ast$ on C3} \\
\rev{miscalibrated $K_a$} & \rev{E21} & \rev{robust on C2; sensitive on C1, C3 at the training latent} \\
\rev{OpQ-soft vs.\ $\beta$} & \rev{E22} & \rev{no divergence in 108 runs; little late switching} \\
\rev{P\ref{thm:exact_rep} hidden latent} & \rev{E23} & \rev{exact plug-in of the belief mean; mixed speed vs.\ belief-mean DQN} \\
\rev{T\ref{thm:sampled} sampled convergence} & \rev{E24} & \rev{error to $Q^\beta\le0.01$; greedy score equals exact $Q^\beta$} \\
\bottomrule
\end{tabular}
\end{table}

\begin{table}[h]
\centering\scriptsize\setlength{\tabcolsep}{3pt}\renewcommand{\arraystretch}{1.25}
\begin{tabular}{lp{2.3cm}p{3.4cm}p{4.7cm}}
\toprule
Exp & Tests & Setup & Headline result \\
\midrule
E7 & Prop.~\ref{prop:lse_lipschitz_new} &
  Commuting/generic pairs; $K{=}N{=}25$ construction; $80$-digit checks &
  Commuting ratio $\le1.000000$; $4.304$ vs.\ $2.25$; non-commuting $0.97/1.17/1.48$ at $\beta{=}1/5/10$ \\
E8, E13 & Prop.~\ref{thm:upper} &
  Random orthogonal, $N{=}5$, $K{=}4$, $T{=}6$ & OpQ at $d{=}5$: $2.6\times10^{-30}$ (125 params); best baseline LSTM 0.033 (17729 params) \\
E9 & Prop.~\ref{prop:reduction} &
  Diagonal unitaries, $N\in\{1,4\}$ & Coincides with soft $Q$-learning to $1.6\times10^{-14}$ \\
E10 & Prop.~\ref{thm:horizon} &
  $\mathfrak{so}(3)$ vs.\ $\mathfrak{n}_3$, matched norm, $T\le10$ &
  $\mathfrak{so}(3)$: constant per-depth term $0.221$; $\mathfrak n_3$: zero beyond depth one \\
E12 & Prop.~\ref{prop:monotone_new} &
  $K{=}4$, $N{=}5$, $\beta{=}2$ & $0/2000$ commuting vs.\ $72/2000$ generic violations; persists at scale $10^{-4}$ \\
E14 & Prop.~\ref{thm:upper} & $\mathfrak{so}(3)$, $K{=}6$, $T{=}8$ & OpQ at $d{=}3$: $6.0\times10^{-32}$ (63 params); best baseline LSTM 0.003 (18241 params) \\
E15 & Prop.~\ref{thm:upper} & Displacements, $N_F{=}6$ & OpQ at $d{=}12$: $9.7\times10^{-15}$ (1008 params); best baseline GRU 0.042 (1121 params) \\
E16 & Prop.~\ref{thm:upper} & $\mathfrak{se}(3)$ hinge chain & OpQ at $d{=}4$: $2.2\times10^{-31}$ (112 params); best baseline GRU 0.092 (13697 params) \\
E17 & P\ref{thm:exact_rep}, T\ref{thm:td}, T\ref{thm:certificate}, P\ref{thm:representation_gap}, P\ref{thm:upper} &
  Random OLDP, $|\mathcal S|{=}2$, $K{=}N{=}3$, $H{=}4$ & See Sec.~\ref{sec:experiments} \\
E18 & End-to-end control & C1--C3, 7 agents, 5 seeds & Table~\ref{tab:e18}, Fig.~\ref{fig:e18} \\
\rev{E19} & \rev{Knowledge of $K_a$; planners} & \rev{C1--C3, $2\times2$ + controls, exhaustive and UCT search, $H\le20$} & \rev{Table~\ref{tab:e19}; App.~\ref{app:e19}} \\
\rev{E20} & \rev{Cor.~\ref{cor:anybound}, Prop.~\ref{prop:nearopt}} & \rev{Exact model; SDP aggregator, $M$ operators, slack spread, leaf value} & \rev{$M{=}8$--$32$ closes the gap on C1, C3; App.~\ref{app:e20}} \\
\rev{E21} & \rev{Miscalibration; cost of $M$} & \rev{$\tilde U_a=U_ae^{\varepsilon A_a}$, observed and tracked latent; $M$-operator lower bound and cost} & \rev{App.~\ref{app:e21}} \\
\rev{E22} & \rev{OpQ-soft vs.\ $\beta$} & \rev{$\beta\in[0.5,20]$, C1--C3, both regimes, 3 seeds} & \rev{no divergence; App.~\ref{app:e22}} \\
\rev{E23} & \rev{Hidden initial latent} & \rev{prior with known mean; belief-mean agents; plug-in} & \rev{App.~\ref{app:e23}} \\
\rev{E24} & \rev{Thm.~\ref{thm:sampled}} & \rev{C1--C3, two step-size rules, 3 seeds, 30,000 episodes} & \rev{converges to $Q^\beta$; App.~\ref{app:e24}} \\
\bottomrule
\end{tabular}
\caption{All experiments at a glance. \rev{Experiment IDs are fixed labels shared with the scripts
(\texttt{experiments/eNN\_*.py}) and the results files. E1--E6 and E11 were preliminary
experiments that are not part of the paper; the remaining IDs are kept unchanged so that the
paper, the code and the discussion refer to the same runs. Section~\ref{sec:experiments}
presents the experiments by kind of evidence (checks, representation, control), not by ID.}}
\label{tab:expsummary}
\end{table}

\section{Additional Results}\label{app:prior}

\subsection{Finite-instance checks of the theory (E17, E7, E9, E12)}
\textbf{E17} uses a random OLDP with $|\mathcal S|{=}2$, $K{=}N{=}3$, $H{=}4$
(Appendix~\ref{sec:e17-protocol}).
\emph{Proposition~\ref{thm:exact_rep}}: the readout matches the exact
expected return to $6.7\times10^{-16}$ (CP) and $3.6\times10^{-16}$ (MP).
\emph{Theorem~\ref{thm:td}}: operator TD reduces
$\max_{s,a}\nrm{Q_k-Q^\pi}$ from $1.69$ to $0.14$ over $4\times10^5$ steps.
\emph{Theorem~\ref{thm:certificate}}: for $\beta\in\{1,5,20,50\}$ there
are $0/30$ violations, and $\lambda_{\min}(Q^\beta-Q^\pi)>1.7$. The mean slack
($1.2$--$1.9$, against a mean $|q^*|$ of $3.45$) does not vanish as $\beta$
grows, which isolates its non-commutative part. In the commuting sector the
slack at pure eigenstates is $4.8\times10^{-12}$ at $\beta{=}200$.
\emph{Proposition~\ref{thm:representation_gap}}: a swapped prefix gives
$r(a,b,c)=-1.143$ and $r(b,a,c)=-0.807$, so any order-blind scalar errs by at
least $0.168$.
\emph{Proposition~\ref{thm:upper}}: the Hankel rank is $9=N^2$.
\textbf{E7}:
over $20{,}000$ commuting configurations the largest operator-norm
ratio is $1.000000$ (Prop.~\ref{prop:lse_lipschitz_new}a);
the construction of Prop.~\ref{prop:lse_lipschitz_new}(c) at
$K{=}N{=}25$, $\beta{=}5$ gives $4.304$, exceeding
$1+\beta\,\mathrm{diam}/4=2.25$, with $\ell_2$-aggregated ratio $0.861$;
in $80$-digit arithmetic, non-commuting ratios reach $0.970$, $1.171$
and $1.477$ at $\beta=1,5,10$.
\textbf{E9}: with commuting actions, the operator recursion and the sample
updates coincide with $N$ decoupled soft $Q$-learners to $1.6\times10^{-14}$
(Prop.~\ref{prop:reduction}).
\textbf{E12}: monotonicity holds in all $2000$ commuting trials but fails in
$72/2000$ generic ones, and continues to fail at perturbation scale
$10^{-4}$ (Prop.~\ref{prop:monotone_new}). The proof of the certificate
does not require monotonicity.

\subsection{Representation (E8, E10, E13--E16)}
\begin{figure}[h]
  \centering
  \includegraphics[width=\columnwidth]{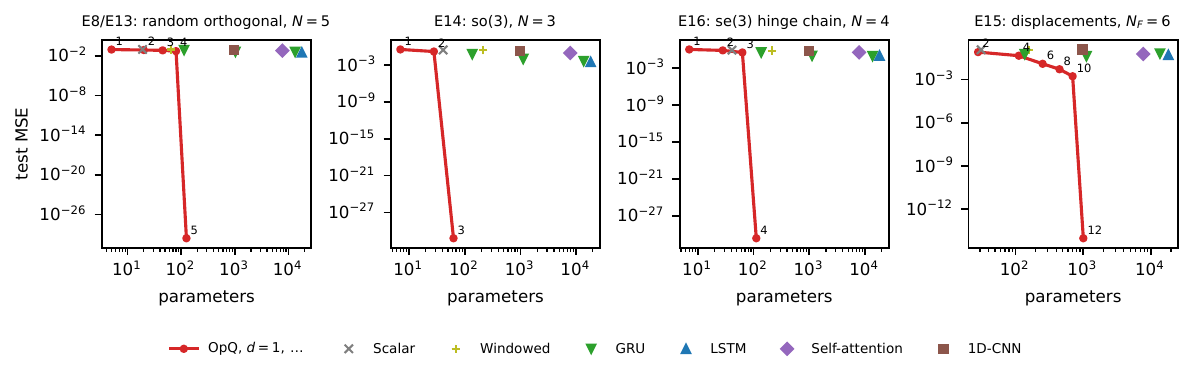}
  \caption{\textbf{E8, E13--E16.} Test MSE against the number of parameters.
  OpQ$(d)$ (red, labelled by $d$) against sequence-model and table
  baselines.}
  \label{fig:rep}
\end{figure}

\begin{figure}[h]
  \centering
  \includegraphics[width=0.8\columnwidth]{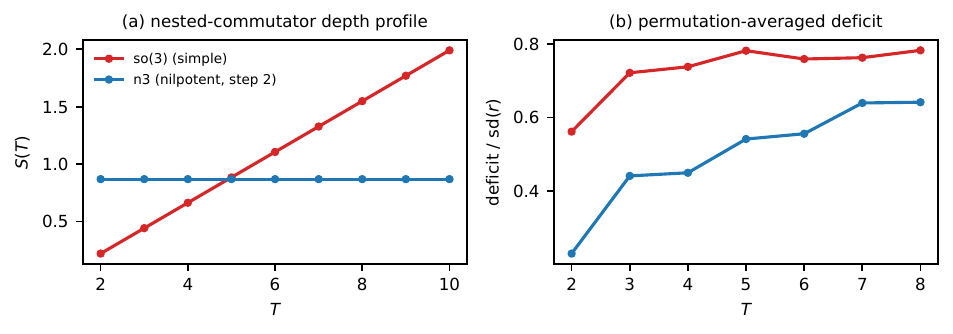}
  \caption{\textbf{E10.} (a) The nested-commutator depth profile has a
  constant per-depth contribution for $\mathfrak{so}(3)$ and is exactly zero
  beyond depth one for $\mathfrak n_3$. (b) The permutation-averaged deficit,
  normalised by the spread of returns, is bounded for both algebras, as
  $\lvert r\rvert\le\nrm{W}$ requires. Nilpotency controls the \emph{size} of
  an exact history statistic (Prop.~\ref{thm:horizon}), not whether order
  matters.}
  \label{fig:e10}
\end{figure}

Each benchmark fits the ordered-product readout of Proposition~\ref{thm:upper} to
exact returns ($3000$ training and $1000$ test words)
\rev{by supervised regression, not by TD: it tests the representational capacity of
the operator class, while E18--E23 test learning and control}. The benchmarks are
random orthogonal ($N{=}5$, E8/E13), $\mathfrak{so}(3)$ ($N{=}3$, E14), an
$\mathfrak{se}(3)$ hinge chain ($N{=}4$, E16) and truncated bosonic
displacements (real dimension $12$, E15). OpQ$(d)$ becomes exact at the
predicted dimension and not below it: test MSE is $\le3\times10^{-30}$ in
the first three benchmarks and $9.7\times10^{-15}$ for displacements, with
$63$--$1008$ parameters, against $4^6$--$6^8$ history-table entries. The
best other models are recurrent (GRU or LSTM), with test MSE
$0.003$--$0.09$ (return variances $0.09$--$1.28$). Attention, the
1D-CNN and the windowed table stay close to the order-blind baseline
(Table~\ref{tab:rep}, Fig.~\ref{fig:rep}). \textbf{E10} compares
$\mathfrak{so}(3)$ with the Heisenberg algebra $\mathfrak n_3$ at matched
generator norm. The per-depth commutator term is constant ($0.221$) for
$\mathfrak{so}(3)$ and exactly zero beyond depth one for $\mathfrak n_3$, the
nilpotency step that makes exact history statistics polynomial
(Proposition~\ref{thm:horizon}). The permutation-averaged deficit stays bounded for
both algebras (Fig.~\ref{fig:e10}), as $|r|\le\nrm W$ requires.

\subsection{E18: full control results}\label{app:e18full}

\begin{table}[h]
\centering\scriptsize\setlength{\tabcolsep}{2.2pt}
\begin{tabular}{lcc|cc|cc|cc|cc|cc}
\toprule
& \multicolumn{4}{c|}{C1 random OLDP} & \multicolumn{4}{c|}{C2 $SO(3)$ pointing} & \multicolumn{4}{c}{C3 two-qubit prep.}\\
& \multicolumn{2}{c|}{fixed} & \multicolumn{2}{c|}{random} & \multicolumn{2}{c|}{fixed} & \multicolumn{2}{c|}{random} & \multicolumn{2}{c|}{fixed} & \multicolumn{2}{c}{random}\\
Agent & 100 & 3000 & 100 & 3000 & 100 & 3000 & 100 & 3000 & 100 & 3000 & 100 & 3000\\
\midrule
OpQ-soft (learned $R$) & 0.77 & 0.84$\pm$0.01 & 0.95 & 0.95$\pm$0.00 & 0.94 & 0.98$\pm$0.00 & 0.95 & 0.99$\pm$0.00 & 0.38 & 0.83$\pm$0.08 & 0.58 & 0.68$\pm$0.00 \\
OpQ-soft (known $R$) & 0.78 & 0.84$\pm$0.01 & 0.95 & 0.95$\pm$0.00 & 0.95 & 0.98$\pm$0.00 & 0.96 & 0.99$\pm$0.00 & 0.67 & 0.79$\pm$0.00 & 0.63 & 0.68$\pm$0.00 \\
OpQ-PI & 0.84 & 0.86$\pm$0.00 & 0.96 & 0.97$\pm$0.01 & 0.95 & 0.91$\pm$0.04 & 0.96 & 0.98$\pm$0.00 & 0.28 & 0.28$\pm$0.00 & 0.59 & 0.70$\pm$0.00 \\
Scalar $Q$ on $(t,m)$ & 0.68 & 0.87$\pm$0.06 & 0.42 & 0.51$\pm$0.03 & 0.37 & 0.69$\pm$0.01 & 0.05 & 0.06$\pm$0.07 & 0.22 & 0.19$\pm$0.42 & 0.02 & 0.10$\pm$0.12 \\
Scalar $Q$, history table & 0.33 & 0.94$\pm$0.03 & 0.11 & 0.50$\pm$0.05 & 0.31 & 0.75$\pm$0.02 & -0.01 & 0.05$\pm$0.04 & 0.28 & 0.28$\pm$0.00 & -0.04 & -0.04$\pm$0.00 \\
DQN on $\rho_t$ & 0.96 & 1.00$\pm$0.00 & 0.89 & 0.99$\pm$0.00 & 0.85 & 0.99$\pm$0.00 & 0.73 & 0.99$\pm$0.00 & -0.08 & 1.00$\pm$0.00 & 0.04 & 0.87$\pm$0.03 \\
DRQN (GRU) on history & 0.76 & 1.00$\pm$0.00 & 0.44 & 0.95$\pm$0.01 & 0.69 & 0.97$\pm$0.01 & 0.28 & 0.94$\pm$0.01 & 0.14 & 0.96$\pm$0.08 & -0.01 & -0.01$\pm$0.05 \\
\bottomrule
\end{tabular}
\caption{E18: normalised return after $100$ and $3000$ training episodes
(mean; $\pm$ one standard deviation over $5$ seeds at $3000$).}
\label{tab:e18app}
\end{table}

\begin{table}[h]
\centering\scriptsize\setlength{\tabcolsep}{3pt}
\begin{tabular}{lcccc|c}
\toprule
& \multicolumn{4}{c|}{Greedy readout of exact $Q^\beta$ (known model)} & Exact OpQ-PI \\
Environment / regime & $\beta{=}1$ & $\beta{=}2$ & $\beta{=}5$ & $\beta{=}10$ & (Prop.~\ref{prop:improve}) \\
\midrule
random OLDP, fixed & 0.76 (2.30) & 0.76 (1.94) & 0.86 (1.88) & 0.87 (1.94) & 0.86 \\
random OLDP, random & 0.97 (1.73) & 0.95 (1.48) & 0.95 (1.43) & 0.96 (1.48) & 0.97 \\
$SO(3)$ pointing, fixed & 0.88 (2.44) & 0.99 (1.72) & 0.99 (1.32) & 0.98 (1.19) & 0.88 \\
$SO(3)$ pointing, random & 0.98 (2.43) & 0.99 (1.69) & 0.99 (1.30) & 0.99 (1.18) & 0.98 \\
two-qubit preparation, fixed & 0.28 (11.69) & 0.79 (6.17) & 0.79 (3.02) & 0.79 (2.01) & 0.28 \\
two-qubit preparation, random & 0.70 (12.60) & 0.69 (6.63) & 0.68 (3.23) & 0.68 (2.16) & 0.70 \\
\bottomrule
\end{tabular}
\caption{E18: exact-planning references with the known model. These are the
ceilings that the learned OpQ agents approach (compare Table~\ref{tab:e18}).
Numbers in parentheses give the certificate
$\mathbb E\,\max_a\tau(\rho_0Q^\beta_0(s_0,a))$ relative to $V^*$; it is
always at least $1$, as Theorem~\ref{thm:certificate} requires.}
\label{tab:e18ref}
\end{table}

\begin{figure}[h]
  \centering
  \includegraphics[width=\columnwidth]{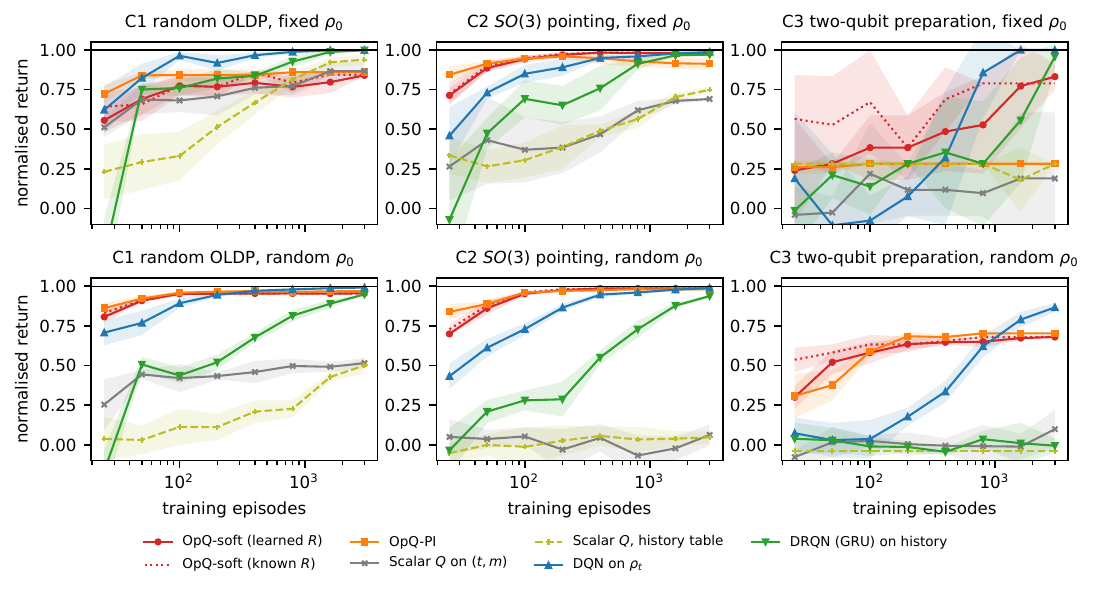}
  \caption{\textbf{E18.} Learning curves: normalised return of the greedy
  policy, evaluated exactly, against the number of training episodes (mean
  $\pm$ one standard deviation over $5$ seeds).}
  \label{fig:e18}
\end{figure}

\subsection{Representation benchmark: full results}\label{app:repfull}

\begin{table}[h]
\centering\scriptsize\setlength{\tabcolsep}{2.4pt}
\begin{tabular}{lrr|rr|rr|rr}
\toprule
& \multicolumn{2}{c|}{Orthogonal (E8/E13)} & \multicolumn{2}{c|}{$\mathfrak{so}(3)$ (E14)} & \multicolumn{2}{c|}{$\mathfrak{se}(3)$ (E16)} & \multicolumn{2}{c}{Displacements (E15)}\\
Model & params & test MSE & params & test MSE & params & test MSE & params & test MSE \\
\midrule
Scalar order-blind $(t,a_t)$ & 19 & 0.095 & 41 & 0.303 & 41 & 0.861 & 31 & 0.116 \\
Windowed table $(t,a_{t-1},a_t)$ & 64 & 0.094 & 216 & 0.277 & 216 & 0.730 & 156 & 0.115 \\
GRU, hidden 4 & 113 & 0.077 & 137 & 0.062 & 137 & 0.363 & 137 & 0.058 \\
GRU, hidden 16 & 1025 & 0.041 & 1121 & 0.010 & 1121 & 0.106 & 1121 & 0.042 \\
GRU, hidden 64 & 13313 & 0.040 & 13697 & 0.005 & 13697 & 0.092 & 13697 & 0.064 \\
LSTM, hidden 64 & 17729 & 0.033 & 18241 & 0.003 & 18241 & 0.107 & 18241 & 0.051 \\
Self-attention, width 32 & 7713 & 0.064 & 7841 & 0.096 & 7841 & 0.366 & 7777 & 0.059 \\
1D-CNN on action embeddings & 973 & 0.083 & 979 & 0.202 & 979 & 0.578 & 979 & 0.122 \\
OpQ, $d{=}1$ & 5 & 0.091 & 7 & 0.361 & 7 & 1.108 & -- & -- \\
OpQ, $d{=}2$ & 20 & 0.078 & 28 & 0.165 & 28 & 0.773 & 28 & 0.080 \\
OpQ, $d{=}3$ & 45 & 0.069 & 63 & $6.0\times10^{-32}$ & 63 & 0.361 & -- & -- \\
OpQ, $d{=}4$ & 80 & 0.052 & -- & -- & 112 & $2.2\times10^{-31}$ & 112 & 0.045 \\
OpQ, $d{=}5$ & 125 & $2.6\times10^{-30}$ & -- & -- & -- & -- & -- & -- \\
OpQ, $d{=}6$ & -- & -- & -- & -- & -- & -- & 252 & 0.012 \\
OpQ, $d{=}8$ & -- & -- & -- & -- & -- & -- & 448 & 0.005 \\
OpQ, $d{=}10$ & -- & -- & -- & -- & -- & -- & 700 & 0.002 \\
OpQ, $d{=}12$ & -- & -- & -- & -- & -- & -- & 1008 & $9.7\times10^{-15}$ \\
\midrule
Planted solution & & $1.1\times10^{-32}$ & & $2.1\times10^{-32}$ & & $1.1\times10^{-31}$ & & $2.1\times10^{-32}$ \\
Var(test returns) & & 0.094 & & 0.412 & & 1.282 & & 0.117 \\
History table & 4096 & $0$ & 1679616 & $0$ & 1679616 & $0$ & 46656 & $0$ \\
\bottomrule
\end{tabular}
\caption{Representation benchmark (E8, E13--E16): held-out test MSE. OpQ$(d)$
is exact at $d=N$ (real form $2N_F=12$ for displacements, to $10^{-14}$), as
Proposition~\ref{thm:upper} predicts.}
\label{tab:rep}
\end{table}

\clearpage
\section{\rev{E19: Model-Matched Baselines and Planners}}\label{app:e19}

\rev{This appendix gives the full results and protocol of E19 (Sec.~\ref{sec:e19}): the
$2\times2$ comparison in both regimes with every control, zero-shot transfer, planners
with the same knowledge as OpQ, longer horizons, and cost.}

\subsection{\rev{Further tables}}

\begin{table}[t]
\centering
\small
\setlength{\tabcolsep}{4pt}
\resizebox{\textwidth}{!}{%
\begin{tabular}{@{}lcccccc@{}}
\toprule
 & \multicolumn{2}{c}{C1} & \multicolumn{2}{c}{C2} & \multicolumn{2}{c}{C3} \\
\cmidrule(lr){2-3}\cmidrule(lr){4-5}\cmidrule(lr){6-7}
Agent & train $\rho_0$ & held-out & train $\rho_0$ & held-out & train $\rho_0$ & held-out \\
\midrule
DQN on $\rho_t$ & $1.00${\scriptsize$\pm$0.00} & $0.86${\scriptsize$\pm$0.04} & $0.98${\scriptsize$\pm$0.01} & $0.96${\scriptsize$\pm$0.02} & $1.00${\scriptsize$\pm$0.00} & $0.46${\scriptsize$\pm$0.11} \\
DQN + $K_a$ data & $1.00${\scriptsize$\pm$0.00} & $\mathbf{0.99}${\scriptsize$\pm$0.00} & $0.98${\scriptsize$\pm$0.01} & $0.98${\scriptsize$\pm$0.00} & $1.00${\scriptsize$\pm$0.00} & $\mathbf{0.84}${\scriptsize$\pm$0.03} \\
DQN + $K_a$ data, reachable $\rho$ & $1.00${\scriptsize$\pm$0.00} & $0.88${\scriptsize$\pm$0.02} & $0.98${\scriptsize$\pm$0.01} & $0.96${\scriptsize$\pm$0.01} & $1.00${\scriptsize$\pm$0.00} & $0.59${\scriptsize$\pm$0.04} \\
Linear-Q & $0.99${\scriptsize$\pm$0.00} & $0.75${\scriptsize$\pm$0.03} & $0.95${\scriptsize$\pm$0.02} & $0.78${\scriptsize$\pm$0.02} & $0.28${\scriptsize$\pm$0.00} & $0.37${\scriptsize$\pm$0.05} \\
\upd{Linear-Q + $K_a$ data} & \upd{$0.86${\scriptsize$\pm$0.00}} & \upd{$0.97${\scriptsize$\pm$0.01}} & \upd{$0.91${\scriptsize$\pm$0.02}} & \upd{$0.97${\scriptsize$\pm$0.00}} & \upd{$0.73${\scriptsize$\pm$0.24}} & \upd{$0.71${\scriptsize$\pm$0.01}} \\
OpQ-soft & $0.84${\scriptsize$\pm$0.01} & $0.95${\scriptsize$\pm$0.00} & $0.98${\scriptsize$\pm$0.00} & $\mathbf{0.99}${\scriptsize$\pm$0.00} & $0.83${\scriptsize$\pm$0.08} & $0.68${\scriptsize$\pm$0.01} \\
\midrule
Model + exhaustive & $1.00${\scriptsize$\pm$0.00} & $1.00${\scriptsize$\pm$0.00} & $1.00${\scriptsize$\pm$0.00} & $1.00${\scriptsize$\pm$0.00} & $1.00${\scriptsize$\pm$0.00} & $1.00${\scriptsize$\pm$0.00} \\
\bottomrule
\end{tabular}}
\caption{E19, zero-shot transfer. Agents train for 3000 episodes at one fixed initial latent and are evaluated, without retraining, at that latent and on 8 held-out latents (normalised return, mean $\pm$ s.d.\ over 5 seeds). Bold marks the best held-out score among the first four agents of the $2\times2$ comparison.}
\label{tab:e19-transfer}
\end{table}

\begin{table}[t]
\centering
\small
\setlength{\tabcolsep}{5pt}
\begin{tabular}{@{}lccc@{}}
\toprule
 & C1 & C2 & C3 \\
\midrule
\multicolumn{4}{@{}l}{\emph{Normalised return at the training latent / on held-out latents}} \\
Exhaustive search, model from 25 episodes & $0.99\,/\,0.97$ & $0.99\,/\,1.00$ & $0.83\,/\,0.88$ \\
Exhaustive search, model from 50 episodes & $1.00\,/\,1.00$ & $1.00\,/\,1.00$ & $0.96\,/\,0.94$ \\
Exhaustive search, model from 100 episodes & $1.00\,/\,1.00$ & $1.00\,/\,1.00$ & $1.00\,/\,1.00$ \\
MCTS, 100 simulations, model from 100 episodes & $0.89\,/\,0.93$ & $0.95\,/\,0.95$ & $0.38\,/\,0.61$ \\
MCTS, 1000 simulations, model from 100 episodes & $0.93\,/\,0.96$ & $0.99\,/\,0.99$ & $0.83\,/\,0.73$ \\
OpQ-soft, 100 episodes (E18, fixed latent) & $0.77$ & $0.94$ & $0.38$ \\
OpQ-soft, 3000 episodes (E18, fixed latent) & $0.84\,/\,0.95$ & $0.98\,/\,0.99$ & $0.83\,/\,0.68$ \\
\midrule
\multicolumn{4}{@{}l}{\emph{Time per decision at $t=0$ (ms)}} \\
OpQ-soft & $0.026$ & $0.034$ & $0.030$ \\
MCTS, 100 simulations & $20$ & $21$ & $23$ \\
MCTS, 1000 simulations & $241$ & $241$ & $264$ \\
Exhaustive search & $4$ & $1597$ & $1432$ \\
Exhaustive search, nodes expanded & $5.5\times10^{3}$ & $2.0\times10^{6}$ & $4.9\times10^{5}$ \\
\bottomrule
\end{tabular}
\caption{E19, planners that know what OpQ knows. Each planner knows $K_a$ and uses $\hat R$ (least squares) and $\hat P$ (counts) learned from uniformly random episodes at the fixed training latent; held-out latents are not used for learning. Exhaustive search is evaluated exactly, MCTS by Monte Carlo over mode paths. Means over 5 seeds. Horizons are $H=6$ (C1) and $H=8$ (C2, C3).}
\label{tab:e19-planners}
\end{table}

\subsection{\rev{All controls}}

\rev{Tables~\ref{tab:e19-2x2-fixed} and~\ref{tab:e19-2x2-random} report every agent of the
$2\times2$ study, including the soft and max backups of Linear-Q, the four-update variants
of DQN and DQN\,+\,$K_a$ data, the variant with the true $R$, and the variant restricted to
reachable latents, in the fixed and random regimes.}

\subsection{\rev{Protocol}}

\rev{The environments, regimes, seeds, evaluation and normalisation are those of the E18
protocol (Appendix~\ref{app:expdetails}). New code: \texttt{opq/planners.py},
\texttt{opq/agents\_ext.py}, the class \texttt{DQNAug} in \texttt{opq/deep\_agents.py},
\texttt{experiments/e19\_planners.py}, \texttt{experiments/e19\_horizon.py} and
\texttt{run\_e19.py}, which runs every job in a separate process and can resume.}

\begin{table}[t]
\centering
\small
\setlength{\tabcolsep}{4pt}
\resizebox{\textwidth}{!}{%
\begin{tabular}{@{}llccccccc@{}}
\toprule
 & $H$ & OpQ-soft & Linear-Q & DQN & DQN + $K_a$ & MCTS-100 & MCTS-1000 & Exhaustive \\
\midrule
C1 & 6 & $0.81$ & $\mathbf{0.96}$ & $\mathbf{0.96}$ & $\mathbf{0.96}$ & $0.90$ & $0.94$ & $1.00$ \\
 & 8 & $0.81$ & $0.94$ & $0.94$ & $0.95$ & $0.82$ & $\mathbf{0.96}$ & $1.00$ \\
 & 10 & $0.81$ & $0.94$ & $\mathbf{0.96}$ & $\mathbf{0.96}$ & $0.85$ & $0.95$ & $1.00$ \\
 & 12 & $0.85$ & $0.96$ & $\mathbf{1.00}$ & $\mathbf{1.00}$ & $0.88$ & $0.98$ & -- \\
\midrule
C2 & 8 & $\mathbf{0.98}$ & $0.96$ & $\mathbf{0.98}$ & $\mathbf{0.98}$ & $0.94$ & $\mathbf{0.98}$ & $1.00$ \\
 & 10 & $0.99$ & $0.95$ & $\mathbf{1.00}$ & $\mathbf{1.00}$ & $0.95$ & $0.98$ & -- \\
 & 12 & $0.99$ & $0.94$ & $0.99$ & $\mathbf{1.00}$ & $0.93$ & $0.97$ & -- \\
 & 16 & $0.98$ & $0.88$ & $\mathbf{1.00}$ & $0.98$ & $0.92$ & $0.96$ & -- \\
 & 20 & $0.98$ & $0.91$ & $\mathbf{1.00}$ & $\mathbf{1.00}$ & $0.91$ & $0.96$ & -- \\
\midrule
C3 & 8 & $0.86$ & $0.28$ & $\mathbf{1.00}$ & $\mathbf{1.00}$ & $0.45$ & $0.86$ & $1.00$ \\
 & 10 & $0.61$ & $0.61$ & $\mathbf{1.00}$ & $0.94$ & $0.61$ & $0.61$ & -- \\
 & 12 & $0.73$ & $0.61$ & $\mathbf{1.00}$ & $\mathbf{1.00}$ & $0.86$ & $0.93$ & -- \\
\bottomrule
\end{tabular}}
\caption{E19, longer horizons, training latent. Score $=(V-V_{\mathrm{rand}})/(V_{\mathrm{ref}}-V_{\mathrm{rand}})$, with $V_{\mathrm{ref}}=V^\ast$ where the dynamic program fits in memory (C1 up to $H=10$, C2 at $H=8$) or where $V^\ast=1$ by construction (C3), and otherwise the best value found by any method, which lower-bounds $V^\ast$. OpQ-soft, Linear-Q, DQN and DQN + $K_a$ train for 3000 episodes at the fixed latent (3 seeds). MCTS and exhaustive search use $\hat R$, $\hat P$ from 400 random episodes. Policies are evaluated by Monte Carlo over 32 mode paths (12 for MCTS), so they differ from the exact evaluation of the base-horizon tables by a few hundredths. Bold marks the best score excluding exhaustive search, which attains 1 where it can run.}
\label{tab:e19-horizon}
\end{table}

\begin{table}[t]
\centering
\small
\setlength{\tabcolsep}{4pt}
\resizebox{\textwidth}{!}{%
\begin{tabular}{@{}llccccccc@{}}
\toprule
 & $H$ & OpQ-soft & Linear-Q & DQN & DQN + $K_a$ & MCTS-100 & MCTS-1000 & Exhaustive \\
\midrule
C1 & 6 & $0.98$ & $0.73$ & $0.82$ & $\mathbf{1.00}$ & $0.93$ & $0.97$ & $1.00$ \\
 & 8 & $0.99$ & $0.90$ & $0.86$ & $\mathbf{1.00}$ & $0.95$ & $0.99$ & $1.00$ \\
 & 10 & $0.98$ & $0.86$ & $0.86$ & $\mathbf{0.99}$ & $0.93$ & $0.98$ & $1.00$ \\
 & 12 & $0.99$ & $0.87$ & $0.90$ & $\mathbf{1.00}$ & $0.93$ & $0.96$ & -- \\
\midrule
C2 & 8 & $\mathbf{1.00}$ & $0.82$ & $0.98$ & $\mathbf{1.00}$ & $0.95$ & $0.99$ & $1.00$ \\
 & 10 & $\mathbf{1.00}$ & $0.83$ & $0.98$ & $0.99$ & $0.95$ & $0.99$ & -- \\
 & 12 & $\mathbf{1.00}$ & $0.83$ & $0.97$ & $\mathbf{1.00}$ & $0.94$ & $0.99$ & -- \\
 & 16 & $\mathbf{1.00}$ & $0.81$ & $0.99$ & $\mathbf{1.00}$ & $0.92$ & $0.99$ & -- \\
 & 20 & $0.98$ & $0.85$ & $0.99$ & $\mathbf{1.00}$ & $0.91$ & $0.98$ & -- \\
\midrule
C3 & 8 & $0.84$ & $0.30$ & $0.35$ & $\mathbf{0.90}$ & $0.58$ & $0.78$ & $1.00$ \\
 & 10 & $0.72$ & $0.30$ & $0.42$ & $\mathbf{0.82}$ & $0.64$ & $0.73$ & -- \\
 & 12 & $0.64$ & $0.36$ & $0.61$ & $0.75$ & $0.63$ & $\mathbf{0.77}$ & -- \\
\bottomrule
\end{tabular}}
\caption{E19, longer horizons, zero-shot transfer: score on the first two held-out latents for agents trained at the fixed latent. Normalisation as in Table~\ref{tab:e19-horizon}.}
\label{tab:e19-horizon-heldout}
\end{table}

\rev{\textbf{DQN\,+\,$K_a$ data.} The network, optimiser and replay buffer are those of DQN in
E18. Each minibatch is half real and half synthetic. For a synthetic transition we take
$(t,m,a,m',\mathrm{done})$ from a stored real transition, draw a pure latent
$\rho=N vv^*$ with $v$ uniform on the unit sphere (complex for C3), set
$\rho'=\mathcal K_a(\rho)$ with the true operator, and take the reward $\tau(\hat R(m,a)\rho')$
from the least-squares estimate (the true $R$ in the known-$R$ control). The mode transition
is therefore resampled from the real data, and only the latent is synthetic. The on-support
control instead draws $\rho_t$ by applying $t$ uniformly random actions to the initial latent
(to a random pure state in the random regime).}

\rev{\textbf{Linear-Q.} The value is $Q(t,m,a;\rho)=\theta_{t,m,a}^{\top}\phi(\rho)$, with
$\phi$ the coordinates of $\rho$ in an orthonormal Hermitian basis divided by $N$. The update
is $\theta\leftarrow\theta+\alpha\,\delta\,\phi/(\phi^{\top}\phi+10^{-8})$ with
$\alpha=\max(0.02,n^{-0.6})$, where $\delta$ is the temporal-difference error of the soft
backup with $\beta=5$ (or of the max backup), evaluated at the next latent that the agent
observes.}

\upd{\textbf{Linear-Q\,+\,$K_a$ data.} The linear class is trained like DQN\,+\,$K_a$ data. Each
environment step makes the real normalised-LMS update of Linear-Q, followed by 32 synthetic
soft-backup updates. A synthetic update takes the action and mode from a stored real
transition, draws a fresh pure latent $\rho$, forms $\rho'=\mathcal K_a(\rho)$, and uses the
reward $\tau(\hat R\rho')$ from the same least-squares reward model as OpQ. The next mode is
resampled from the real data. Features, step size and $\beta=5$ are those of Linear-Q.}

\begin{table}[t]
\centering
\small
\setlength{\tabcolsep}{4pt}
\begin{tabular}{@{}llcccccc@{}}
\toprule
 & $H$ & $V_{\mathrm{ref}}$ & cert.$/V_{\mathrm{ref}}$ & nodes & OpQ-soft & MCTS-1000 & Exhaustive \\
\midrule
C1 & 6 & $4.17$ & $1.88$ & $5.5\times10^{3}$ & $0.026$ & $236$ & $4$ \\
 & 8 & $5.25$ & $1.95$ & $8.7\times10^{4}$ & $0.025$ & $264$ & $58$ \\
 & 10 & $6.47$ & $1.93$ & $1.4\times10^{6}$ & $0.024$ & $303$ & $943$ \\
 & 12 & $7.15$$^{\dagger}$ & $2.02$ & $2.2\times10^{7}$ & $0.030$ & $344$ & -- \\
\midrule
C2 & 8 & $6.20$ & $1.32$ & $2.0\times10^{6}$ & $0.034$ & $249$ & $1512$ \\
 & 10 & $7.38$$^{\dagger}$ & $1.33$ & $7.3\times10^{7}$ & $0.034$ & $277$ & -- \\
 & 12 & $8.49$$^{\dagger}$ & $1.33$ & $2.6\times10^{9}$ & $0.036$ & $304$ & -- \\
 & 16 & $10.40$$^{\dagger}$ & $1.33$ & $3.4\times10^{12}$ & $0.035$ & $373$ & -- \\
 & 20 & $11.95$$^{\dagger}$ & $1.33$ & $4.4\times10^{15}$ & $0.035$ & $417$ & -- \\
\midrule
C3 & 8 & $1.00$ & $3.02$ & $4.9\times10^{5}$ & $0.030$ & $264$ & $1374$ \\
 & 10 & $1.00$ & $3.67$ & $1.2\times10^{7}$ & $0.031$ & $351$ & -- \\
 & 12 & $1.00$ & $4.30$ & $3.1\times10^{8}$ & $0.030$ & $330$ & -- \\
\bottomrule
\end{tabular}
\caption{E19, cost and certificate by horizon. $V_{\mathrm{ref}}$ is the reference value at the training latent ($^{\dagger}$: best value found, a lower bound on $V^\ast$). cert.\ is the operator log-sum-exp certificate ($\beta=5$) of Theorem~\ref{thm:certificate}, an upper bound on $V^\ast$. Nodes is the number of node expansions of exhaustive search at $t=0$. Remaining columns give wall-clock time per decision in ms at $t=0$.}
\label{tab:e19-cost}
\end{table}

\begin{table}[t]
\centering
\small
\setlength{\tabcolsep}{4pt}
\resizebox{\textwidth}{!}{%
\begin{tabular}{@{}llcccccc@{}}
\toprule
 & & \multicolumn{2}{c}{C1 random OLDP} & \multicolumn{2}{c}{C2 SO(3) pointing} & \multicolumn{2}{c}{C3 two-qubit} \\
\cmidrule(lr){3-4}\cmidrule(lr){5-6}\cmidrule(lr){7-8}
Agent & $K_a$ & 100 & 3000 & 100 & 3000 & 100 & 3000 \\
\midrule
DQN on $\rho_t$ (nonlinear) & unknown & $\mathbf{0.96}${\scriptsize$\pm$0.02} & $\mathbf{1.00}${\scriptsize$\pm$0.00} & $0.85${\scriptsize$\pm$0.06} & $\mathbf{0.98}${\scriptsize$\pm$0.01} & $-0.08${\scriptsize$\pm$0.16} & $\mathbf{1.00}${\scriptsize$\pm$0.00} \\
DQN + $K_a$ data (nonlinear) & known & $0.91${\scriptsize$\pm$0.08} & $\mathbf{1.00}${\scriptsize$\pm$0.00} & $0.81${\scriptsize$\pm$0.03} & $\mathbf{0.98}${\scriptsize$\pm$0.01} & $0.26${\scriptsize$\pm$0.33} & $\mathbf{1.00}${\scriptsize$\pm$0.00} \\
Linear-Q (linear) & unknown & $0.86${\scriptsize$\pm$0.08} & $0.99${\scriptsize$\pm$0.00} & $0.73${\scriptsize$\pm$0.06} & $0.95${\scriptsize$\pm$0.02} & $0.21${\scriptsize$\pm$0.14} & $0.28${\scriptsize$\pm$0.00} \\
OpQ-soft (linear) & known & $0.77${\scriptsize$\pm$0.06} & $0.84${\scriptsize$\pm$0.01} & $\mathbf{0.94}${\scriptsize$\pm$0.02} & $\mathbf{0.98}${\scriptsize$\pm$0.00} & $\mathbf{0.38}${\scriptsize$\pm$0.20} & $0.83${\scriptsize$\pm$0.08} \\
\midrule
\upd{Linear-Q + $K_a$ data} & \upd{known} & \upd{$0.88${\scriptsize$\pm$0.04}} & \upd{$0.86${\scriptsize$\pm$0.00}} & \upd{$0.93${\scriptsize$\pm$0.02}} & \upd{$0.91${\scriptsize$\pm$0.02}} & \upd{$0.17${\scriptsize$\pm$0.13}} & \upd{$0.73${\scriptsize$\pm$0.24}} \\
Linear-Q, max backup & unknown & $0.86${\scriptsize$\pm$0.08} & $0.99${\scriptsize$\pm$0.00} & $0.75${\scriptsize$\pm$0.06} & $0.97${\scriptsize$\pm$0.00} & $0.28${\scriptsize$\pm$0.00} & $0.46${\scriptsize$\pm$0.27} \\
DQN, 4 updates & unknown & $0.92${\scriptsize$\pm$0.03} & $1.00${\scriptsize$\pm$0.00} & $0.79${\scriptsize$\pm$0.11} & $0.99${\scriptsize$\pm$0.01} & $0.17${\scriptsize$\pm$0.38} & $1.00${\scriptsize$\pm$0.00} \\
DQN + $K_a$ data, 4 updates & known & $0.96${\scriptsize$\pm$0.02} & $1.00${\scriptsize$\pm$0.00} & $0.93${\scriptsize$\pm$0.02} & $0.99${\scriptsize$\pm$0.00} & $0.36${\scriptsize$\pm$0.22} & $1.00${\scriptsize$\pm$0.00} \\
DQN + $K_a$ data, true $R$ & known & $0.96${\scriptsize$\pm$0.04} & $1.00${\scriptsize$\pm$0.00} & $0.84${\scriptsize$\pm$0.05} & $0.99${\scriptsize$\pm$0.01} & $0.24${\scriptsize$\pm$0.32} & $1.00${\scriptsize$\pm$0.00} \\
DQN + $K_a$ data, reachable $\rho$ & known & $0.96${\scriptsize$\pm$0.03} & $1.00${\scriptsize$\pm$0.00} & $0.88${\scriptsize$\pm$0.04} & $0.98${\scriptsize$\pm$0.01} & $0.27${\scriptsize$\pm$0.29} & $1.00${\scriptsize$\pm$0.00} \\
Model + exhaustive & known & $1.00${\scriptsize$\pm$0.00} & $1.00${\scriptsize$\pm$0.00} & $1.00${\scriptsize$\pm$0.00} & $1.00${\scriptsize$\pm$0.00} & $1.00${\scriptsize$\pm$0.00} & $1.00${\scriptsize$\pm$0.00} \\
\bottomrule
\end{tabular}}
\caption{E19, all agents, fixed initial latent. Normalised return after 100 and 3000 episodes; mean $\pm$ s.d.\ over 5 seeds. Bold marks the best of the first four rows.}
\label{tab:e19-2x2-fixed}
\end{table}

\rev{\textbf{Planners.} The model has the true $K_a$, $\hat R$ by least squares and
$\hat P$ by counts from uniformly random episodes at the fixed training latent (the horizon
study uses 400 episodes). Exhaustive search is a vectorised dynamic program over all
$K^{H-t}$ histories; with the true model it equals $V^\ast$. UCT uses exploration constant~1,
uniformly random rollouts, returns normalised by their running minimum and maximum, and
chance nodes over the pair (action, next mode). Policies of the exhaustive planner are
evaluated exactly. Policies that include search are evaluated by Monte Carlo over sampled
mode paths with common random numbers across methods (32 paths, 12 for UCT). Decision time
is wall-clock time of the first decision with an empty cache, on one CPU core.}

\rev{\textbf{Longer horizons.} We use $H\in\{6,8,10,12\}$ on C1, $\{8,10,12,16,20\}$ on C2
and $\{8,10,12\}$ on C3, with 3 seeds. On C3 the target is a word of length $H$, so
$V^\ast=1$. Otherwise $V_{\mathrm{ref}}=V^\ast$ when the dynamic program fits in
$4\times10^{8}$ bytes, and the best value found by any method when it does not; the latter
lower-bounds $V^\ast$, so scores at those horizons can be slightly optimistic and can exceed
1 through sampling alone. The operator log-sum-exp certificate is computed at
$\beta\in\{2,5\}$ and reported at $\beta=5$. As a conditioning diagnostic for
$\sum_i e^{\beta x_i}$ we record the product of $\beta$ and the spectral spread of the
operators; it stays below 25 in all runs, the threshold under which we treat double
precision as reliable.}

\begin{table}[t]
\centering
\small
\setlength{\tabcolsep}{4pt}
\resizebox{\textwidth}{!}{%
\begin{tabular}{@{}llcccccc@{}}
\toprule
 & & \multicolumn{2}{c}{C1 random OLDP} & \multicolumn{2}{c}{C2 SO(3) pointing} & \multicolumn{2}{c}{C3 two-qubit} \\
\cmidrule(lr){3-4}\cmidrule(lr){5-6}\cmidrule(lr){7-8}
Agent & $K_a$ & 100 & 3000 & 100 & 3000 & 100 & 3000 \\
\midrule
DQN on $\rho_t$ (nonlinear) & unknown & $0.89${\scriptsize$\pm$0.03} & $\mathbf{0.99}${\scriptsize$\pm$0.00} & $0.73${\scriptsize$\pm$0.04} & $\mathbf{0.99}${\scriptsize$\pm$0.00} & $0.04${\scriptsize$\pm$0.12} & $\mathbf{0.87}${\scriptsize$\pm$0.02} \\
DQN + $K_a$ data (nonlinear) & known & $0.93${\scriptsize$\pm$0.03} & $\mathbf{0.99}${\scriptsize$\pm$0.00} & $0.82${\scriptsize$\pm$0.03} & $\mathbf{0.99}${\scriptsize$\pm$0.00} & $0.22${\scriptsize$\pm$0.15} & $0.86${\scriptsize$\pm$0.02} \\
Linear-Q (linear) & unknown & $0.77${\scriptsize$\pm$0.06} & $0.96${\scriptsize$\pm$0.02} & $0.57${\scriptsize$\pm$0.04} & $0.95${\scriptsize$\pm$0.01} & $0.25${\scriptsize$\pm$0.10} & $0.54${\scriptsize$\pm$0.08} \\
OpQ-soft (linear) & known & $\mathbf{0.95}${\scriptsize$\pm$0.01} & $0.95${\scriptsize$\pm$0.00} & $\mathbf{0.95}${\scriptsize$\pm$0.01} & $\mathbf{0.99}${\scriptsize$\pm$0.00} & $\mathbf{0.58}${\scriptsize$\pm$0.07} & $0.68${\scriptsize$\pm$0.00} \\
\midrule
\upd{Linear-Q + $K_a$ data} & \upd{known} & \upd{$0.96${\scriptsize$\pm$0.01}} & \upd{$0.97${\scriptsize$\pm$0.00}} & \upd{$0.91${\scriptsize$\pm$0.02}} & \upd{$0.97${\scriptsize$\pm$0.01}} & \upd{$0.41${\scriptsize$\pm$0.11}} & \upd{$0.69${\scriptsize$\pm$0.01}} \\
Linear-Q, max backup & unknown & $0.77${\scriptsize$\pm$0.05} & $0.97${\scriptsize$\pm$0.01} & $0.62${\scriptsize$\pm$0.04} & $0.96${\scriptsize$\pm$0.01} & $0.29${\scriptsize$\pm$0.07} & $0.53${\scriptsize$\pm$0.07} \\
DQN, 4 updates & unknown & $0.89${\scriptsize$\pm$0.04} & $0.99${\scriptsize$\pm$0.00} & $0.82${\scriptsize$\pm$0.04} & $0.99${\scriptsize$\pm$0.00} & $0.04${\scriptsize$\pm$0.11} & $0.87${\scriptsize$\pm$0.03} \\
DQN + $K_a$ data, 4 updates & known & $0.96${\scriptsize$\pm$0.02} & $1.00${\scriptsize$\pm$0.00} & $0.90${\scriptsize$\pm$0.03} & $1.00${\scriptsize$\pm$0.00} & $0.38${\scriptsize$\pm$0.09} & $0.90${\scriptsize$\pm$0.01} \\
DQN + $K_a$ data, true $R$ & known & $0.94${\scriptsize$\pm$0.02} & $0.99${\scriptsize$\pm$0.00} & $0.81${\scriptsize$\pm$0.01} & $0.99${\scriptsize$\pm$0.00} & $0.19${\scriptsize$\pm$0.14} & $0.89${\scriptsize$\pm$0.02} \\
DQN + $K_a$ data, reachable $\rho$ & known & $0.94${\scriptsize$\pm$0.02} & $1.00${\scriptsize$\pm$0.00} & $0.82${\scriptsize$\pm$0.04} & $0.99${\scriptsize$\pm$0.00} & $0.19${\scriptsize$\pm$0.14} & $0.87${\scriptsize$\pm$0.03} \\
Model + exhaustive & known & $1.00${\scriptsize$\pm$0.00} & $1.00${\scriptsize$\pm$0.00} & $1.00${\scriptsize$\pm$0.00} & $1.00${\scriptsize$\pm$0.00} & $1.00${\scriptsize$\pm$0.00} & $1.00${\scriptsize$\pm$0.00} \\
\bottomrule
\end{tabular}}
\caption{E19, all agents, random initial latent. Normalised return after 100 and 3000 episodes; mean $\pm$ s.d.\ over 5 seeds. Bold marks the best of the first four rows.}
\label{tab:e19-2x2-random}
\end{table}

\upd{\textbf{Linear class with operator data.} With random latents, Linear-Q\,+\,$K_a$ data scores
0.96/0.91/0.41 after 100 episodes and 0.97/0.97/0.69 after 3000 on C1/C2/C3 (OpQ-soft:
0.95/0.95/0.58 and 0.95/0.99/0.68). With a fixed latent it scores 0.88/0.93/0.17 and
0.86/0.91/0.73 (OpQ-soft: 0.77/0.94/0.38 and 0.84/0.98/0.83). At held-out latents it scores
0.97/0.97/0.71 (OpQ-soft 0.95/0.99/0.68; DQN\,+\,$K_a$ data 0.99/0.98/0.84). Operator-generated
data therefore closes most of the early gap between the linear class and OpQ on C1 and C2, and
not on C3.}

\begin{figure}[t]
\centering
\includegraphics[width=0.95\linewidth]{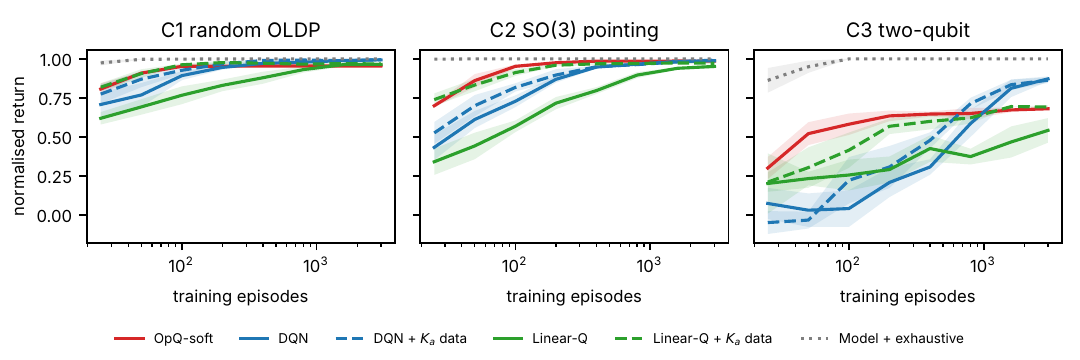}\\[4pt]
\includegraphics[width=0.95\linewidth]{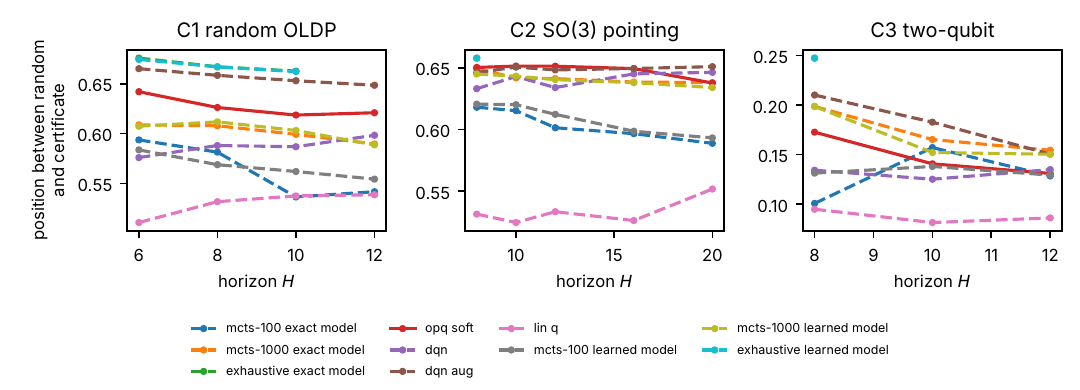}
\caption{\upd{E19, learning curves with random initial latents: normalised return against training
episodes (mean $\pm$ s.d.\ over 5 seeds); and value against horizon, as the position between the
random policy and the certificate ($\beta=5$) at the training latents.}}
\label{fig:e19-app}
\end{figure}

\section{\rev{E20: Tightening the Single-Operator Certificate}}\label{app:e20}

\rev{E20 uses the exact model of each environment of E18 (\texttt{opq/envs.py}, seed~0, base
horizons) and the same 8 held-out initial latents. No learning is involved, so every number
is the exact value of a greedy policy, normalised as in E18. Code:
\texttt{experiments/e20\_tightening.py}; one CPU core, about six minutes in total.}

\begin{table}[t]
\centering\small\setlength{\tabcolsep}{4pt}
\resizebox{\textwidth}{!}{%
\begin{tabular}{@{}lcccccc@{}}
\toprule
 & \multicolumn{2}{c}{C1 random OLDP} & \multicolumn{2}{c}{C2 SO(3) pointing} & \multicolumn{2}{c}{C3 two-qubit} \\
\cmidrule(lr){2-3}\cmidrule(lr){4-5}\cmidrule(lr){6-7}
Aggregator & score & cert./$V^\ast$ & score & cert./$V^\ast$ & score & cert./$V^\ast$ \\
\midrule
$\LSE_\beta$, $\beta=5$ (OpQ) & 0.86 / 0.95 & 1.88 / 1.44 & 0.99 / 0.99 & 1.32 / 1.30 & 0.79 / 0.68 & 3.02 / 3.23 \\
$\LSE_\beta$, $\beta=20$      & 0.87 / 0.97 & 2.00 / 1.51 & 0.98 / 0.99 & 1.13 / 1.12 & 0.79 / 0.68 & 1.50 / 1.62 \\
trace-minimal SDP (Cor.~\ref{cor:anybound}) & 0.84 / 0.96 & 1.67 / 1.27 & 0.87 / 0.98 & 1.64 / 1.58 & 0.28 / 0.64 & 1.79 / 1.83 \\
\bottomrule
\end{tabular}}
\caption{E20(a): greedy control and certificate for three L\"owner upper-bound aggregators, at
the training latent / averaged over the 8 held-out latents. The first row reproduces
Table~\ref{tab:e18ref}. The SDP is solved with SCS (tolerance $10^{-9}$); its largest
constraint violation is below $4\times10^{-9}$ and is removed by a shift of the identity, so
the certificate remains valid.}
\label{tab:e20a}
\end{table}

\begin{table}[t]
\centering\small\setlength{\tabcolsep}{5pt}
\begin{tabular}{@{}lccccccc@{}}
\toprule
$M$ & 1 & 2 & 4 & 8 & 16 & 32 & 64 \\
\midrule
C1 training latent & 0.89 & 0.86 & 0.96 & 0.98 & 0.97 & 1.00 & 1.00 \\
C1 held-out        & 0.78 & 0.85 & 0.90 & 0.94 & 0.98 & 0.99 & 1.00 \\
C2 training latent & 0.83 & 0.84 & 0.81 & 0.93 & 0.95 & 0.98 & 0.98 \\
C2 held-out        & 0.71 & 0.80 & 0.78 & 0.84 & 0.89 & 0.95 & 0.98 \\
C3 training latent & 0.53 & 0.77 & 0.67 & 1.00 & 1.00 & 1.00 & 1.00 \\
C3 held-out        & 0.64 & 0.73 & 0.78 & 0.82 & 0.88 & 0.92 & 0.95 \\
\bottomrule
\end{tabular}
\caption{E20(b): greedy score with $M$ operators per $(t,m)$, obtained by point-based backups
at $M$ latents reachable from the training latent (training rows) or from 64 random pure
latents (held-out rows); mean over 5 draws of the points. For comparison, one $\LSE_\beta$
operator scores $0.86/0.95$, $0.99/0.99$ and $0.79/0.68$ on C1, C2 and C3.}
\label{tab:e20b}
\end{table}

\rev{\textbf{Protocol.} (a) The aggregators replace $\LSE_\beta$ in the exact finite-horizon
recursion~(4); the SDP minimises $\tau(Y)$ subject to $Y\succeq Q(s',a')$ for all $a'$. (b) For
each $t$ we draw $M$ latents by applying $t$ uniformly random actions to the training latent
(or to one of 64 random pure latents), back up one operator per point and mode by the operator
analogue of a point-based Bellman backup, and act greedily on the one-step lookahead
$\tau(R\rho')+\gamma\sum_{m'}P\max_{\Gamma}\tau(\rho'\Gamma)$. For finite horizons each
backed-up operator is the value of a conditional plan, so the family is a lower bound that
becomes exact once it contains the maximising plan at every reachable node. (c)
$q^\ast_t$ is computed by exhaustive search from every node reached by the greedy policy; the
spread is weighted by the probability of reaching the node. (d) Exact evaluation (the
environment is deterministic).}

\begin{table}[t]
\centering\small\setlength{\tabcolsep}{5pt}
\begin{tabular}{@{}lcccccc@{}}
\toprule
 & $V^\ast$ & loss $V^\ast-V^{\hat\pi}$ & bound of Prop.~\ref{prop:nearopt} & mean spread & mean spread$/V^\ast$ & min.\ slack \\
\midrule
C1 training latent & 4.17 & 0.581 & 3.544 & 0.645 & 15.5\% & $-3\cdot10^{-16}$ \\
C1 held-out (mean) & 5.98 & 0.275 & 2.767 & 0.511 & 8.6\%  & $-7\cdot10^{-16}$ \\
C2 training latent & 6.20 & 0.041 & 0.847 & 0.121 & 2.0\%  & $-3\cdot10^{-16}$ \\
C2 held-out (mean) & 6.06 & 0.049 & 1.137 & 0.157 & 2.7\%  & $-3\cdot10^{-16}$ \\
C3 training latent & 1.00 & 0.146 & 0.666 & 0.083 & 8.3\%  & $0$ \\
C3 held-out (mean) & 0.92 & 0.217 & 0.678 & 0.085 & 9.2\%  & $-1\cdot10^{-16}$ \\
\bottomrule
\end{tabular}
\caption{E20(c): slack of the $\LSE_\beta$ certificate ($\beta=5$) on the nodes reached by its
greedy policy. The bound of Proposition~\ref{prop:nearopt} holds in every case; the minimum
slack is zero up to rounding, as Theorem~\ref{thm:certificate} requires.}
\label{tab:e20c}
\end{table}

\begin{table}[t]
\centering\small\setlength{\tabcolsep}{5pt}
\begin{tabular}{@{}lccccc@{}}
\toprule
 & certificate greedy & depth~3, zero leaf & depth~3, cert.\ leaf & depth~5, zero leaf & depth~5, cert.\ leaf \\
$H$ & & & & & \\
\midrule
8  & 0.85 / 0.73 & 0.85 / 0.80 & 0.85 / 0.84 & 1.00 / 0.84 & 1.00 / 0.87 \\
10 & 0.73 / 0.74 & 0.73 / 0.77 & 0.85 / 0.85 & 0.85 / 0.83 & 0.86 / 0.90 \\
12 & 0.85 / 0.80 & 0.85 / 0.72 & 0.85 / 0.85 & 1.00 / 0.87 & 1.00 / 0.92 \\
\bottomrule
\end{tabular}
\caption{E20(d): two-qubit preparation with the target a word of length $H$ (the E19
protocol), true model. Expected terminal fidelity at $|00\rangle$ / averaged over 4 held-out
initial states. A depth-$d$ search enumerates all $K^d$ continuations and, at its frontier,
uses either zero or the certificate $\max_a\tau(\rho Q^\beta_t(s,a))$ ($\beta=5$).}
\label{tab:e20d}
\end{table}

\section{\rev{Scope of the Guarantees and of the Multi-Operator Extension}}\label{app:scope}

\subsection{\rev{What is proved, and under which conditions}}

\rev{Table~\ref{tab:scope} separates the theoretical statements of the paper. In particular,
the name ``operator Q-learning'' should be read with these qualifications: the sampled control
update converges for a finite horizon (Theorem~\ref{thm:sampled}), but to the certificate
$Q^\beta$ rather than to $Q^\ast$, and no convergence is proved for an infinite horizon
outside the contraction region.}

\begin{table}[t]
\centering\scriptsize\setlength{\tabcolsep}{3pt}\renewcommand{\arraystretch}{1.2}
\begin{tabular}{@{}p{0.2\textwidth}p{0.42\textwidth}p{0.34\textwidth}@{}}
\toprule
Statement & What is proved & Conditions and limits \\
\midrule
Policy evaluation (Prop.~\ref{thm:exact_rep}, Thm.~\ref{thm:td})
 & Exact operator representation of latent-blind values; operator TD converges a.s.
 & Fixed policy; $\gamma\kappa<1$; full-support sampling; Robbins--Monro steps. \\
Policy improvement (Props.~\ref{prop:improve}, \ref{prop:robust})
 & Latent-aware greedy improvement over a latent-blind base; robust to $\varepsilon$-accurate estimates.
 & One improvement step (OpQ-PI); no iteration guarantee. \\
Finite-horizon certificate (Thm.~\ref{thm:certificate}, Cor.~\ref{cor:anybound})
 & $q^\pi_t\le\tau(\rho Q^\beta_t)$ for every policy, latent-aware included.
 & Exact model; CP; any L\"owner upper-bound aggregator. \\
Infinite-horizon certificate (Thm.~\ref{thm:certificate})
 & $n$-step truncations are certificates up to $\gamma^nR_{\max}\tau(\rho)/(1-\gamma)$; the fixed point is one if $\mathcal T^\beta$ contracts.
 & Contraction holds on the commuting sector; in generality needs $\gamma e^{2\beta M_\beta}<1$. \\
Sampled soft update, finite horizon (OpQ-soft; Thm.~\ref{thm:sampled})
 & \rev{Converges a.s.\ to the certificate $Q^\beta$ at every reachable entry.}
 & \rev{Robbins--Monro steps; every reachable entry updated infinitely often; $\hat R\to R$. Limit is $Q^\beta$, not $Q^\ast$. E18--E23 use a step floor (E24 measures its effect; E22: no divergence in 108 runs).} \\
Sampled soft update, infinite horizon
 & \rev{No convergence result.}
 & \rev{$\LSE_\beta$ is not monotone or non-expansive (Props.~\ref{prop:lse_lipschitz_new}, \ref{prop:monotone_new}); contraction needs $\gamma e^{2\beta M_\beta}<1$.} \\
Greedy control (Prop.~\ref{prop:nearopt})
 & Loss $\le\sum_t\gamma^t\varepsilon_t$ (slack spread).
 & Sufficient, not necessary; $\varepsilon_t$ needs the model; conservative by $5$--$21\times$ (E20). \\
$M$ operators per state (App.~\ref{app:scope})
 & Each operator is the value of a conditional plan, so the family is a lower bound on $V^\ast$; exact once it contains the maximising plan at every reachable node.
 & Exact-model construction only; no guarantee for a sampled version. \\
\rev{Hidden latent (Prop.~\ref{thm:exact_rep}; E23)}
 & \rev{Plugging the belief mean into $\tau(\rho Q)$ gives the exact expected return of policies of the action and mode history.}
 & \rev{Known operators and prior mean; excludes policies that condition on past rewards.} \\
Representation size
 & Memory $|S|KN^2$; update $O(N^3)$; independent of the horizon.
 & \rev{$N=2^n$ for $n$ qubits; experiments use $N\le4$; no compressed parameterisation is studied.} \\
\bottomrule
\end{tabular}
\caption{Theoretical statements, what each establishes, and its conditions.}
\label{tab:scope}
\end{table}

\subsection{\rev{One operator or several}}

\rev{\textbf{When one operator suffices.} On the commuting sector the soft recursion is soft
$Q$-learning and its slack vanishes at pure eigenstates as $\beta\to\infty$
(Prop.~\ref{prop:reduction}). More generally, Proposition~\ref{prop:nearopt} shows that one
operator gives near-optimal greedy control when its slack varies little across actions. This
is the case on SO(3) pointing, where the mean spread is $2\%$ of $V^\ast$ and the single
operator scores $0.99$, better than point-based sets up to $M=64$
(Table~\ref{tab:e20b}).}

\rev{\textbf{When several are needed.} For a finite horizon, $q^\ast_t(s,\cdot,a)$ is the
maximum over conditional plans of functions linear in $\rho$, and one operator cannot
represent a maximum whose maximiser changes with the latent. This is the situation on the
random OLDP and on two-qubit preparation, where the slack spread is $8$--$15\%$ of $V^\ast$
and the single operator scores $0.86$ and $0.79$.}

\rev{\textbf{Relation to point-based methods.} The $M$-operator family is point-based value
iteration \citep{pineau2003point} written with operators, and we do not claim it as new. What
the operator setting changes is the upper side. Point-based POMDP solvers that bracket
$V^\ast$, such as HSVI \citep{smith2004heuristic}, initialise the upper bound from the
fast-informed bound, a lattice join; no such join exists here (Remark~\ref{rem:pomdp}), and
Theorem~\ref{thm:certificate} (or any aggregator of Corollary~\ref{cor:anybound}) supplies it
instead.}

\rev{\textbf{Lower bound, and how to choose $M$.} In the point-based backup of
Appendix~\ref{app:e20}, every stored operator has the form
$\Gamma=\mathcal K_a^{\dagger}\bigl(R(m,a)+\gamma\sum_{m'}P(m'\mid m,a)\Gamma_{m'}\bigr)$ with
each $\Gamma_{m'}$ stored at the next step. By induction from $\Gamma_H=0$ and
Proposition~\ref{thm:exact_rep}, $\Gamma$ is the value operator of the conditional plan
``take $a$, then follow the plan of $\Gamma_{m'}$ in the observed mode $m'$''. Such a plan is
a valid policy, so $\tau(\rho\Gamma)\le V^\ast_t(m,\rho)$ for every $\rho$ and every stored
$\Gamma$, and the family gives a lower bound $L_M(\rho)=\max_\Gamma\tau(\rho\Gamma)$. With the
certificate this gives a model-based sandwich
\[
L_M(\rho)\;\le\;V^\ast_0(\rho)\;\le\;\max_a\tau\bigl(\rho\,Q^\beta_0(s_0,a)\bigr),
\]
computable without $V^\ast$. We choose $M$ by doubling it until $L_M$ at the latents of
interest stops increasing (Table~\ref{tab:e21c}). The upper side is loose (Table~\ref{tab:e20a}),
so the sandwich certifies progress in $M$ rather than closeness to $V^\ast$.}

\rev{\textbf{Cost.} The construction backs up $M$ points per $(t,m)$; each backup reads out
$M\,M_{\mathrm{modes}}$ stored operators per next mode and applies one dual map, for
$O\bigl(HM_{\mathrm{modes}}M\,K(M_{\mathrm{modes}}M N^2+N^3)\bigr)$ time, quadratic in $M$,
and $HM_{\mathrm{modes}}MN^2$ stored numbers. A decision reads out
$O(KM_{\mathrm{modes}}M)$ operators instead of $K$. Table~\ref{tab:e21b} gives measured times.}

\section{\rev{E21: Miscalibrated Operators, and the Cost of $M$ Operators}}\label{app:e21}

\rev{\textbf{Protocol.} The agent's model replaces each action unitary by
$\tilde U_a=U_a\exp(\varepsilon A_a)$, with $A_a$ a random anti-Hermitian (real
antisymmetric on C1, C2) matrix of unit Frobenius norm, 5 draws per $\varepsilon$; $P$ and
$R$ are exact. OpQ is the greedy readout of the exact soft recursion~(4) on the model
($\beta=5$), and the planner is exhaustive search on the model. Both are evaluated exactly on
the true environment. In the observed case the agent sees the true latent; in the tracked
case it reconstructs the latent from $\rho_0$ and its own actions with $\tilde U_a$, as an
agent without access to the latent would. Code: \texttt{experiments/e21\_calibration.py}.}

\rev{\textbf{Results.} On SO(3) pointing OpQ is as robust as planning on the same model: with
a tracked latent both keep $\ge0.95$ up to $\varepsilon=0.1$ and fall together at
$\varepsilon=0.3$ ($0.73$ and $0.74$), where the tracking error dominates. On the random OLDP
and on two-qubit preparation the single-operator greedy policy is markedly less robust than
planning at the training latent (C3: $0.79$ to $0.49$ at $\varepsilon=0.03$, while planning
stays at $1.00$), although its held-out scores are stable up to $\varepsilon=0.1$. This is
consistent with Proposition~\ref{prop:nearopt}: where the certificate's slack varies across
actions, greedy decisions are close to ties, and small model errors flip them. Uncertain gate
sets are therefore a limitation of the single-operator method on such tasks.}

\begin{table}[t]
\centering\small\setlength{\tabcolsep}{5pt}
\begin{tabular}{@{}llcccc@{}}
\toprule
 & $\varepsilon$ & OpQ, observed & OpQ, tracked & exhaustive, observed & exhaustive, tracked \\
\midrule
C1 random OLDP & 0    & 0.86 / 0.97 & 0.86 / 0.97 & 1.00 / 1.00 & 1.00 / 1.00 \\
 & 0.01 & 0.77 / 0.97 & 0.77 / 0.97 & 1.00 / 1.00 & 1.00 / 1.00 \\
 & 0.03 & 0.77 / 0.97 & 0.80 / 0.97 & 1.00 / 1.00 & 1.00 / 1.00 \\
 & 0.1  & 0.80 / 0.97 & 0.82 / 0.96 & 0.99 / 1.00 & 1.00 / 0.98 \\
 & 0.3  & 0.82 / 0.94 & 0.79 / 0.85 & 0.93 / 0.94 & 0.95 / 0.88 \\
\midrule
C2 SO(3) pointing & 0    & 0.99 / 1.00 & 0.99 / 1.00 & 1.00 / 1.00 & 1.00 / 1.00 \\
 & 0.01 & 0.99 / 1.00 & 0.99 / 1.00 & 1.00 / 1.00 & 1.00 / 1.00 \\
 & 0.03 & 0.99 / 1.00 & 0.98 / 0.99 & 1.00 / 1.00 & 0.99 / 1.00 \\
 & 0.1  & 0.98 / 0.99 & 0.96 / 0.97 & 0.99 / 0.99 & 0.95 / 0.97 \\
 & 0.3  & 0.94 / 0.96 & 0.73 / 0.77 & 0.93 / 0.95 & 0.74 / 0.77 \\
\midrule
C3 two-qubit & 0    & 0.79 / 0.71 & 0.79 / 0.71 & 1.00 / 1.00 & 1.00 / 1.00 \\
 & 0.01 & 0.79 / 0.71 & 0.79 / 0.71 & 1.00 / 1.00 & 1.00 / 1.00 \\
 & 0.03 & 0.59 / 0.71 & 0.49 / 0.73 & 1.00 / 1.00 & 1.00 / 1.00 \\
 & 0.1  & 0.63 / 0.70 & 0.43 / 0.72 & 1.00 / 0.99 & 1.00 / 0.98 \\
 & 0.3  & 0.43 / 0.62 & 0.70 / 0.54 & 0.87 / 0.80 & 0.87 / 0.80 \\
\bottomrule
\end{tabular}
\caption{E21(a): normalised return on the true environment of policies built from a
miscalibrated model, at the training latent / averaged over 4 held-out latents (mean over 5
draws of the perturbation).}
\label{tab:e21a}
\end{table}

\begin{table}[t]
\centering\scriptsize\setlength{\tabcolsep}{3.5pt}
\resizebox{\textwidth}{!}{%
\begin{tabular}{@{}lccccccccc@{}}
\toprule
 & $V^\ast$ & certificate & $L_1$ & $L_2$ & $L_4$ & $L_8$ & $L_{16}$ & $L_{32}$ & $L_{64}$ \\
\midrule
C1 random OLDP, training latent & 4.17 & 7.83 & 2.92 & 3.28 & 3.63 & 3.91 & 3.95 & 4.13 & 4.15 \\
C1 random OLDP, held-out        & 5.89 & 8.47 & 0.91 & 2.42 & 3.62 & 3.93 & 5.09 & 5.42 & 5.70 \\
C2 SO(3) pointing, training latent & 6.20 & 8.18 & 4.66 & 4.94 & 5.02 & 5.74 & 5.96 & 6.09 & 6.13 \\
C2 SO(3) pointing, held-out     & 6.08 & 7.87 & 2.61 & 3.21 & 3.49 & 3.90 & 5.07 & 5.30 & 5.60 \\
C3 two-qubit, training latent   & 1.00 & 3.02 & 0.67 & 0.84 & 0.77 & 1.00 & 1.00 & 1.00 & 1.00 \\
C3 two-qubit, held-out          & 0.92 & 2.99 & 0.24 & 0.39 & 0.50 & 0.54 & 0.66 & 0.65 & 0.75 \\
\bottomrule
\end{tabular}}
\caption{E21(c): the sandwich of Appendix~\ref{app:scope}. $L_M$ is the lower bound from $M$
operators per state (mean over 5 draws of the points; held-out values averaged over 4
latents), the certificate is the $\LSE_\beta$ upper bound ($\beta=5$). $L_M\le V^\ast$ in
every case.}
\label{tab:e21c}
\end{table}

\begin{table}[t]
\centering\small\setlength{\tabcolsep}{5pt}
\begin{tabular}{@{}rcccccc@{}}
\toprule
 & \multicolumn{2}{c}{C1} & \multicolumn{2}{c}{C2} & \multicolumn{2}{c}{C3} \\
\cmidrule(lr){2-3}\cmidrule(lr){4-5}\cmidrule(lr){6-7}
$M$ & build(s)/decision(ms) & numbers & build(s)/decision(ms) & numbers & build(s)/decision(ms) & numbers \\
\midrule
1  & 0.001 / 0.11 & 108  & 0.003 / 0.07 & 144  & 0.006 / 0.06 & 1152 \\
8  & 0.018 / 0.50 & 864  & 0.048 / 0.24 & 1152 & 0.093 / 0.14 & 9216 \\
32 & 0.196 / 1.16 & 3456 & 0.560 / 0.85 & 4608 & 1.031 / 0.97 & 36864 \\
64 & 0.736 / 1.92 & 6912 & 1.662 / 1.72 & 9216 & 3.858 / 0.82 & 73728 \\
\bottomrule
\end{tabular}
\caption{E21(b): cost of $M$ operators per $(t,m)$ on one CPU core: construction time, time
of the first decision, and stored numbers. $M=1$ is the cost of a point-based single operator;
the $\LSE$ operator of OpQ has the same size.}
\label{tab:e21b}
\end{table}

\section{\rev{E22: Sensitivity of the Sampled Soft Update to $\beta$}}\label{app:e22}

\rev{Theorem~\ref{thm:sampled} gives convergence for Robbins--Monro steps but says nothing
about how the limit depends on $\beta$, or about the step-size floor used in practice. E22
measures this: OpQ-soft with learned $R$, $\beta\in\{0.5,1,2,5,10,20\}$, C1--C3 in both
regimes, 3 seeds, 3000 episodes, otherwise as in E18. For each run we record the score, the
score of the exact-model readout at the same $\beta$, the largest operator norm reached, the
relative change of the whole table $Q$ between episodes 2500 and 3000 (drift), and the
fraction of 600 probe nodes $(t,m,\rho)$, drawn by random rollouts from the evaluation
latents, whose greedy action changes between those checkpoints (switching). Code:
\texttt{experiments/e22\_beta\_sweep.py}.}

\begin{table}[t]
\centering\scriptsize\setlength{\tabcolsep}{3pt}
\resizebox{\textwidth}{!}{%
\begin{tabular}{@{}lcccccc@{}}
\toprule
$\beta=$ & 0.5 & 1 & 2 & 5 & 10 & 20 \\
\midrule
C1 fixed  & 0.86/0.86/0\% & 0.79/0.76/0\% & 0.74/0.76/0\%  & 0.84/0.86/0\% & 0.87/0.87/0\% & 0.87/0.87/1\% \\
C1 random & 0.96/0.96/1\% & 0.97/0.97/2\% & 0.96/0.95/1\%  & 0.95/0.95/0\% & 0.96/0.96/0\% & 0.97/0.97/1\% \\
C2 fixed  & 0.88/0.88/5\% & 0.91/0.88/7\% & 0.94/0.99/14\% & 0.98/0.99/3\% & 0.98/0.98/1\% & 0.98/0.98/5\% \\
C2 random & 0.97/0.98/2\% & 0.98/0.98/2\% & 0.98/0.99/2\%  & 0.99/0.99/2\% & 0.99/0.99/0\% & 0.99/0.99/0\% \\
C3 fixed  & 0.28/0.28/0\% & 0.28/0.28/0\% & 0.62/0.79/1\%  & 0.86/0.79/5\% & 0.45/0.79/6\% & 0.45/0.79/0\% \\
C3 random & 0.70/0.70/0\% & 0.70/0.70/0\% & 0.69/0.69/0\%  & 0.68/0.68/0\% & 0.68/0.68/1\% & 0.65/0.68/1\% \\
\bottomrule
\end{tabular}}
\caption{E22: learned score after 3000 episodes / exact-model score at the same $\beta$ /
late-training switching, mean over 3 seeds.}
\label{tab:e22}
\end{table}

\rev{\textbf{Results.} No run diverged: all 108 runs stayed finite, the largest operator norm
was 25.2 and the largest drift 0.010. The learned score is within $0.03$ of the exact-model
score at the same $\beta$ in 30 of 36 settings. The exceptions are two-qubit preparation from
a fixed latent, where reward arrives only at the last step and some seeds stay on a plateau of
$0.28$ for $\beta\in\{2,10,20\}$, and SO(3) pointing from a fixed latent at $\beta=2$.
Switching is at most $6\%$ of probe nodes in 34 of 36 settings ($7\%$ and $14\%$ for SO(3)
fixed at $\beta=1$ and $2$), so the failure of L\"owner monotonicity does not produce
persistent policy chattering here. The score also depends on $\beta$ through the exact
recursion itself (e.g.\ C1 fixed: $0.86$ at $\beta=0.5$, $0.76$ at $\beta=1$--$2$, $0.87$ at
$\beta\ge10$), so $\beta$ is a modelling choice as well as a learning parameter; $\beta=5$ is
within $0.02$ of the best exact score on C1 and C2.}

\section{\rev{E23: Control with a Hidden Initial Latent}}\label{app:e23}

\rev{\textbf{Setting.} The initial latent is drawn from a prior around the nominal state of
each environment, $v=(v_{\mathrm{nom}}+\sigma g)/\lVert v_{\mathrm{nom}}+\sigma g\rVert$,
$\rho_0=N vv^*$, $g\sim\mathcal N(0,I)$ (complex on C3), $\sigma=0.5$, and is \emph{never
observed}: agents see the time, the mode and the rewards. They know the operators and the
prior mean $\bar\rho_0$ (a calibration constant, estimated once from 20,000 prior draws), and
track the belief mean $\bar\rho_{t+1}=\mathcal K_{a_t}(\bar\rho_t)$ from their own actions.
Because rewards are linear in $\rho_t$, the expected return over the prior of any policy of
$(t,m)$ and the history of actions and modes equals its value at $\bar\rho_0$
(Proposition~\ref{thm:exact_rep}); the optimal such policy (the \emph{blind optimum}) is
therefore computed exactly by dynamic programming at $\bar\rho_0$, and every agent is
evaluated exactly at $\bar\rho_0$. A Monte Carlo check over 400 hidden draws confirms this
(Table~\ref{tab:e23ref}). This policy class does not condition on past rewards. On C1 and C2
per-step rewards depend on $\rho_0$, so a policy that infers $\rho_0$ from them by Bayesian
filtering could exceed the blind optimum; the latent-aware oracle bounds what such inference
could gain. On C3 the reward arrives only at the last step, so no reward is available for
inference before the final action, as in the one-copy quantum setting. \emph{Plug-in} agents
are trained with the latent observed, under the same prior, and deployed on the belief mean;
for operator values plugging in the mean is exact,
$\mathbb E\,\tau(\rho Q)=\tau(\bar\rho Q)$, while for a nonlinear $Q$-network it is not.
Score $=(V-V_{\mathrm{rand}})/(V^\ast_{\mathrm{blind}}-V_{\mathrm{rand}})$; 3 seeds, 3000
episodes. Code: \texttt{experiments/e23\_hidden\_latent.py}.}

\rev{\textbf{Results.} (i) With the reward observable known (natural when the target state or
orientation is given), OpQ learns fastest on C2 and C3 and is level on C1: after 100 episodes
it scores $0.84/0.97/0.53$ on C1/C2/C3, against $0.85/0.64/-0.06$ for the DQN on the belief
mean, and reaches $0.98$ on SO(3) pointing. On two-qubit preparation it stays at its
single-operator ceiling ($0.79$), and the belief-mean DQN reaches $1.00$. (ii) With matched
knowledge (OpQ learning $R$ from rewards and the belief mean), the belief-mean DQN learns
faster early on C1 and C2 ($0.85/0.64$ vs.\ $0.30/0.07$ at 100 episodes); at 3000 episodes OpQ
is ahead on C2 ($0.97$ vs.\ $0.91$), level on C1 ($0.96$ vs.\ $0.97$), and far behind on C3
($0.46$ vs.\ $1.00$): estimating $R$ from belief means is harder than from observed latents.
(iii) Plug-in transfer is exact for OpQ and not for DQN. OpQ trained with the latent and
deployed on the belief mean scores exactly what blind training gives ($0.96/0.98/0.79$); a DQN
trained the same way scores $0.85/0.95/0.45$, against $0.97/0.91/1.00$ when trained on the
belief mean. (iv) \emph{Decision cost.} OpQ decides in about $0.01$\,ms; exhaustive search for
the blind optimum takes $940$ and $740$\,ms per decision on C2 and C3. (v) The oracle scores
($1.5$--$3.3$) show how much the hidden latent costs; no method that lacks it can recover this
gap.}

\begin{table}[t]
\centering\small\setlength{\tabcolsep}{4pt}
\resizebox{\textwidth}{!}{%
\begin{tabular}{@{}llcccc@{}}
\toprule
Agent & knows & latent in training & C1 & C2 & C3 \\
\midrule
OpQ-soft, learned $R$ & $K_a$ & no & 0.30 / 0.96{\scriptsize$\pm$0.00} & 0.07 / 0.97{\scriptsize$\pm$0.00} & 0.11 / 0.46{\scriptsize$\pm$0.24} \\
OpQ-soft, known $R$ & $K_a$, $R$ & no & 0.84 / 0.96{\scriptsize$\pm$0.00} & 0.97 / 0.98{\scriptsize$\pm$0.00} & 0.53 / 0.79{\scriptsize$\pm$0.00} \\
DQN on belief mean & $K_a$ (via $\bar\rho_t$) & no & 0.85 / 0.97{\scriptsize$\pm$0.02} & 0.64 / 0.91{\scriptsize$\pm$0.04} & $-0.06$ / 1.00{\scriptsize$\pm$0.00} \\
History table & none & no & 0.48 / 0.97{\scriptsize$\pm$0.01} & 0.32 / 0.72{\scriptsize$\pm$0.01} & 0.28 / 0.28{\scriptsize$\pm$0.00} \\
OpQ-soft, known $R$, plug-in & $K_a$, $R$ & yes & 0.84 / 0.96{\scriptsize$\pm$0.00} & 0.96 / 0.98{\scriptsize$\pm$0.00} & 0.38 / 0.79{\scriptsize$\pm$0.00} \\
DQN, plug-in & none & yes & 0.86 / 0.85{\scriptsize$\pm$0.08} & 0.66 / 0.95{\scriptsize$\pm$0.01} & $-0.14$ / 0.45{\scriptsize$\pm$0.24} \\
\bottomrule
\end{tabular}}
\caption{E23: score after 100 / 3000 episodes ($\pm$ s.d.\ over 3 seeds) with the initial latent
hidden. ``Knows'' lists model knowledge beyond the sampled episodes; all agents know the prior
mean.}
\label{tab:e23}
\end{table}

\begin{table}[t]
\centering\small\setlength{\tabcolsep}{4pt}
\resizebox{\textwidth}{!}{%
\begin{tabular}{@{}lcccccc@{}}
\toprule
 & belief purity & $V^\ast_{\mathrm{blind}}$ & $V_{\mathrm{rand}}$ & oracle score & value at $\bar\rho_0$ / MC over draws & decision (ms): OpQ / blind search \\
\midrule
C1 random OLDP    & 0.49 & 3.913 & $-0.027$ & 1.49 & 3.743 / 3.775{\scriptsize$\pm$0.081} & 0.011 / 2 \\
C2 SO(3) pointing & 0.50 & 4.197 & 2.683    & 2.32 & 4.180 / 4.224{\scriptsize$\pm$0.083} & 0.013 / 941 \\
C3 two-qubit      & 0.32 & 0.472 & 0.266    & 3.27 & 0.429 / 0.437{\scriptsize$\pm$0.011} & 0.012 / 737 \\
\bottomrule
\end{tabular}}
\caption{E23 references. Belief purity is $\tau(\bar\rho_{20})/N$ (1 for a pure state, $1/N$
for complete ignorance). The oracle score is that of the latent-aware optimum, which no agent
without the latent can reach. The MC column compares the exact value at $\bar\rho_0$ of a
belief-tracking policy with its mean over 400 hidden draws ($\pm$ s.e.).}
\label{tab:e23ref}
\end{table}

\rev{\textbf{What E23 does and does not show.} The exactness in (iii) is a direct consequence
of Proposition~\ref{thm:exact_rep}, so it holds by construction; the experiment shows that
this costs OpQ nothing and that it fails for a nonlinear value function. E23 assumes known
operators and a known prior mean, the setting of Proposition~\ref{thm:exact_rep}; it does not
test learning the operators or the prior.}

\section{\rev{E24: Convergence of the Sampled Soft Update}}\label{app:e24}

\rev{E24 checks Theorem~\ref{thm:sampled} numerically. OpQ-soft ($\beta=5$) is trained on
C1--C3 with random initial latents for 30,000 episodes, 3 seeds, with two step-size rules:
$\alpha_n=n^{-0.6}$, which satisfies the theorem, and $\alpha_n=\max(0.02,n^{-0.6})$, the rule
of E18--E23, which does not. Exploration is $\varepsilon$-greedy with $\varepsilon\ge0.05$, as
in E18. At each checkpoint we compare every visited entry with the exact recursion
$Q^\beta_t$ of Theorem~\ref{thm:certificate}, computed by backward induction on the true model,
and report the largest error $\max_{t,m,a}\lVert Q_t(m,a)-Q^\beta_t(m,a)\rVert_{\mathrm{op}}$
divided by $\max_{t,m,a}\lVert Q^\beta_t(m,a)\rVert_{\mathrm{op}}$. Every reachable entry is
visited (48, 90 and 40 entries; at least 317 updates each by the end), and unreachable entries
never affect the readout. Code: \texttt{experiments/e24\_convergence.py}.}

\begin{table}[t]
\centering\small\setlength{\tabcolsep}{4pt}
\resizebox{\textwidth}{!}{%
\begin{tabular}{@{}lcccccccc@{}}
\toprule
 & \multicolumn{3}{c}{largest error, $\alpha_n=n^{-0.6}$} & floor & \multicolumn{2}{c}{mean error, 30,000} & \multicolumn{2}{c}{score} \\
\cmidrule(lr){2-4}\cmidrule(lr){5-5}\cmidrule(lr){6-7}\cmidrule(lr){8-9}
Environment & 100 ep. & 1000 ep. & 30,000 ep. & 30,000 ep. & $n^{-0.6}$ & floor & learned & exact $Q^\beta$ \\
\midrule
C1 random OLDP     & 0.25 & 0.019  & 0.0098 & 0.0093 & 0.0018 & 0.0024 & 0.952 & 0.952 \\
C2 SO(3) pointing  & 0.42 & 0.0095 & 0.0027 & 0.0036 & 0.0008 & 0.0011 & 0.986 & 0.986 \\
C3 two-qubit prep. & 0.30 & 0.0045 & $1.5\times10^{-9}$ & $1.5\times10^{-9}$ & $1\times10^{-9}$ & $1\times10^{-9}$ & 0.680 & 0.680 \\
\bottomrule
\end{tabular}}
\caption{E24: distance of the learned operators from the exact certificate $Q^\beta_t$
(relative operator norm, mean over 3 seeds; learned $R$; the known-$R$ variant is within the
same range). Floor: $\alpha_n=\max(0.02,n^{-0.6})$. Score: normalised return of the greedy
readout after 30,000 episodes, and of the exact $Q^\beta$.}
\label{tab:e24}
\end{table}

\rev{\textbf{Results.} (i) The iterates converge to $Q^\beta$, as Theorem~\ref{thm:sampled}
predicts. The largest relative error falls from $0.25$--$0.42$ after 100 episodes to $0.010$
(C1) and $0.003$ (C2) after 30,000, and is still decreasing. On C3 the observable transitions
are deterministic and the reward is noise-free, so the targets carry no sampling noise and the
error falls to $1.5\times10^{-9}$, the bias of the ridge term in the reward estimate
($10^{-14}$ with $R$ known). (ii) The greedy policy reaches the certificate's greedy policy.
After 30,000 episodes the score equals that of the exact $Q^\beta$ on all three environments
($0.952$, $0.986$, $0.680$), and already does so after 1000 episodes on C1 and C2. The
remaining gap to the optimum on C1 and C3 is therefore the gap of the certificate itself
(Proposition~\ref{prop:nearopt}, E20), not a failure to converge. (iii) The step-size floor
costs little here. The two rules coincide until an entry has been updated about 680 times;
after 30,000 episodes the floor leaves a mean error of $0.0024$ instead of $0.0018$ on C1 and
$0.0011$ instead of $0.0008$ on C2, with identical scores. The largest error, at rarely
visited entries, differs little ($0.0093$ vs.\ $0.0098$ on C1, $0.0036$ vs.\ $0.0027$ on C2).}

\begin{figure}[t]
\centering
\includegraphics[width=\textwidth]{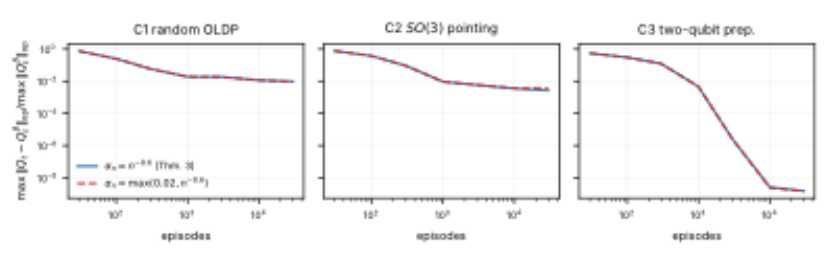}
\caption{E24: largest relative distance from the exact certificate $Q^\beta_t$ against episodes
(geometric mean over 3 seeds, learned $R$).}
\label{fig:e24}
\end{figure}

\section{Experimental Details}\label{app:expdetails}

All code is in the anonymized repository at \coderepo{}; \texttt{run\_e18.sh} and
\texttt{run\_rest.sh} reproduce every number of E7--E18, \rev{\texttt{run\_e19.py} those of
E19 and \texttt{run\_rebuttal.sh} those of E20--E24, and} \texttt{make\_figures.py}
every figure and table. Everything runs on \rev{CPU: E7--E18 take about three hours on one
core, E19 about \upd{seven} CPU-hours, and E20--E24 about 100 minutes on two cores}. Each script
fixes its random seeds\rev{, and the raw outputs of every reported run are in
\texttt{results/}}.

\subsection{E18: end-to-end control}\label{sec:e18-protocol}

\paragraph{Environments.} All three are OLDPs in the conjugation picture with
observable state $s_t=(t,m_t)$: a step counter and an observable mode whose
transitions do not depend on the latent.
\emph{C1 (random OLDP):} $N{=}3$, $K{=}4$, two modes, $H{=}6$,
$\gamma{=}0.95$; $U_a=\exp(0.8(G_a-G_a^\top))$ with $G_a$ standard Gaussian;
$R(m,a)$ random symmetric; $P(\cdot\mid m,a)$ flat Dirichlet; Gaussian reward
noise of standard deviation $0.1$.
\emph{C2 ($SO(3)$ pointing):} the latent is the dyad $\rho=3nn^\top$ of a body
axis $n$ ($n_0=e_3$); $K{=}6$ actions rotate by $\pm0.6$\,rad about the three
body axes; two modes with targets $t_1,t_2$ (random unit vectors), reward
$r_t=(t_{m_t}\cdot n_{t+1})^2=\tau(t_mt_m^\top\rho_{t+1})$, mode transitions
$P(m'\mid0)=(0.8,0.2)$ and $P(m'\mid1)=(0.3,0.7)$ for every action; $H{=}8$,
$\gamma{=}0.95$.
\emph{C3 (two-qubit preparation):} $N{=}4$, gates
$\{H\otimes I,I\otimes H,T\otimes I,I\otimes T,\mathrm{CNOT}\}$,
$\rho_0=4\lvert00\rangle\langle00\rvert$, $H{=}8$, $\gamma{=}1$; the target
$\lvert\psi\rangle$ is the output of a random 8-gate word, redrawn until it is
entangled (reduced purity $<0.9$) and far from $\lvert00\rangle$
($\lvert\langle00\vert\psi\rangle\rvert^2<0.5$), so the optimum is $1$; the
reward is the fidelity $\tau(\lvert\psi\rangle\langle\psi\rvert\rho_H)$ at the
last step and $0$ before.

\paragraph{Regimes.} \emph{Fixed}: one initial latent (C1: a random density
operator; C2: $n_0=e_3$; C3: $\lvert00\rangle$). \emph{Random}: at every
episode $\rho_0=Nvv^*$ with $v$ uniform on the real (C1, C2) or complex (C3)
unit sphere; evaluation averages over $8$ held-out draws.
\emph{Transfer} (Table~\ref{tab:e18}, last column per environment): agents are
trained in the fixed regime and evaluated, without retraining, on the same
$8$ held-out initial latents.

\paragraph{Agents.} \rev{All agents observe $s_t$ and the rewards, and either $\rho_t$ (OpQ, DQN),
the action history with $\rho_0$ at step $0$ (DRQN), or the action and mode history only
(history table). The agents do not have the same knowledge: OpQ also uses the dual maps
$\mathcal K_a^\dagger$, the history-table key omits $\rho_0$ (so it is a sufficient statistic only
in the fixed regime), and DRQN's $\rho_0$ input is constant in fixed-regime training. E19
(Appendix~\ref{app:e19}) gives the comparisons with matched knowledge.}
\emph{OpQ-soft}: one $N\times N$ operator per $(t,m,a)$, updated by the
sample form of~\eqref{eq:soft},
$Q(s,a)\leftarrow Q(s,a)+\alpha\bigl(\Kd a(\hat R(m,a)+\gamma\LSE_\beta(Q(s',\cdot)))-Q(s,a)\bigr)$,
with $\beta{=}5$, and acting greedily on $\tau(\rho_tQ(s_t,a))$. The update
is unbiased for~\eqref{eq:soft} because~\eqref{eq:soft} is linear in
$P(\cdot\mid s,a)$. The reward observable $\hat R(m,a)$ is estimated online
by least squares from the scalar rewards (ridge $10^{-6}$, orthonormal
Hermitian basis), since $r=\tau(R\rho')$ is linear in $\rho'$; the
\emph{known-$R$} variant uses the true $R$.
\emph{OpQ-PI}: operator TD~\eqref{eq:update} evaluating the uniform policy
(behaviour is uniform throughout), followed by the latent-aware greedy
improvement of Prop.~\ref{prop:improve}.
\emph{Scalar $Q$ on $(t,m)$} and \emph{scalar $Q$, history table}
(keyed by $(t,m,a_1m_1\dots a_tm_t)$): tabular $Q$-learning.
The step size of all tabular and operator agents is
$\alpha=\max(0.02,n^{-0.6})$, where $n$ counts visits to the entry.
\emph{DQN on $\rho_t$}: MLP $[\,$one-hot $t$, one-hot $m$, coordinates of
$\rho_t\,]\to64\to64\to K$ (ReLU), Adam ($10^{-3}$), batch $64$, replay
$5\times10^4$, target network every $250$ steps, one update per step.
\emph{DRQN}: GRU with $64$ hidden units on
$[\,$previous action, previous reward, mode, one-hot $t$, $\rho_0$ at step $0\,]$,
Adam ($10^{-3}$), $32$ episodes per batch, $4$ updates per episode, target
network every $50$ episodes, replay of $5000$ episodes.
Exploration is $\epsilon$-greedy, annealed linearly from $1$ to $0.05$ over
the first $60\%$ of $3000$ episodes, for every agent except OpQ-PI.

\paragraph{Evaluation.} The greedy policy of each agent is evaluated
\emph{exactly}: all mode branches are enumerated and mean rewards are used,
so the reported curves contain no Monte-Carlo noise. $V^*$ is computed by
backward dynamic programming over all $K^H$ action histories and modes;
$V_{\mathrm{rand}}$ comes from Proposition~\ref{thm:exact_rep} applied to the
uniform policy. Scores are $(V-V_{\mathrm{rand}})/(V^*-V_{\mathrm{rand}})$,
averaged over initial latents; we report the mean and standard deviation
over $5$ seeds. Code: \texttt{opq/envs.py}, \texttt{opq/agents.py},
\texttt{opq/deep\_agents.py}, \texttt{experiments/e18\_control.py},
\texttt{experiments/e18\_transfer.py}.

\subsection{E8, E13--E16: representation benchmark}\label{sec:rep-protocol}

The target is $f(w)=\tfrac1N\operatorname{Re}\Tr(WA_{a_1}\cdots A_{a_T})$ for
uniformly random words ($3000$ training, $1000$ test).
\emph{Orthogonal} (E8/E13): $N{=}5$, $K{=}4$, $T{=}6$,
$A_a=\exp(G_a-G_a^\top)$ with $G_a$ having i.i.d.\ $\mathcal N(0,0.6^2)$ entries,
$W$ random symmetric.
\emph{$\mathfrak{so}(3)$} (E14): $K{=}6$ rotations by $\pm0.6$\,rad about the
coordinate axes, $T{=}8$, $W$ Gaussian.
\emph{$\mathfrak{se}(3)$ hinge chain} (E16): $N{=}4$ (homogeneous
coordinates); each of $K{=}6$ creases is a revolute joint with a random unit
axis $n_a$, a Gaussian hinge point $p_a$ and an angle
$\theta_a\sim U[0.5,1.5]$, with twist
$\Omega_a=\begin{psmallmatrix}\theta_a[n_a]_\times&-\theta_a[n_a]_\times p_a\\0&0\end{psmallmatrix}$;
$T{=}8$, $W$ Gaussian.
\emph{Displacements} (E15): Fock space truncated at $N_F{=}6$,
$D(\alpha_a)=\exp(\alpha_aa^\dagger-\bar\alpha_aa)$ for $K{=}6$ random
$\alpha_a$ ($\lvert\alpha_a\rvert\approx0.5$), $T{=}6$, $W$ random Hermitian;
complex matrices are encoded in real form as
$X\mapsto\begin{psmallmatrix}\operatorname{Re}X&-\operatorname{Im}X\\\operatorname{Im}X&\operatorname{Re}X\end{psmallmatrix}$,
so the real operator dimension that suffices is $2N_F{=}12$.
In the untruncated algebra $\ad^2_{\Omega_a}=0$, but truncation breaks
nilpotency. The depth profile $S(10)$ is $3.0\times10^4$, $1.1\times10^6$
and $2.4\times10^7$ at $N_F=6,12,24$, so it grows rather than vanishes. The
order-dependent phase of one swapped pair, $\arg\langle0\vert
D(\alpha)D(\beta)\vert0\rangle/\langle0\vert D(\beta)D(\alpha)\vert0\rangle
=-0.3596$\,rad, matches the untruncated value
$2\operatorname{Im}(\alpha\bar\beta)$ to four digits. In the $\mathfrak{se}(3)$
chain the same two folds applied in opposite orders move the marked point
$(1,1,0)$ by up to $2.00$.

Before any fitting, a planted-solution check evaluates the operator model with
the \emph{true} generators; its test MSE is at floating-point level in every
environment (Table~\ref{tab:rep}).
\emph{OpQ$(d)$}: $K$ learned $d\times d$ operators and a readout,
$g(w)=\Tr(W_dQ_{a_1}\cdots Q_{a_T})$, initialised at $Q_a=\exp$(random
skew-symmetric, scale $0.3$); full-batch Adam ($0.02$, cosine decay, $1500$
epochs, gradient clipping at $1$), best of $4$ restarts by training loss,
followed by at most $2000$ L-BFGS iterations.
\emph{Baselines}: ordinary least squares on one-hot $(t,a_t)$ (order-blind)
and on one-hot $(t,a_{t-1},a_t)$ (windowed table); GRUs with $4,16,64$ hidden
units and an LSTM with $64$; a single-head self-attention block (width $32$,
learned positional embeddings, feed-forward block, mean pooling); and a
1D-CNN (learned $3$-dimensional action embedding, two convolutions with
kernel $3$ and $16$ channels, mean pooling). Neural baselines are trained
with full-batch Adam ($0.01$; $0.005$ for attention; cosine decay,
$1000$ epochs). For linear models the parameter count is the rank of the
design matrix.
Code: \texttt{opq/rep\_models.py}, \texttt{experiments/rep\_benchmark.py}.

\subsection{E10: nilpotent versus non-nilpotent algebras}\label{sec:e10-protocol}
$\mathfrak{so}(3)$ (generators $L_x,L_y,L_z$) and the Heisenberg algebra
$\mathfrak n_3$ (strictly upper-triangular $X,Y,Z$ with $[X,Y]=Z$), both in
dimension $3$ with every generator rescaled to Frobenius norm $\sqrt2$, so that
the nilpotency step is the only difference; $W$ is a general Gaussian matrix
(a symmetric $W$ would annihilate every $\mathfrak{so}(3)$ bracket term). (a)
$S(T)=\sum_{t=2}^T\max_{a\ne b}\lvert\Tr(W\ad_{\Omega_a}^{t-1}\Omega_b)\rvert/N$ by
exact iterated bracketing. (b) For $T=2,\dots,8$ and $400$ random words
($100$ for $T\ge7$), the return is averaged over all $T!$ reorderings of the
word; this average is the best predictor available from the action counts.
We record the mean absolute deviation of the true return from it, normalised
by the standard deviation of returns at that $T$.
Code: \texttt{experiments/e10\_depth\_profile.py}.

\subsection{E7, E9, E12: the operator log-sum-exp and the commutative reduction}\label{sec:e7-protocol}\label{sec:e12-protocol}\label{sec:e12-results}
\emph{E7.}
\begin{itemize}
\item[(i)] $20{,}000$ commuting configurations ($N\in\{2,\dots,5\}$,
$K\in\{2,3,4\}$, $\beta$ cycling through $\{0.05,0.1,0.25,0.5,1,2,5,10\}$,
shared random orthonormal eigenbasis, eigenvalues in $[-1,1]$); half use
local perturbations and half independent draws.
\item[(ii)] The construction of Prop.~\ref{prop:lse_lipschitz_new}(c) at
$K{=}N{=}25$, $\beta{=}5$, $\epsilon{=}10^{-6}$.
\item[(iii)] $80$ non-commuting configurations per $\beta\in\{1,5,10\}$
($N,K\in\{2,3\}$, perturbations at scale $10^{-12}$), evaluated in $80$-digit
arithmetic with \texttt{mpmath}. In double precision, $\sum_ie^{\beta x_i}$
loses its smallest eigenvalues once $e^{2\beta M}$ approaches $10^{16}$, which
produces spurious ratios in the thousands.
\end{itemize}
\emph{E9.} Three observable states, $K{=}3$, $\gamma{=}0.9$, $\beta{=}3$,
diagonal unitaries $U_a=\operatorname{diag}(e^{i\phi})$ and diagonal
$R(s,a)$, for $N\in\{1,4\}$. We compare (a) $200$ iterations of the exact
operator recursion against $N$ independent scalar soft Bellman iterations, and
(b) $20{,}000$ sample-based updates fed to both the operator learner and $N$
scalar soft $Q$-learners with the same samples and step sizes.
\emph{E12.} We apply the aggregation $x\mapsto R+\gamma\LSE_\beta(x)$ to
$K{=}4$ self-adjoint $5\times5$ inputs ($\gamma{=}0.95$, $\beta{=}2$). We draw
$y_i=x_i+P_i$ with $P_i\succeq0$ at scale $0.5$ ($2000$ trials each for
commuting and generic inputs), then sweep the scale over
$\{0.5,0.1,0.01,10^{-3},10^{-4}\}$ with $1000$ generic trials each. A trial
counts as a violation when $\lambda_{\min}$ of the output difference is below
$-10^{-12}$.
Code: \texttt{experiments/e07\_lse\_lipschitz.py},
\texttt{e09\_commutative\_reduction.py}, \texttt{e12\_monotonicity.py}.

\subsection{E17: exact checks of the compositional recursion}\label{sec:e17-protocol}
A random OLDP with $|\mathcal S|{=}2$, $K{=}3$, $N{=}3$, $H{=}4$,
$\gamma{=}0.9$, $P$ flat Dirichlet, $U_a=\exp(0.8(G-G^\top))$, $R$ random
symmetric.
\emph{Proposition~\ref{thm:exact_rep}}: $Q^\pi_0$ from the backward recursion is
compared with the exact expected return, obtained by enumerating every branch,
at $20$ random $(s,a,\rho)$, for $\rho$ a density operator (CP) and a rotation
(MP).
\emph{Theorem~\ref{thm:td}}: the infinite-horizon $Q^\pi$ is computed by $400$
Bellman iterations; operator TD uses i.i.d.\ uniform $(s,a)$ and
$\alpha_k=(1+k/200)^{-0.7}$.
\emph{Theorem~\ref{thm:certificate}}: $Q^\beta_0$ for $\beta\in\{1,5,20,50\}$,
with $\LSE_\beta$ evaluated in $120$-digit arithmetic, is compared with $q^*$
from exhaustive latent-aware dynamic programming at $30$ random $(s,a,\rho)$,
and $\lambda_{\min}(Q^\beta_0-Q^\pi_0)$ is taken over $5$ random latent-blind
policies. The commuting check uses diagonal sign matrices and diagonal $R$ at
$\beta{=}200$ and pure eigenstates.
\emph{Proposition~\ref{thm:representation_gap}}: three generators at scale $0.6$,
Gaussian $W$, $T{=}3$.
\emph{Proposition~\ref{thm:upper}}: a Hankel block over the $40$ words of length
$\le3$, with numerical rank taken at relative threshold $10^{-9}$.
Code: \texttt{experiments/e17\_exact\_checks.py}.

\subsection{Computational requirements}\label{sec:compute}
Python~3.12 with NumPy, SciPy, mpmath, Matplotlib, JAX (CPU) and Optax. On one CPU core: E18 learning curves $\approx$ 70 minutes; E18 transfer $\approx$ 25 minutes; E17 $\approx$ 1 minute; E7 $\approx$ 10 minutes (dominated by 80-digit arithmetic); E9, E10, E12 under 2 minutes each; representation benchmark $\approx$ 15 minutes per environment. \rev{E19 runs every job in a separate process and can resume; E20 takes about six minutes and E21 about twenty.}

\end{document}